\documentclass[nohyperref]{article}

\usepackage{microtype}
\usepackage{graphicx}
\usepackage{booktabs} % for professional tables

\usepackage[dvipsnames]{xcolor}
\definecolor{royalblue}{RGB}{65, 105, 225}

\usepackage{hyperref}

\usepackage[accepted]{icml2022}

\usepackage{amsmath,amssymb,amsfonts}

\usepackage{algorithm}
\usepackage[noend]{algpseudocode}
\usepackage{graphicx}
\usepackage{graphics}
\usepackage{dblfloatfix}
\usepackage{bbm}
\usepackage{arydshln}
\usepackage{multirow} %added new
\usepackage{booktabs} %added new
\usepackage{amssymb} %added new
\usepackage{microtype}
\usepackage{booktabs}
\usepackage{mathtools}
\DeclarePairedDelimiterX{\infdivx}[2]{(}{)}{%
  #1\;\delimsize\|\;#2%
}
\newcommand{\infdiv}{\mathcal{D}_{KL}\infdivx}

\newcommand*\circled[1]{\tikz[baseline=(char.base)]{
            \node[shape=circle,draw,inner sep=1pt] (char) {#1};}}

\DeclareMathOperator*{\argmaxB}{argmax}
\usepackage{graphicx}
\usepackage{graphics}
\usepackage{multirow} %added new
\usepackage{booktabs} %added new
\usepackage{amssymb} %added new
\usepackage{hhline}
\usepackage{appendix}
\usepackage{caption}
\usepackage{subcaption}
\usepackage{enumitem}

\usepackage{amsmath}
\usepackage{amssymb}
\usepackage{mathtools}
\usepackage{amsthm}

\usepackage[capitalize,noabbrev]{cleveref}

\usepackage[textsize=tiny]{todonotes}

\icmltitlerunning{SoQal: Selective Oracle Questioning for Consistency Based Active Learning of Cardiac Signals}

\begin{document}

\twocolumn[
\icmltitle{SoQal: Selective Oracle Questioning for Consistency Based \\ Active Learning of Cardiac Signals}



\icmlsetsymbol{equal}{*}

\begin{icmlauthorlist}
\icmlauthor{Dani Kiyasseh}{A}
\icmlauthor{Tingting Zhu}{B}
\icmlauthor{David Clifton}{B}
\end{icmlauthorlist}

\icmlaffiliation{A}{Department of Computing and Mathematical Sciences, California Institute of Technology, California, USA}
\icmlaffiliation{B}{Department of Engineering Science, University of Oxford, Oxford, UK}

\icmlcorrespondingauthor{Dani Kiyasseh}{dkiyass1@caltech.edu}

\icmlkeywords{Machine Learning, ICML}

\vskip 0.3in
]



\printAffiliationsAndNotice{}  % leave blank if no need to mention equal contribution

\begin{abstract}
Clinical settings are often characterized by abundant unlabelled data and limited labelled data. This is typically driven by the high burden placed on oracles (e.g., physicians) to provide annotations. One way to mitigate this burden is via active learning (AL) which involves the (a) acquisition and (b) annotation of informative unlabelled instances. Whereas previous work addresses either one of these elements independently, we propose an AL framework that addresses both. For acquisition, we propose Bayesian Active Learning by Consistency (BALC), a sub-framework which perturbs both instances and network parameters and quantifies changes in the network output probability distribution. For annotation, we propose SoQal, a sub-framework that dynamically determines whether, for each acquired unlabelled instance, to request a label from an oracle or to pseudo-label it instead. We show that BALC can outperform start-of-the-art acquisition functions such as BALD, and SoQal outperforms baseline methods even in the presence of a noisy oracle.
\end{abstract}

\section{Introduction}

Deep learning algorithms often need access to abundant, high-quality, labelled data. Such access, though, is a rarity within healthcare. For example, in low-resource clinical settings, data can be collected in abundance via wearable sensors however physicians, who are typically required to annotate such data, are either in low-supply or ill-trained to complete the task. As such, these settings are often characterized by abundant \textit{unlabelled} data and limited \textit{labelled} data. In light of this observation, we focus on the following question: \textit{how can we design clinical algorithms that exploit abundant, unlabelled data and limited, labelled data while minimizing the labelling burden placed on physicians?}

One way to address this question is via the active learning (AL) framework \cite{Settles2009} which iterates over three main steps. 1) \textit{acquisition} - a learner (e.g., neural network) is tasked with acquiring unlabelled instances 2) \textit{annotation} - an oracle (e.g., physician) is tasked with labelling such acquired instances, and 3) \textit{aggregation} - the learner is trained on the existing and newly-labelled instances. Whereas previous work addresses either one of the first two steps of active learning, in this work, we aim to modify both. 

\begin{figure}[!t]
        \centering
        \begin{subfigure}{1\columnwidth}
        \centering
        \includegraphics[width=\textwidth]{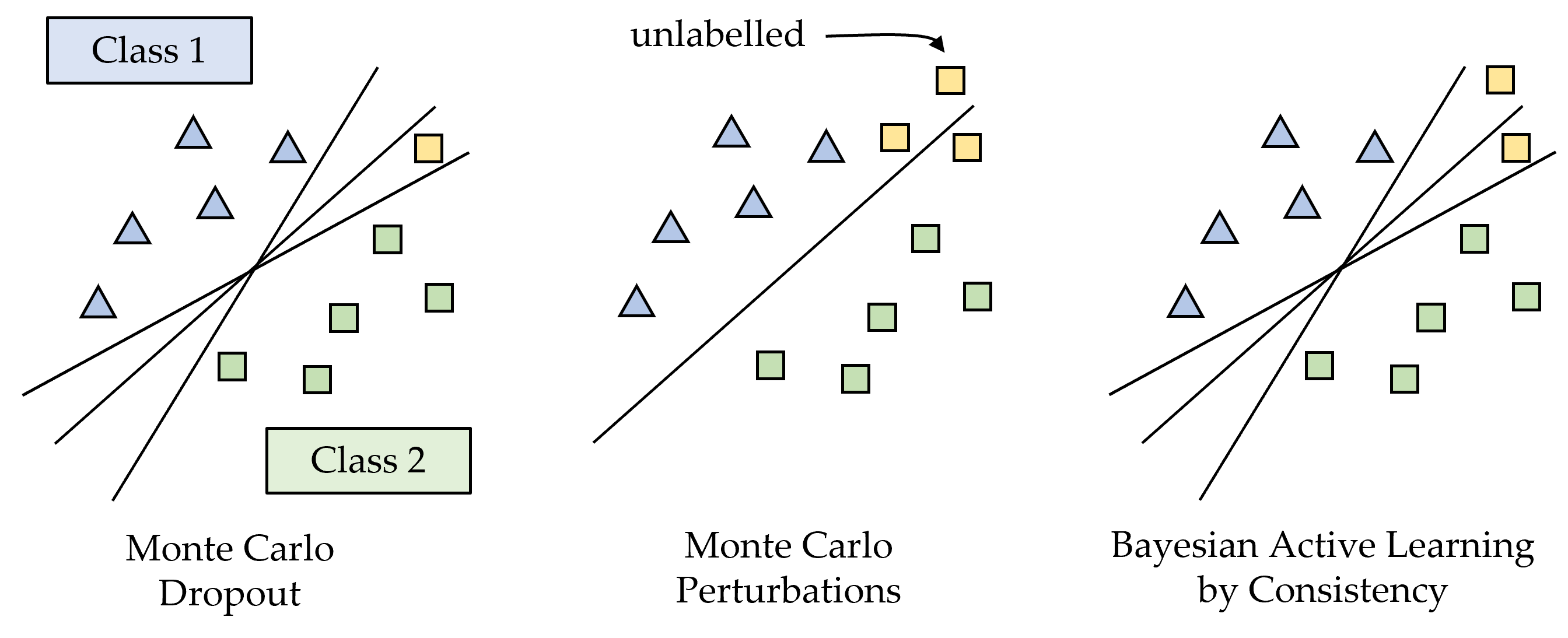}
        \end{subfigure}
\caption{\textbf{Perturbation frameworks used in active learning.} Labelled instances from two classes and unlabelled instances (yellow) with the hypotheses of \textbf{(Left)} Monte Carlo Dropout (MCD), where each MC sample is a distinct hypothesis, \textbf{(Centre)} Monte Carlo Perturbations (MCP), with one hypothesis but several perturbed variants of the unlabelled instance, and \textbf{(Right)} Bayesian Active Learning by Consistency (BALC) with multiple hypotheses and perturbed instances. In this paper, we introduce MCP and BALC as alternatives to MCD in order to acquire unlabelled instances.}
\label{fig:version_space}
\end{figure}

The acquisition of unlabelled instances is commonly performed via an acquisition function, an example of which is Bayesian Active Learning by Disagreement (BALD) \cite{Houlsby2011}. It quantifies the degree to which a network is uncertain about the classification of an instance. To do so, Monte Carlo Dropout (MCD) \cite{Gal2016} is exploited wherein network parameters are stochastically perturbed while an unlabelled instance is passed through a network. These parameter perturbations manifest in the form of distinct hypotheses (decision boundaries), as shown in Fig.~\ref{fig:version_space} (left). However, MCD, as we later outline, can erroneously overlook informative instances for acquisition due to its misspecification of parameter perturbations. As for annotating acquired instances, it is often assumed that oracles are continuously available and sufficiently skilled to provide such annotations, an assumption that is rarely satisfied within healthcare settings. 

In this paper, we make the following contributions. \textbf{First}, we propose an AL framework, Monte Carlo Perturbations (MCP), that involves perturbing instances and observing concomitant changes in the network output distribution. We show that MCP performs on par with MCD in several settings. \textbf{Second}, we take inspiration from consistency training and propose an AL framework, Bayesian Active Learning by Consistency (BALC), that involves perturbing \textit{both} instances and parameters and observing changes in the output distribution. Moving away from uncertainty-based acquisition functions, we introduce two \textit{consistency-based} acquisition functions, $\mathrm{BALC_{KLD}}$ and $\mathrm{BALC_{JSD}}$, and show that they can outperform state-of-the-art acquisition functions such as $\mathrm{BALD}$. \textbf{Third}, existing acquisition functions are static; they determine the informativeness of an instance at a single snapshot in time. Instead, we propose the simple modification of \textit{tracking} the value of an acquisition function over time (epochs). We show that acquisition functions with such temporal information can outperform their static counterparts. \textbf{Fourth}, we take inspiration from selective classification and propose $\mathrm{SoQal}$, a framework which determines whether, for each acquired unlabelled instance, to request a label from an oracle or to pseudo-label it instead. We show that $\mathrm{SoQal}$ outperforms state-of-the-art selective classification methods. To the best of our knowledge, we are the first to explore these avenues in the context of cardiac signals.

\section{Related Work}

\subsection{Active Learning in Healthcare} 
For medical images, previous work has acquired instances by measuring their distance in a latent space to images in the training set \cite{Smailagic2018, Smailagic2020}. For time-series data, researchers have acquired instances from electronic health record (EHR) \cite{Gong2019} and electrocardiogram (ECG) \cite{Wiens2010,Pasolli2010,Wang2019,Kiyasseh2021} databases. For example, although \citet{Kiyasseh2021} acquire cardiac signals, they do so in the context of continual learning and do not explore how to annotate such cardiac signals. More generally, \citet{Gal} adopt BALD \cite{Houlsby2011} in the context of Monte Carlo Dropout to acquire instances that maximize the Jensen-Shannon divergence (JSD) across MC samples. For an in-depth review of active learning, we refer readers to \citet{Settles2009}.

\subsection{Oracles in Active Learning}
Previous work attempts to learn from multiple or imperfect oracles \cite{Dekel2012,Zhang2015,Sinha2019}. For example, Urner \textit{et al.} \cite{Urner2012} identify a suitable oracle to label a particular instance. In contrast to our work, they do not explore the independence of a learner from an oracle. Although \citet{Yan2016} consider oracle abstention, we place the decision of abstention under the control of the learner. To the best of our knowledge, we are the first to explore a dynamic oracle selection strategy in the context of cardiac signals.

\subsection{Consistency Training} 
Consistency training helps enforce the smoothness assumption \cite{McCallumzy1998,Zhu2005,Verma2019}. For example, \citet{Xie2019} penalize networks for generating drastically different outputs in response to perturbed instances, resulting in perturbation-invariant representations. Most similar to our work is that of \citet{Gao2020} which designs a consistency-loss based on the Kullback-Leibler divergence $\mathcal{D}_{KL}$. Whereas they actively acquire instances using the variance of the probability assigned to each class by the network in response to perturbed versions of the same instance, we design distinct consistency-based acquisition functions. 

\subsection{Selective Classification} 
Selective classification imbues a network with the ability to abstain from making predictions \cite{Chow1970,El-Yaniv2010,Cortes2016}. \citet{Wiener2011} use a support vector machine to rank and reject instances based on the degree of disagreement between hypotheses. In some frameworks, these are the same instances that active learning views as most informative. Although \citet{Ziyin2019} propose the gambler's loss to learn a selection function that determines whether instances are rejected, this approach is not implemented in the context of active learning. SelectiveNet \cite{Geifman2019} introduces a multi-head architecture similar to ours, however it assumes the presence of ground-truth labels and, therefore, does not trivially extend to unlabelled instances.

\section{Background}

\subsection{Active Learning}
\label{sec:AL_101}

Consider a learner $f_{\theta}: \boldsymbol{x} \in \mathbb{R}^{D} \rightarrow \boldsymbol{v} \in \mathbb{R}^{E}$, parameterized by $\boldsymbol{\theta}$, that maps a $D$-dimensional instance, $\boldsymbol{x}$, to an $E$-dimensional representation, $\boldsymbol{v}$. Further consider a function, $p_{\omega}: \boldsymbol{v} \in \mathbb{R}^{E} \rightarrow y \in \mathbb{R}^{C}$, that maps an $E$-dimensional representation, $\boldsymbol{v}$, to a $C$-dimensional output, $y$, where $C$ is the number of classes. After training on a pool of labelled data, $\mathcal{D}_{L} = \{ \boldsymbol{X_{L}},\boldsymbol{Y_{L}} \} $ for $\tau$ epochs, the learner is tasked with querying the unlabelled pool of data, $\boldsymbol{X_{U}}$, and acquiring the top $b$ fraction of instances, $\{\boldsymbol{x}_{i}\}_{i=1}^{b} \sim \boldsymbol{X_{U}}$, that it deems to be most informative. The degree of informativeness of an instance, $\boldsymbol{x}$, is determined by an acquisition function, $\alpha(\boldsymbol{x}) \in \mathbb{R}$, such as Bayesian Active Learning by Disagreement ($\mathrm{BALD}$) \cite{Houlsby2011}. These functions are typically used alongside Monte Carlo Dropout ($\mathrm{MCD}$) \cite{Gal2016} to identify instances that lie in the region of uncertainty, a region in which hypotheses disagree the most about instances (Fig.~\ref{fig:version_space} left).

\subsection{Consistency Training}
Consider an unlabelled instance, $\boldsymbol{x} \sim \boldsymbol{X_{U}}$, and its perturbed counterpart, $\boldsymbol{x'} = \boldsymbol{x} + \epsilon$, where $\epsilon$ is some perturbation. A network is said to be invariant to such a perturbation if its outputs, $p_{\omega}(\boldsymbol{x})$ and $p_{\omega}(\boldsymbol{x'})$, are similar to one another. Consistency training is one way to encourage such an invariance. In this work, we exploit this intuition to design acquisition functions, as outlined next. 

\section{Methods}

\subsection{Monte Carlo Perturbations}
\label{sec:MCP}

\begin{figure}[!t]
    \centering
    \begin{subfigure}{1\columnwidth}
        \centering
        \includegraphics[width=1\textwidth]{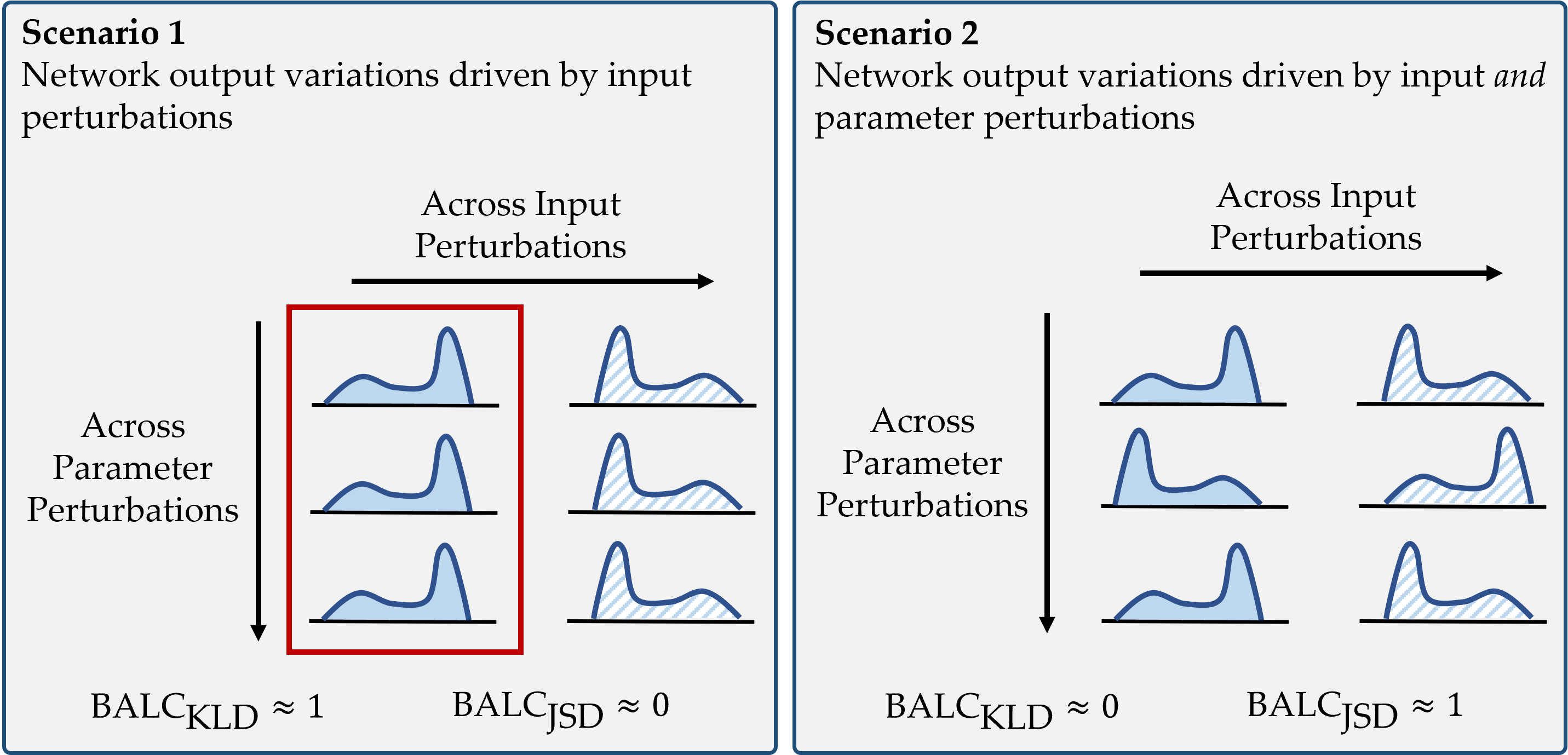}
    \end{subfigure}
    \caption{\textbf{Scenarios demonstrating the effect of input and parameter perturbations on the behaviour of the proposed acquisition functions, $\mathrm{BALC}_{\mathrm{KLD}}$ and $\mathrm{BALC}_{\mathrm{JSD}}$.} \textbf{(Scenario 1)} network output variations caused primarily by input perturbations. The red rectangle illustrates how parameter perturbations alone, as with MCD, can be insufficient in causing changes to the network outputs, thus failing to acquire informative unlabelled instances. \textbf{(Scenario 2)} network output variations caused by both input and parameter perturbations. We show that while $\mathrm{BALC}_{\mathrm{KLD}}$ is likely to acquire instances due to input perturbations, $\mathrm{BALC}_{\mathrm{JSD}}$ considers both input and parameter perturbations when performing acquisitions.}
    \label{fig:intuition_figure}
\end{figure}

Acquisition functions dependent upon parameter perturbations, such as in $\mathrm{MCD}$, can overlook, and thus fail to acquire, informative unlabelled instances. To see this, and without loss of generality, consider an unlabelled instance which is (a) in proximity to a decision boundary, thus deeming it informative for training \cite{Settles2009} and (b) classified by a network into an arbitrary class. In this setting, a total of $T$ parameter perturbations results in $T$ hypotheses, altering the outputs of a network, $\{ p_{\omega}^{i} \}_{i=1}^{T}$, in response to an unlabelled instance. We visualize the distribution of such outputs in Fig.~\ref{fig:intuition_figure} (red rectangle), after having applied $T=3$ parameter perturbations. If the perturbations happen to be too small in magnitude, for example, then a network will exhibit a similar output distribution ($p_{\omega}^{1} = p_{\omega}^{2} = \cdots$) across the parameter perturbations. However, since acquisition functions assign value based on \textit{changes} in the output distribution, this informative unlabelled instance (due to its proximity to decision boundary) would be \textit{erroneously} deemed uninformative.

One way to avoid missing these informative instances is by stochastically perturbing instances (instead of network parameters) and observing changes in the network outputs, a setup we refer to as Monte Carlo Perturbations ($\mathrm{MCP}$). This results in a single hypothesis but multiple perturbed variants of the instance (see Fig.~\ref{fig:version_space} centre). The intuition is that network outputs will differ more significantly for an instance in proximity to the decision boundary than for an instance farther away. By quantifying these output changes, as is done with almost any acquisition function, we can identify informative instances for acquisition. The main advantage of MCP over MCD is the increased control and interpretability of the applied perturbations; perturbations applied to instances are likely to be more understandable than those applied to parameters. A formal derivation of $\mathrm{MCP}$ can be found Appendix~\ref{appendix:mcp_derivation}.

\subsection{Bayesian Active Learning by Consistency}
\label{sec:BALC}

\begin{figure}[!h]
\centering
\begin{subfigure}{1\columnwidth}
        \centering
        \includegraphics[width=\textwidth]{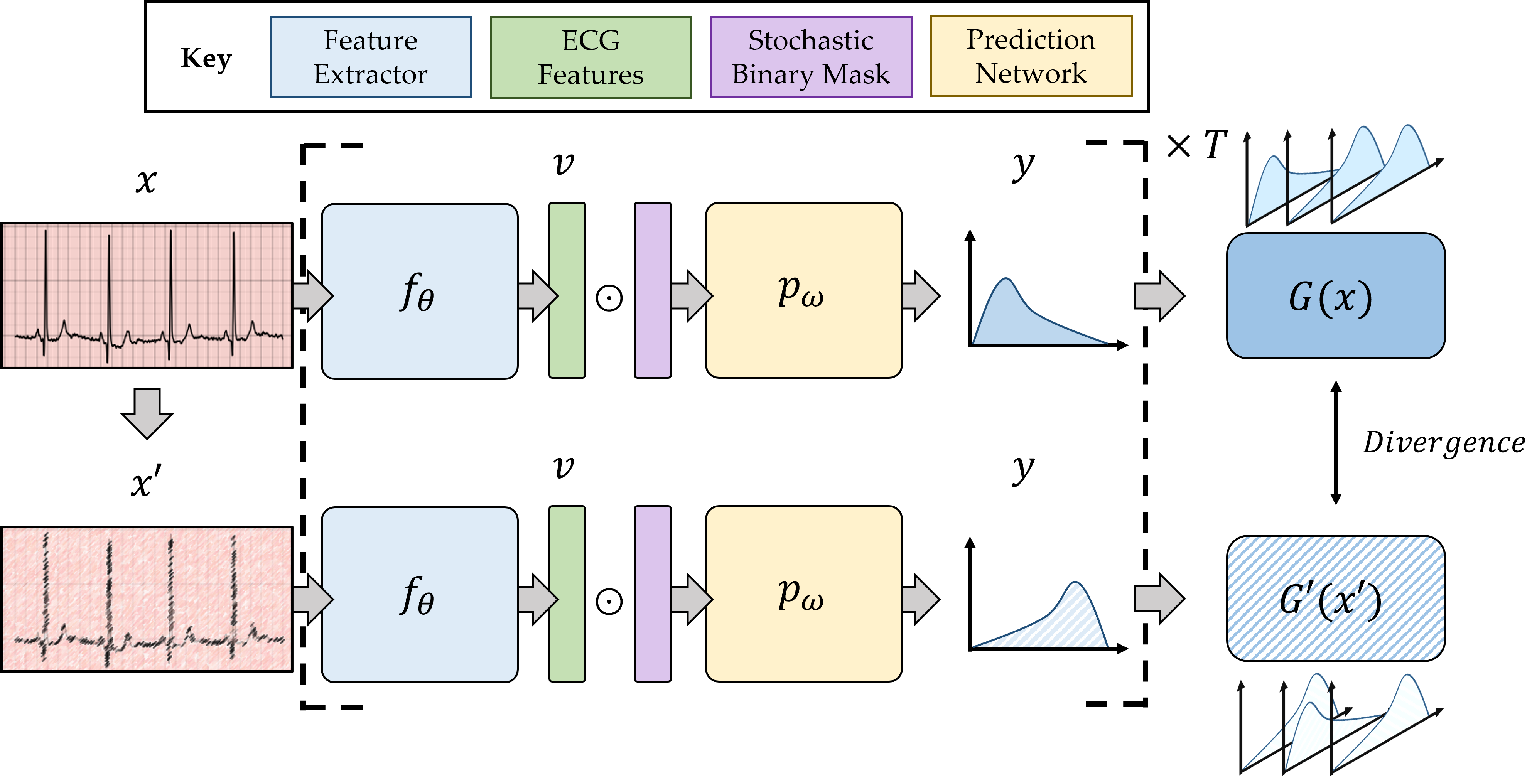}
\end{subfigure}
        \caption{\textbf{Consistency-based active-learning framework.} We perturb an instance, $\boldsymbol{x}$, to generate, $\boldsymbol{x'}$, and extract their corresponding representations, $v$. We apply a stochastic dropout mask to these representations and obtain an output distribution over the classes. When repeated $T$ times, this generates a pair of matrices, $\boldsymbol{G(x)}$ and $\boldsymbol{G'(x')}$, whose divergence is calculated via $\mathrm{BALC}_{\mathrm{KLD}}$ or $\mathrm{BALC}_{\mathrm{JSD}}$.}
        \label{fig:BALC_pipeline}
\end{figure}

It could be argued that the same limitations exhibited by MCD also extend to MCP. After all, both frameworks apply perturbations to either network parameters or instances. Acknowledging this, we propose a framework, entitled Bayesian Active Learning by Consistency ($\mathrm{BALC}$), in which perturbations are simultaneously applied to network parameters \textit{and} instances. As such, this results in multiple decision boundaries and perturbed instances (see Fig.~\ref{fig:version_space} right). $\mathrm{BALC}$ consists of three main steps: 1) perturb an instance, $\boldsymbol{x} \in \mathbb{R}^{D}$, to generate $\boldsymbol{x'} \in \mathbb{R}^{D}$, 2) perturb the network parameters, $\boldsymbol{\theta}$, to generate $\boldsymbol{\theta'}$, and 3) pass both instances, $\boldsymbol{x}$ and $\boldsymbol{x'}$, through the \textit{perturbed} network, generating outputs, $p(y|\boldsymbol{x},\boldsymbol{\theta'})$ and $p(y|\boldsymbol{x'},\boldsymbol{\theta'}) \in \mathbb{R}^{C}$, respectively. Note that we drop the explicit dependence on $\boldsymbol{\omega}$ for clarity. We perform these steps for $T$ stochastic parameter perturbations and generate two matrices of network outputs, $\boldsymbol{G(x)}, \boldsymbol{G'(x')} \in \mathbb{R}^{T \times C}$, as shown in Fig.~\ref{fig:BALC_pipeline}.

The intuition is that the greater the divergence between $\boldsymbol{G}$ and $\boldsymbol{G'}$, the closer an instance is to the decision boundary. To that end, we propose two consistency-based acquisition functions, $\mathrm{BALC}_{\mathrm{KLD}}$ and $\mathrm{BALC}_{\mathrm{JSD}}$. 

To calculate $\mathrm{BALC}_{\mathrm{KLD}}(\boldsymbol{x}) \in \mathbb{R}$ \eqref{eq4_main}, we first empirically fit two $C$-dimensional Gaussian distributions, $\mathcal{N}(\boldsymbol{x})$ and $\mathcal{N}(\boldsymbol{x'})$ to $\boldsymbol{G}$ and $\boldsymbol{G'}$, respectively. For each such matrix, we obtain an empirical mean vector, $\boldsymbol{\mu} = \frac{1}{T} \sum_{i}^{T} \boldsymbol{G}_{i}$ and covariance matrix, $\boldsymbol{\Sigma} =  (\boldsymbol{G} - \boldsymbol{\mu})^{T}(\boldsymbol{G} - \boldsymbol{\mu})$. We then quantify the $D_{KL}$ between $\mathcal{N}(\boldsymbol{x})$ and $\mathcal{N}(\boldsymbol{x'})$ as follows.
\begin{equation} 
\label{eq4_main}
\mathrm{BALC}_{\mathrm{KLD}}(\boldsymbol{x}) = \infdiv{\mathcal{N}(\boldsymbol{x})}{\mathcal{N}(\boldsymbol{x'})}
\end{equation}
where $\mathcal{N}(\boldsymbol{x}) = \mathcal{N}(\boldsymbol{\mu(x)},\boldsymbol{\Sigma(x)})$ and $\mathcal{N}(\boldsymbol{x'}) = \mathcal{N}(\boldsymbol{\mu(x')},\boldsymbol{\Sigma(x')})$. 

We note that $\mathrm{BALC}_{\mathrm{KLD}}$ is likely to detect changes in the network outputs due solely to instance perturbations. To see this, consider Scenario 1 and 2 presented in Fig.~\ref{fig:intuition_figure}. Changes in network outputs are caused by either instance perturbations alone (Scenario 1) or both instance and parameter perturbations (Scenario 2). In these two scenarios, $\mathrm{BALC}_{\mathrm{KLD}} \approx 1$ and $0$, respectively. Since the informativeness of an instance is related to the magnitude of the acquisition function, these scenarios suggest that $\mathrm{BALC}_{\mathrm{KLD}}$ has a preference for instance perturbations. In order to detect changes in the network outputs due to \textit{both} instance and parameter perturbations, we introduce $\mathrm{BALC}_{\mathrm{JSD}}$. Support for this claim can also be found in Fig.~\ref{fig:intuition_figure}.

$\mathrm{BALC}_{\mathrm{JSD}}(\boldsymbol{x}) \in \mathbb{R}$ \eqref{eq3_main} comprises the difference of two terms. The first term, \circled{A}, calculates the $D_{KL}$ of network outputs due to a single instance perturbation and averages this across $T$ parameter perturbations. The second term, \circled{B}, averages the network outputs across parameter perturbations independently for the original and perturbed instance before calculating the $D_{KL}$ of the resulting mean outputs. The full derivation of $\mathrm{BALC}_{\mathrm{JSD}}$ can be found in Appendix~\ref{appendix:balc_derivation}. 
\begin{equation} 
\label{eq3_main}
\begin{split}
\text{BALC}_{\text{JSD}}(\boldsymbol{x}) & = \overbrace{\frac{1}{T}\sum_{i=1}^{T} \left[\infdiv{\boldsymbol{G_{i}(x)}}{\boldsymbol{G'_{i}(x')}} \right]}^{\text{\circled{A} across parameter perturbations}}  - \\ & \overbrace{\infdiv{\frac{1}{T}\sum_{i=1}^{T}\boldsymbol{G_{i}(x)}}{\frac{1}{T}\sum_{i=1}^{T}\boldsymbol{G'_{i}(x')}}}^{\text{\circled{B} across input perturbations}}
\end{split}
\end{equation}

\subsection{Tracked Acquisition Function}

So far, we have discussed active learning frameworks that, similar to those in the literature, quantify the informativeness of an unlabelled instance at a \textit{single} snapshot in time (e.g., at a particular epoch, $\tau$). This static setup, however, faces two limitations. First, it depends on an appropriate choice of epoch for the acquisition of instances, which is non-trivial to identify a priori. For example, an acquisition function can be of little value if calculated when network parameters have yet to be updated sufficiently. Second, the diversity and number of hypotheses obtained via parameter perturbations can be limited by this single-epoch view. This is detrimental given the established benefit of eliminating unsuitable hypotheses at a greater rate ($\downarrow$ version space) \cite{Cohn1994}. 

To overcome such limitations, we propose to \textit{track} this acquisition function over time (e.g., epochs) before deploying it for the acquisition of instances. This is a simple extension to almost any acquisition function. In the process, we become less dependent on the choice of acquisition epoch and are likely to increase the diversity of hypotheses at our disposal. Formally, consider an acquisition function, $\alpha(\boldsymbol{x})$, which would ordinarily be calculated once at epoch, $\tau$, for each instance, $\boldsymbol{x}$. In our formulation, we calculate $\alpha(\boldsymbol{x},t)$ at each epoch, $t \in [1,\tau]$. At epoch, $\tau$, which we refer to as an \textit{acquisition epoch}, the modified acquisition function, $\mathrm{AUTAF(\boldsymbol{x},\alpha)} \in \mathbb{R}$, corresponds to the area under the tracked acquisition function, approximated via the trapezoidal rule.
\begin{equation} 
\label{eq1_main}
\begin{split}
\text{AUTAF}(\boldsymbol{x},\alpha) 
	& \approx \sum_{t=1}^{\tau} \left(\frac{\alpha(\boldsymbol{x},t+\Delta t) + \alpha(\boldsymbol{x},t)}{2}\right) \Delta t
\end{split}
\end{equation}

where $\Delta t$ is the interval between epochs at which acquisition values are calculated.

\subsection{Selective Oracle Questioning}

Up until this point, we have discussed methods that exclusively target the \textit{acquisition} of unlabelled instances. We now direct our attention to the \textit{annotation} of such instances with the goal of minimizing the labelling burden that is placed on an oracle. To do so, we design a framework that, given an acquired unlabelled instance, dynamically determines whether to request a label from an oracle or to pseudo-label that instance instead. This framework, which we refer to as selective oracle questioning in active learning, SoQal, is outlined next.

\begin{figure}[!b]
\centering
\begin{subfigure}{1\columnwidth}
        \centering
        \includegraphics[width=\textwidth]{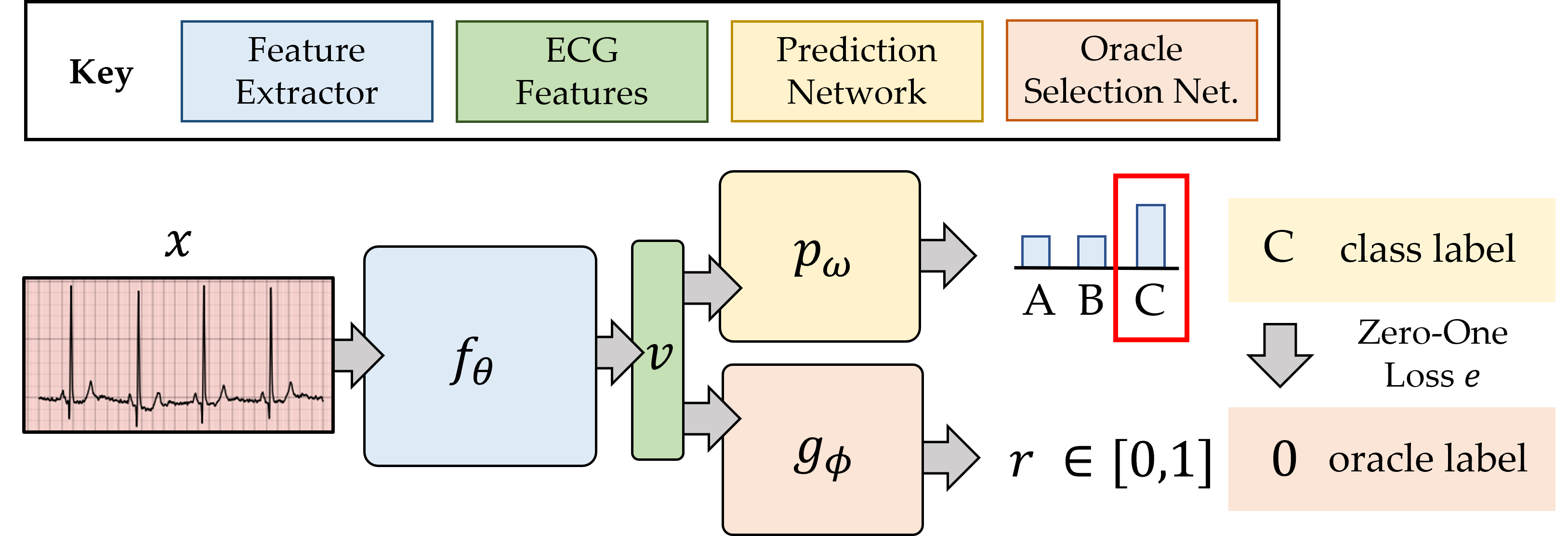}
\end{subfigure}
        \caption{\textbf{Selective oracle questioning framework.} Representation, $\boldsymbol{v}$, of instance, $\boldsymbol{x}$ is passed through $p_{\omega}$ to generate task predictions and $g_{\phi}$ to determine whether a label is requested from an oracle. The zero-one loss of $p_{\omega}$ is the ground-truth label for $r$. High (low) values of $r$ indicate that an instance should be labelled by an oracle (pseudo-labelled by prediction network).}
        \label{fig:network_architecture}
\end{figure}

\paragraph{Oracle selection network} Let us first consider a learner, $g_{\phi}: \boldsymbol{v} \in \mathbb{R}^{E} \rightarrow r \in [0,1]$, parameterized by $\boldsymbol{\phi}$, that maps an $E$-dimensional representation, $\boldsymbol{v}$, to a scalar, $r$, as shown in Fig.~\ref{fig:network_architecture}. We refer to this learner as an oracle selection network.

\paragraph{Learning a proxy for misclassifications} 
The intuition underlying our oracle selection framework is as follows. Given an instance, a network should request a label (i.e., ask for help) from an oracle instead of pseudo-label it if the network is likely to misclassify this instance. This idea operates under the assumption that the network is able to identify whether an instance will be misclassified. While this is possible with \textit{labelled} data, it is non-trivial with \textit{unlabelled} data (due to absence of a ground-truth), the prime focus of active learning. As such, we need a way to identify whether an unlabelled instance would have been misclassified if it were to be pseudo-labelled by the network. In other words, we need a reliable proxy for a misclassification. 

To design this proxy, we exploit the oracle selection network ($g_{\phi}$) and its scalar output, $r$. As we explain next, we learn $\boldsymbol{\phi}$ such that higher values of $r$ correspond to misclassifications. Achieving this requires that $r$ have a supervisory ground-truth label. We choose such a label to be the zero-one loss, $e \in \mathbb{R}$, incurred by the prediction network, $p_{\omega}$, on an instance, $\boldsymbol{x}$. Intuitively, when $e=1$, the network should request a label from an oracle. Note that this ground-truth label, $e$, is only available while training on \textit{labelled} data. We later explain how to exploit $r$ on acquired \textit{unlabelled} instances. As such, while training on labelled data, we optimize \eqref{eq:loss_function} with a categorical cross-entropy loss for the prediction network, $\mathcal{L}_{CPL}$, with ground-truth class labels, $c$, and a weighted binary cross-entropy loss for the oracle selection network, $\mathcal{L}_{OSL}$. In the process, we learn the parameters, $\boldsymbol{\theta}$, $\boldsymbol{\omega}$, and $\boldsymbol{\phi}$ in an end-to-end manner. 
\begin{equation}
\begin{split}
\mathcal{L} = \mathcal{L}_{CPL} + \mathcal{L}_{OSL}
\end{split}
\label{eq:loss_function}
\end{equation}
\begin{equation*}
    \mathcal{L}_{CPL} = - \sum_{i=1}^{B} \log\left (p_{\boldsymbol{\omega}}(y_{i}=c|\boldsymbol{x}_{i},\boldsymbol{\phi}) \right)
\end{equation*}
\begin{equation*}
    \mathcal{L}_{OSL} = - \sum_{i=1}^{B} \beta e_{i}\log g_{\boldsymbol{\phi}}(\boldsymbol{x}_{i}) - (1-e_{i})\log \left(1-g_{\boldsymbol{\phi}}(\boldsymbol{x}_{i}) \right)
\end{equation*}

\textbf{Weighted oracle selection loss.} During the early stages of training on \textit{labelled} data, a network struggles to classify instances correctly. In our framework, this implies that we will encounter more terms with $e=1$ (misclassification) than those with $e=0$ (correct classification). The opposite is true as training progresses and the network becomes more adept at classifying instances. Therefore, regardless of the stage of training (early vs. late), there will be an imbalance in the ground-truth labels, $e$, provided to the oracle selection network. This imbalance sends a strong supervisory signal to $g_{\phi}$ through the oracle selection loss, $\mathcal{L}_{OSL}$, and not accounting for it would lead to systematically higher (lower) $r$ values during the early (late) stages of training. This can reduce the reliability of $r$ as a proxy for misclassifications. As such, we introduce a dynamic weight, $\beta=\frac{\sum \mathbbm{1}({e=0})}{\sum \mathbbm{1}({e=1})}$, where $\mathbbm{1}$ is the indicator function. As training progress, $\beta < 1 \rightarrow \beta > 1$, as the ratio of correctly classified ($e=0$) to misclassified ($e=1$) instances within a mini-batch changes.

\paragraph{Making decisions with proxy} 
\label{sec:making_decisions}

\begin{figure}[!t]
\centering
\begin{subfigure}{0.32\columnwidth}
        \centering
        \includegraphics[width=\textwidth]{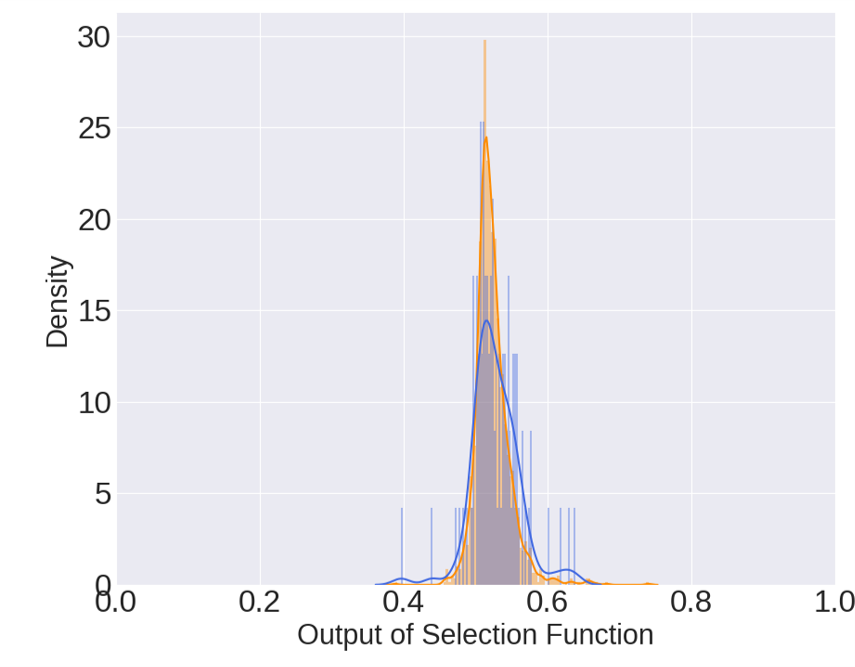}
        \caption{Early stage}
        \label{fig:histogram_separation_early}
\end{subfigure}
\begin{subfigure}{0.32\columnwidth}
        \centering
        \includegraphics[width=\textwidth]{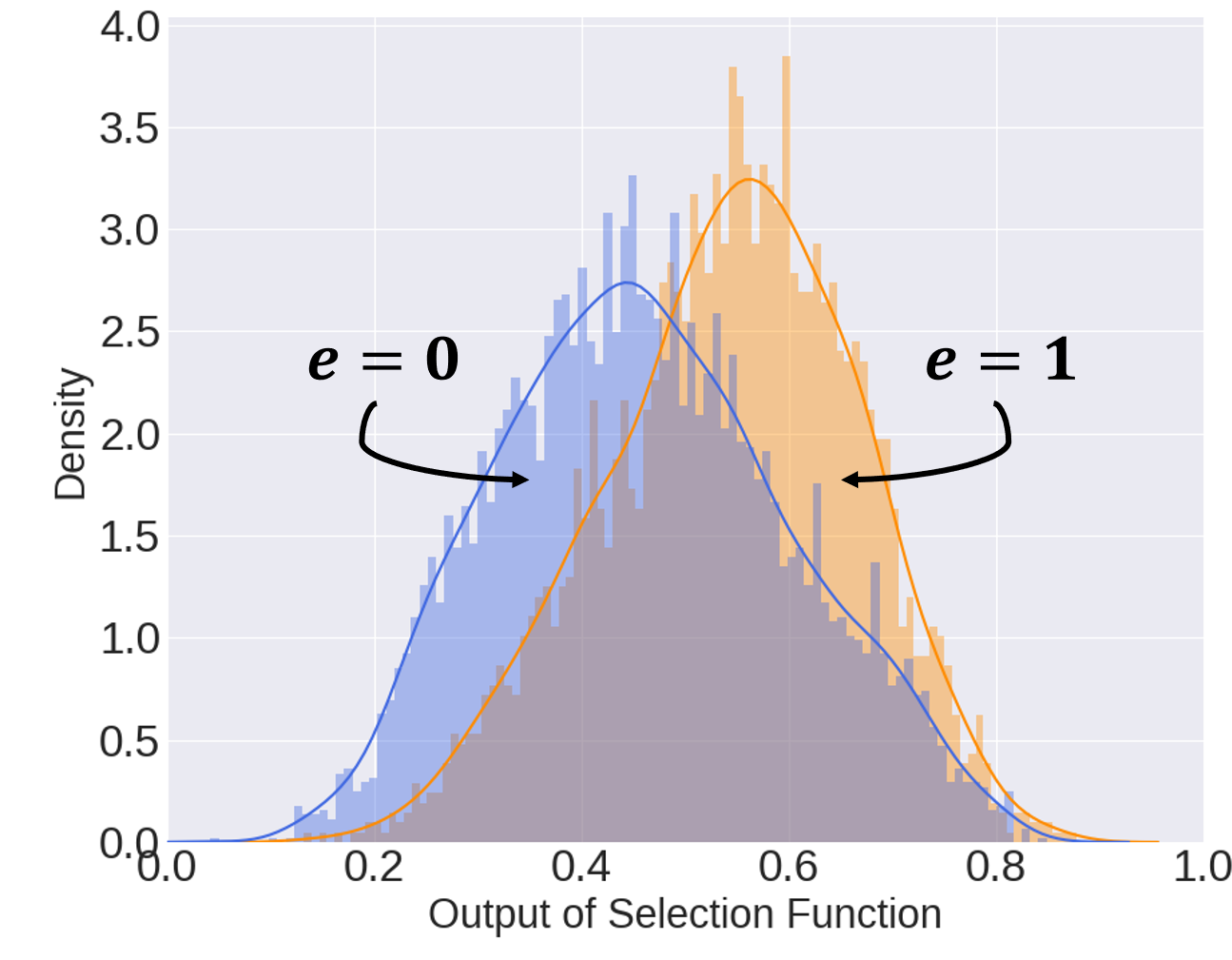}
        \caption{Late stage}
        \label{fig:histogram_separation_late}
\end{subfigure}
\begin{subfigure}{0.31\columnwidth}
        \centering
        \includegraphics[width=\textwidth]{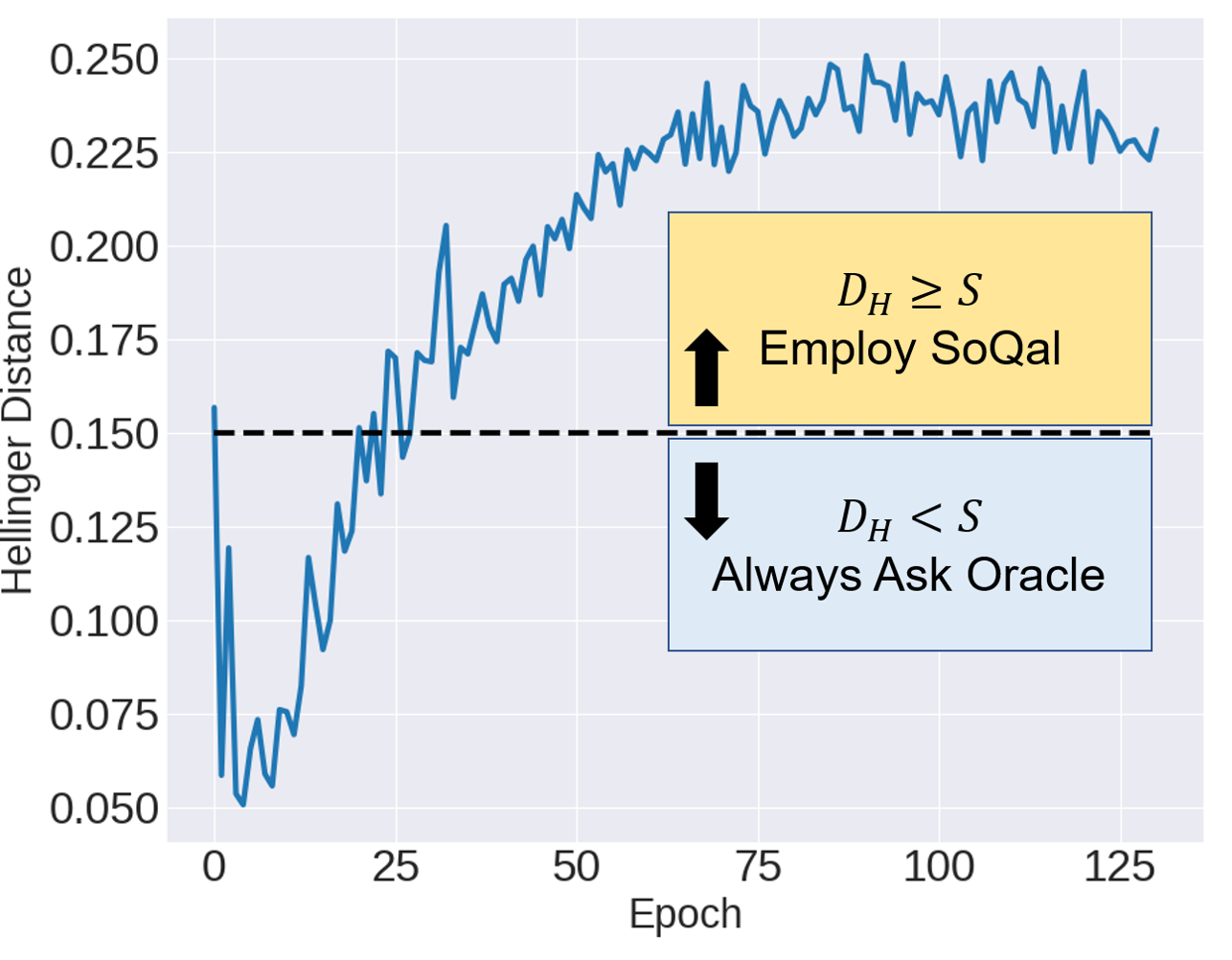}
        \caption{$D_{H}$}
        \label{fig:hellinger_distance}
\end{subfigure}
        \caption{\textbf{Components of selective oracle questioning framework.} Distribution of the outputs, $r$, of $g_{\phi}$ stratified based on the zero-one loss, $e$ during the (a) early and (b) late stages of training. (c) Hellinger distance, $D_{H}$, between the two distributions of $r$ increases during training.}
        % Similar behaviour is observed across datasets (see Appendix~\ref{appendix:density_of_selection_network}).}
        \label{fig:histogram_separation}
\end{figure}

We exploit $r$ for the binary decision of whether to request a label from an oracle or to generate a pseudo-label instead. Since $r \in [0,1]$, one simple rule would be to select an oracle if $r > 0.5$. This threshold, however, may be sub-optimal particularly if the values of $r$ are not calibrated. Therefore, to design a robust rule, we account for the \textit{distribution} of the $r$ values stratified according to the zero-one loss, $e$, on \textit{labelled} data, and the separability of such distributions. We present such empirical stratified distributions in Figs.~\ref{fig:histogram_separation_early} and \ref{fig:histogram_separation_late} during the early and late stages of training, respectively. As expected, we find that correctly-classified instances ($e=0$) tend to have lower $r$ values than those which are misclassified.

We now outline the steps involved in making the binary decision. First, after each training epoch, $t$, we fit two unimodal Gaussian distributions, $\mathcal{N}_{0}(\mu_{0},\sigma_{0}^{2})$ and $\mathcal{N}_{1}(\mu_{1},\sigma_{1}^{2})$ corresponding to $e=0$ and $e=1$, respectively, to the $r$ values (generated on the \textit{labelled} data), similar to those in Fig.~\ref{fig:histogram_separation_late}. Low separability of such distributions would suggest that the oracle selection network has yet to learn to distinguish between correctly-classified and misclassified instances, and is thus generating an unreliable proxy, $r$. We quantify this separability via the Hellinger distance, $D_{H} \in [0,1]$, a metric to which a threshold can be easily applied. In Fig.~\ref{fig:hellinger_distance}, we show that indeed as training progresses, the separability of the two distributions increases, which is suggestive of an increasingly reliable proxy. Based on some user-defined threshold, $S$, low separability (i.e., low proxy reliability) can be defined as $D_{H} < S$. In such an event, we use a conservative strategy that defers labelling to the oracle. This value of $S$ can be altered depending on the relative level of trust placed in the network and oracle. 
% We also explore how $S$ affects performance in Sec.~\ref{sec:soqal_oracle_dependence}. 

\textbf{Pseudo-labelling.} High separability, defined as $D_{H} \geq S$ suggests that the proxy is sufficiently reliable to make decisions. As such, for each acquired \textit{unlabelled} instance, $\boldsymbol{x}_{u} \sim \boldsymbol{X_{U}}$, we obtain $r_{u} = g_{\boldsymbol{\phi}}(\boldsymbol{x}_{u})$ and compare the values of $\mathcal{N}_{0}$ and $\mathcal{N}_{1}$ when evaluated at $r_{u}$. If $\mathcal{N}_{1} > \mathcal{N}_{0}$, then the value of $r$ is too high (indicating a potential network misclassification), and a label is requested from an oracle. Otherwise, the instance is pseudo-labelled via $\argmaxB_{c} p_{\omega}(y=c|\boldsymbol{x})$. The probability of requesting a label from an oracle is denoted by $p(\text{A})$.
\begin{equation}
p(\text{A}) = \begin{dcases}
1, & D_{H} < S \\
1, & \mathcal{N}_{1} > \mathcal{N}_{0} \ \text{and}\ D_{H} \geq S\\
0, & \text{otherwise} \\
\end{dcases}
\label{eq:oracle_strategy}
\end{equation}
where $\mathcal{N}_{1} = \mathcal{N}\left (r=r_{u}|\mu_{1}, \sigma_{1}^{2}, e=1\right)$ and $\mathcal{N}_{0} = \mathcal{N}\left (r=r_{u}|\mu_{0}, \sigma_{0}^{2}, e=0 \right)$. We note that our oracle selection framework is independent of the acquisition of unlabelled instances, making it general enough to be used alongside almost any acquisition function. Pseudo-code for the entire pipeline can be found in Appendix~\ref{appendix:pseudo_code}.

\section{Experimental Design}

\subsection{Datasets}
\label{subsec:datasets}

We conduct our experiments\footnote{Code: https://github.com/danikiyasseh/SoQal} using PyTorch \cite{Paszke2019} on four publicly-available datasets comprising cardiac signals, such as the photoplethysmogram (PPG) and the electrocardiogram (ECG) alongside cardiac arrhythmia labels (abnormalities in the functioning of the heart). \textbf{PhysioNet 2015} ($\mathcal{D}_{1}$) \cite{Clifford2015} consists of PPG data alongside 5 different classes of cardiac arrhythmia. $\mathcal{D}_{2}$ is similar to the first however comprises ECG data. \textbf{PhysioNet 2017} ($\mathcal{D}_{3}$) \cite{Clifford2017} consists of 8,528 single-lead ECG recordings alongside 4 different classes of cardiac arrhythmia. \textbf{Cardiology} ($\mathcal{D}_{4}$) \cite{Hannun2019} consists of single-lead ECG data from 292 patients alongside 12 different classes of cardiac arrhythmia.

In order to evaluate our framework in the limited data regime, we place a fraction, $F \in [0.1, 0.3, 0.5, 0.7, 0.9]$, of the training dataset into a labelled set, $\mathcal{D}_{L}$. and its complement into an unlabelled set, $\boldsymbol{X_{U}}$ (see Table~\ref{table:data_split_main}). Additional details about the datasets and pre-processing steps can be found in Appendix~\ref{appendix:datasets}. 

\begin{table}[!h]
\small
\centering
\begin{tabular}{c | c c c c}
\toprule
 & \multicolumn{2}{c}{\textbf{Train}}  & \multirow{2}{*}{\textbf{Val.}} & \multirow{2}{*}{\textbf{Test}} \\
 Dataset & Labelled & Unlabelled & & \\
\midrule
\multirow{1}{*}{$\mathcal{D}_{1}$}&401&4233&\multirow{1}{*}{1124}&\multirow{1}{*}{1435}\\
\multirow{1}{*}{$\mathcal{D}_{2}$}&401&4233&\multirow{1}{*}{1124}&\multirow{1}{*}{1435}\\
\multirow{1}{*}{$\mathcal{D}_{3}$}&1,776 &16,479&\multirow{1}{*}{4,582}&\multirow{1}{*}{5,824}\\
\multirow{1}{*}{$\mathcal{D}_{4}$}&452 &4,110&\multirow{1}{*}{1,131}&\multirow{1}{*}{1,386}\\
\bottomrule
\end{tabular}
\caption{\textbf{Number of instances in the training, validation, and test sets.} The distribution of labelled instances is shown for a fraction, $F=0.1$, of the original training set. This is also used for the selective oracle questioning experiments. The remaining distributions can be found in Appendix~\ref{appendix:datasets}.}
\label{table:data_split_main}
\end{table}

\subsection{Network Architecture}

Our network architecture involves a lightweight convolutional network which receives a cardiac time-series segment (2500 samples or $\approx$ 5 seconds in duration) as input and returns a probability distribution over cardiac arrhythmias as output. We chose this architecture based on previous research demonstrating its effectiveness in classifying cardiac arrhythmias \cite{Kiyasseh2021CROCS}. Additional details about the network architecture can be found in Appendix~\ref{appendix:implementation_details}.

\subsection{Active Learning Scenarios}
We explore three distinct active learning scenarios characterized by the presence and quality of an oracle. Recall that the motivation behind our framework was to exploit abundant unlabelled and limited labelled data while alleviating (and not necessarily eliminating) the labelling burden placed on an oracle. As such, the target and realistic clinical scenario in which our framework would be deployed is one where an oracle (e.g., physician) is available in some capacity. However, to explore the extreme limits of our approach, and as a stepping stone to the target clinical scenario, we begin our experimentation without an oracle and transition to the scenarios in which an oracle is available.

\textbf{Scenario 1 - Without Oracle.} we assume that a physician is unavailable to provide labels and thus evaluate the performance of our framework \textit{without} an oracle.

\textbf{Scenario 2 - Noise-free Oracle.} we assume that a physician is available and capable of providing accurate labels. 

\textbf{Scenario 3 - Noisy Oracle.} we assume that a physician is either ill-trained or unable to perform a diagnosis due to its difficulty. To simulate this setting, we introduce two types of label noise. We stochastically flip the ground-truth label (unseen by the network) of each unlabelled instance to 1) \textbf{(Random)} any other label randomly, or 2) \textbf{(Nearest Neighbour)} its nearest neighbour, in a smaller dimensional subspace, from a \textit{different} class. Whereas the first form of noise is extreme, the latter is more realistic as it may reflect uncertain physician diagnoses. Furthermore, we simulate noise of different magnitude by injecting it with probability $\gamma=[0.05,0.1,0.2,0.4,0.8]$. 

\subsection{Baselines}

We compare our proposed acquisition functions to the state-of-the-art functions used alongside MCD. These include \textbf{Var Ratio}, \textbf{Entropy}, and \textbf{BALD} \cite{Houlsby2011}, definitions of which can be found in Appendix~\ref{appendix:aq_functions}. We also experiment with several baselines that exhibit varying degrees of oracle dependence. \textbf{$\epsilon$-greedy} \cite{Watkins1989} - a stochastic strategy that we adapt to exponentially decay the reliance of the network on an oracle as a function of the number of acquisition epochs. \textbf{$S$-response} - assumes that high entropy predictions are indicative of instances that the network is unsure of. Therefore, we introduce a threshold, $S_\mathrm{Entropy}$, such that if it is exceeded, an oracle is requested to label the chosen instance (see Appendix~\ref{appendix:implementation_details}). 
% We do not compare our methods to Softmax Response \cite{Geifman2017} and SelectiveNet \cite{Geifman2019}, despite their strong performance for selective classification, as they do not trivially extend to the setting in which labels are unavailable.

\subsection{Hyperparameters}

For all experiments, we chose the number of MC samples $T = 20$ to balance between computational complexity and accuracy of the approximation of the version space. During training, we acquire unlabelled instances at pre-defined epochs (acquisition epochs), $\tau = 5n$, $n \in \mathbb{N}^{+}$. During each acquisition epoch, we acquire $b = 2\%$ of the remaining unlabelled instances. We investigate the effect of such hyperparameters on performance in Appendices~\ref{appendix:effect_of_mcsamples}-\ref{appendix:effect_of_aqinterval}. When experimenting with tracked acquisition functions, we chose the temporal period, $\Delta t=1$, calculating the acquisition function at each epoch of training. Given the increasing trend of $D_{H}$ during training (see Fig.~\ref{fig:hellinger_distance}), we chose $D_{H} \geq S = 0.15$ to balance between the reliability of the proxy and the independence of the network from an oracle. 
% We also explore the sensitivity of SoQal to this choice of $S$ in Sec.~\ref{sec:soqal_oracle_dependence}.

\begin{figure*}[!t]
    \centering
    \begin{subfigure}{0.9\columnwidth}
    \centering
        \includegraphics[width=1\textwidth]{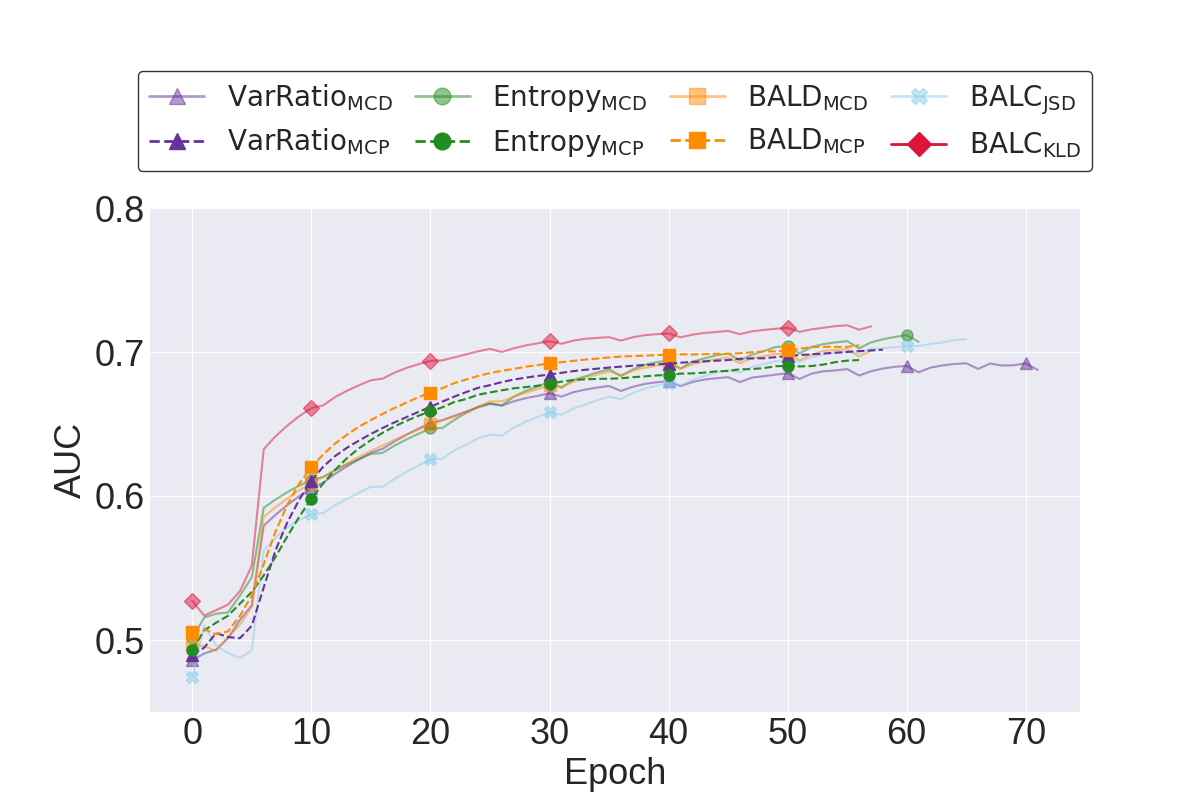}
        \caption{$\mathcal{D}_{2}$ at $F=0.3$}
        \label{fig:validation_AUC_D2}
    \end{subfigure}
    ~
    \begin{subfigure}{0.8\columnwidth}
    \centering
          \includegraphics[width=1\textwidth]{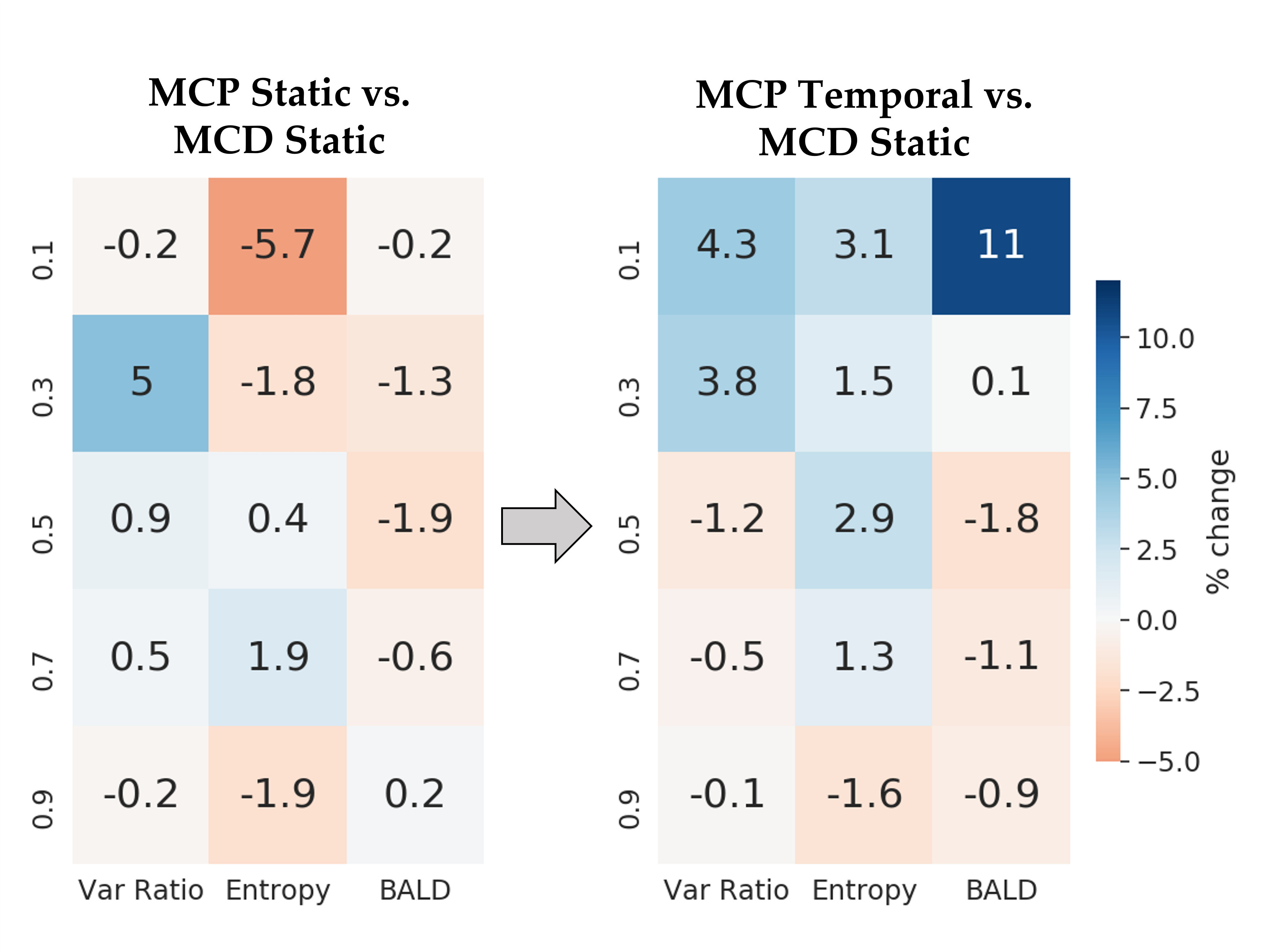}
           \caption{Utility of tracked acquisition functions on $\mathcal{D}_{1}$}
            \label{fig:matrices_D1}
        \end{subfigure}
\caption{\textbf{Performance of active learning frameworks without an oracle.}  \textbf{(a)} Validation AUC on $\mathcal{D}_{2}$ at $F=0.3$. \textbf{(b)} Percent change in AUC when comparing MCP with static and temporal acquisition functions to MCD with static functions on $\mathcal{D}_{1}$. Results are averaged across five seeds. In (a), we show that $\mathrm{BALC}_{\mathrm{KLD}}$ can outperform state-of-the-art acquisition functions. In (b), we show the marginal benefit of temporal acquisition functions.}
\label{fig:results_absence_of_oracle}
\end{figure*}

\section{Experimental Results}

\subsection{Active Learning without Oracle}

We begin by exploring the performance of active learning frameworks without an oracle. In Fig.~\ref{fig:validation_AUC_D2}, we illustrate the validation AUC of a network that is initially exposed to $F=0.3$ of the labelled training data of $\mathcal{D}_{2}$.

We find that a network which exploits $\mathrm{BALC}_{\mathrm{KLD}}$ learns faster than, and outperforms, those which exploit the remaining acquisition functions. For example, $\mathrm{BALC}_{\mathrm{KLD}}$ and $\mathrm{BALD}_{\mathrm{MCD}}$ achieve $\mathrm{AUC} \approx 0.69$ after $20$ and $40$ epochs of training, respectively, reflecting a two-fold increase in learning efficiency. Moreover, $\mathrm{BALC}_{\mathrm{KLD}}$ and $\mathrm{Entropy}_{\mathrm{MCD}}$ achieve a final $\mathrm{AUC} \approx 0.72$ and $\approx 0.70$, respectively. One hypothesis for this improved performance is that $\mathrm{BALC}_{\mathrm{KLD}}$, as a consistency-based active learning framework, acquires unlabelled instances which may still be correctly pseudo-labelled by the network despite the absence of an oracle. This, in turn, facilitates learning. Uncertainty-based acquisition functions, on the other hand, acquire unlabelled instances to which the network is most uncertain. As such, pseudo-labels for these instances are likely to be incorrect. The results for the remaining experiments can be found in Appendix~\ref{appendix:effect_of_perturbations}.

\definecolor{light-gray}{gray}{0.95}
\newcommand{\CC}[1]{\cellcolor{light-gray#1}}

We now transition to quantifying the marginal benefit of incorporating temporal information into the acquisition functions. In Fig.~\ref{fig:matrices_D1}, the panel on the left illustrates the percent change in the AUC when comparing $\mathrm{MCP}$ to $\mathrm{MCD}$ while using \textit{static} acquisition functions. The panel on the right, however, depicts the results after having incorporated temporal information into the acquisition functions used alongside $\mathrm{MCP}$. Hence, the naming $\mathrm{MCP \ Temporal}$. We find that tracked acquisition functions add value when the size of the labelled dataset is small ($\downarrow F$) (red rectangle). For example, transitioning from $\mathrm{MCP \ Static}$ to $\mathrm{MCP \ Temporal}$ when exploiting $\mathrm{BALD}$ at $F=0.1$ improves the $\mathrm{AUC}$ by $11\%$. Similar improvements can be seen in Appendix~\ref{appendix:effect_of_temporal_functions}. We hypothesize that this improvement is due to the increased diversity, and number, of hypotheses available when considering temporal information, thus eliminating unsuitable hypotheses at a greater rate. 

\begin{table}[!b]
\centering
% \small
\resizebox{\columnwidth}{!}{%
\begin{tabular}{c c | c c c}
\toprule
\multirow{2}{*}{Dataset} & \multirow{2}{*}{Ac. Function $\alpha$} & \multicolumn{3}{c}{Oracle Questioning Method} \\ 
 & & $S$-response & $\epsilon$-greedy & SoQal \\
\midrule
\multirow{4}{*}{$\mathcal{D}_{1}$} & $\mathrm{BALD_{MCD}}$ & 49.6 \scriptsize{(3.9}) & 49.1 \scriptsize{(2.8)} & \textbf{62.1 \scriptsize{(2.1)} }\\
& $\mathrm{BALD_{MCP}}$ & 51.7 \scriptsize{(4.3)}  & 50.1 \scriptsize{(4.3)} & \textbf{64.5 \scriptsize{(1.5)} } \\
& $\mathrm{BALC_{KLD}}$ & 54.8 \scriptsize{(3.4)} & 54.8 \scriptsize{(4.2)}  & \textbf{59.8 \scriptsize{(5.5)} } \\ 
& Temp. $\mathrm{BALC_{KLD}}$ & 53.6 \scriptsize{(4.0)} & 52.1 \scriptsize{(5.9)} & \textbf{64.6 \scriptsize{(6.7)}} \\
\hline
\multirow{4}{*}{$\mathcal{D}_{2}$} & $\mathrm{BALD_{MCD}}$ & 58.4 \scriptsize{(4.1)} & 60.9 \scriptsize{(7.1)} & \textbf{70.7 \scriptsize{(3.8)} } \\
& $\mathrm{BALD_{MCP}}$ & 63.8 \scriptsize{(4.3)} & 63.7 \scriptsize{(4.4)} & \textbf{67.7 \scriptsize{(4.2)}} \\
& $\mathrm{BALC_{KLD}}$  & 58.2 \scriptsize{(1.7)} & 64.3 \scriptsize{(3.3)} & \textbf{67.7 \scriptsize{(2.4)}} \\ 
& Temp. $\mathrm{BALC_{KLD}}$ & 61.2 \scriptsize{(5.0)} & 60.5 \scriptsize{(1.9)} & \textbf{64.8 \scriptsize{(5.7)} } \\
\hline
\multirow{4}{*}{$\mathcal{D}_{3}$} & $\mathrm{BALD_{MCD}}$ & 58.8 \scriptsize{(1.3)} & 67.3 \scriptsize{(1.5)} & \textbf{72.1 \scriptsize{(2.5)}} \\
& $\mathrm{BALD_{MCP}}$ & 67.6 \scriptsize{(5.8)} & 66.5 \scriptsize{(2.8)} & \textbf{72.0 \scriptsize{(4.4)}}  \\
& $\mathrm{BALC_{KLD}}$  & 62.9 \scriptsize{(0.4)} & 64.3 \scriptsize{(4.1)} & \textbf{73.1 \scriptsize{(3.3)} } \\ 
& Temp. $\mathrm{BALC_{KLD}}$ & 63.0 \scriptsize{(1.4)} & 65.4 \scriptsize{(1.9)} & \textbf{73.0 \scriptsize{(2.4)}}  \\
\hline
\multirow{4}{*}{$\mathcal{D}_{4}$} & $\mathrm{BALD_{MCD}}$ & 48.9 \scriptsize{(3.0)} & 47.4 \scriptsize{(3.7)} & 46.8 \scriptsize{(2.1)} \\
& $\mathrm{BALD_{MCP}}$ & 50.4 \scriptsize{(2.6)} & 49.2 \scriptsize{(2.4)} & 49.9 \scriptsize{(2.9)} \\
& $\mathrm{BALC_{KLD}}$ & 50.4 \scriptsize{(3.9)} & 47.3 \scriptsize{(1.0)} & 49.5 \scriptsize{(1.2)}  \\ 
& Temp. $\mathrm{BALC_{KLD}}$ & 49.6 \scriptsize{(2.3)} & 49.6 \scriptsize{(2.3)} & 50.3 \scriptsize{(1.0)} \\
\bottomrule
\end{tabular}
}
\caption{\textbf{Test AUC of oracle questioning methods with a noise-free oracle at $F=0.1$.} Mean (standard deviation) is presented across five random seeds for $\mathcal{D}_{1} - \mathcal{D}_{4}$. Bold indicates top-performing method. We show that SoQal outperforms $S$-response and $\epsilon$-greedy across $\mathcal{D}_{1} - \mathcal{D}_{3}$.}
\label{table:effect_of_oracle_strategies}
\end{table}

\subsection{Active Learning with Noise-free Oracle}

Having explored the performance of active learning frameworks without an oracle, we now assume the presence of a noise-free oracle and explore the effect of selective oracle questioning methods. In Table~\ref{table:effect_of_oracle_strategies}, we present the results of these experiments across all datasets at $F=0.1$. 

We find that $\mathrm{SoQal}$ consistently outperforms $S$-response and $\epsilon$-greedy across $\mathcal{D}_{1}$ - $\mathcal{D}_{3}$. For example, when using $\mathrm{BALD}_{\mathrm{MCD}}$ on $\mathcal{D}_{2}$, $\mathrm{SoQal}$ achieves $\mathrm{AUC}=70.7$ whereas $S$-response and $\epsilon$-greedy achieve $\mathrm{AUC}=58.4$ and $60.9$, respectively. Such a finding suggests that $\mathrm{SoQal}$ is well equipped to know \textit{when} a label should be requested from an oracle. This improved performance also coincides with reduced dependence on an oracle (see Appendix~\ref{appendix:soqal_oracle_dependence}). We also hypothesize that the poor performance of all methods on $\mathcal{D}_{4}$ is due to the cold-start problem \cite{Konyushkova2017} where network learning is hindered by the limited availability of initial labelled training data. We support this claim with further experiments in Appendix~\ref{appendix:soqal_oracle_dependence}.

\subsection{Active Learning with Noisy Oracle}

\begin{figure*}[!h]
    \centering
    \begin{subfigure}[h]{0.9\columnwidth}
            \centering
            \includegraphics[width=\columnwidth]{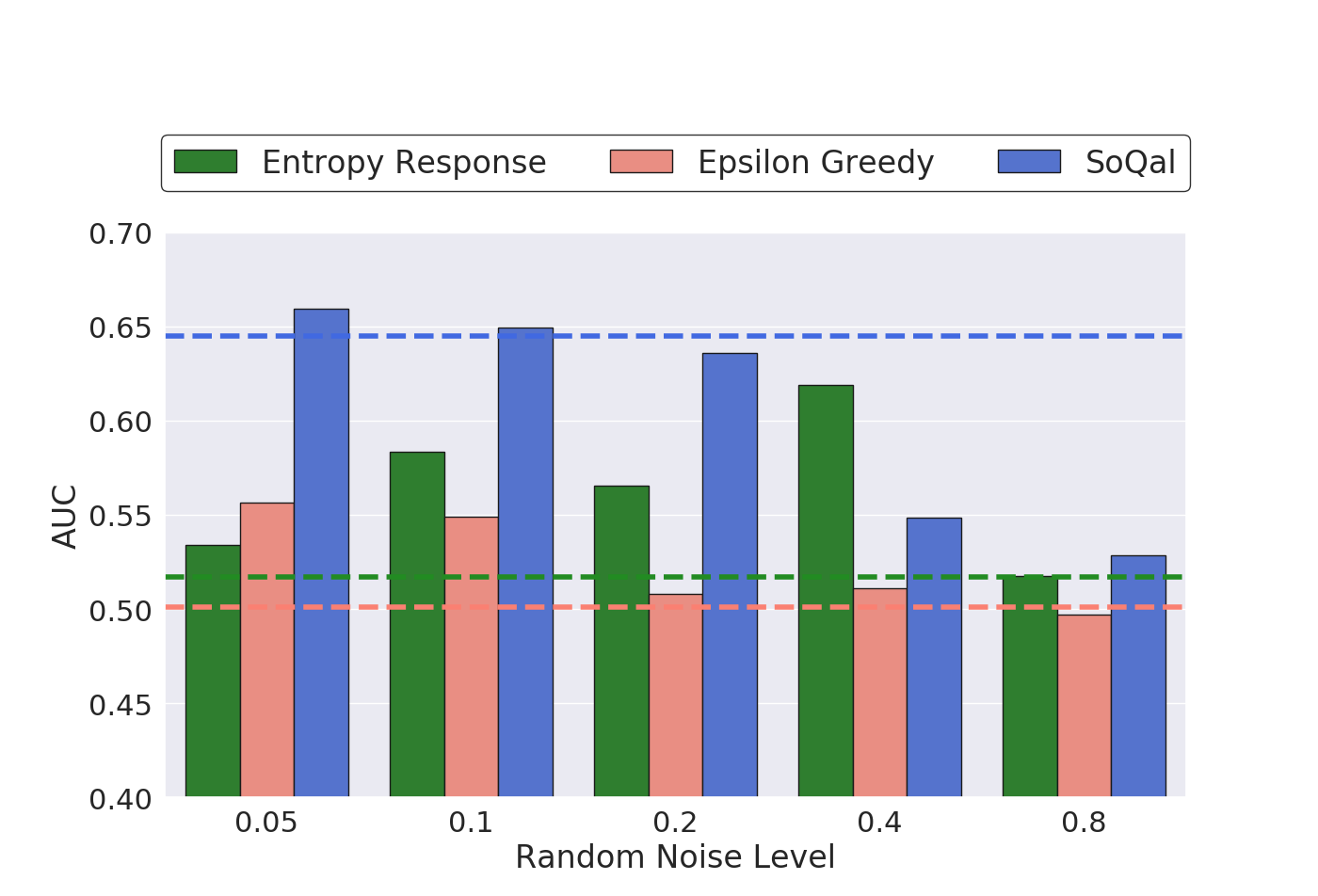}
    \end{subfigure}
    \begin{subfigure}[h]{0.9\columnwidth}
            \centering
            \includegraphics[width=\columnwidth]{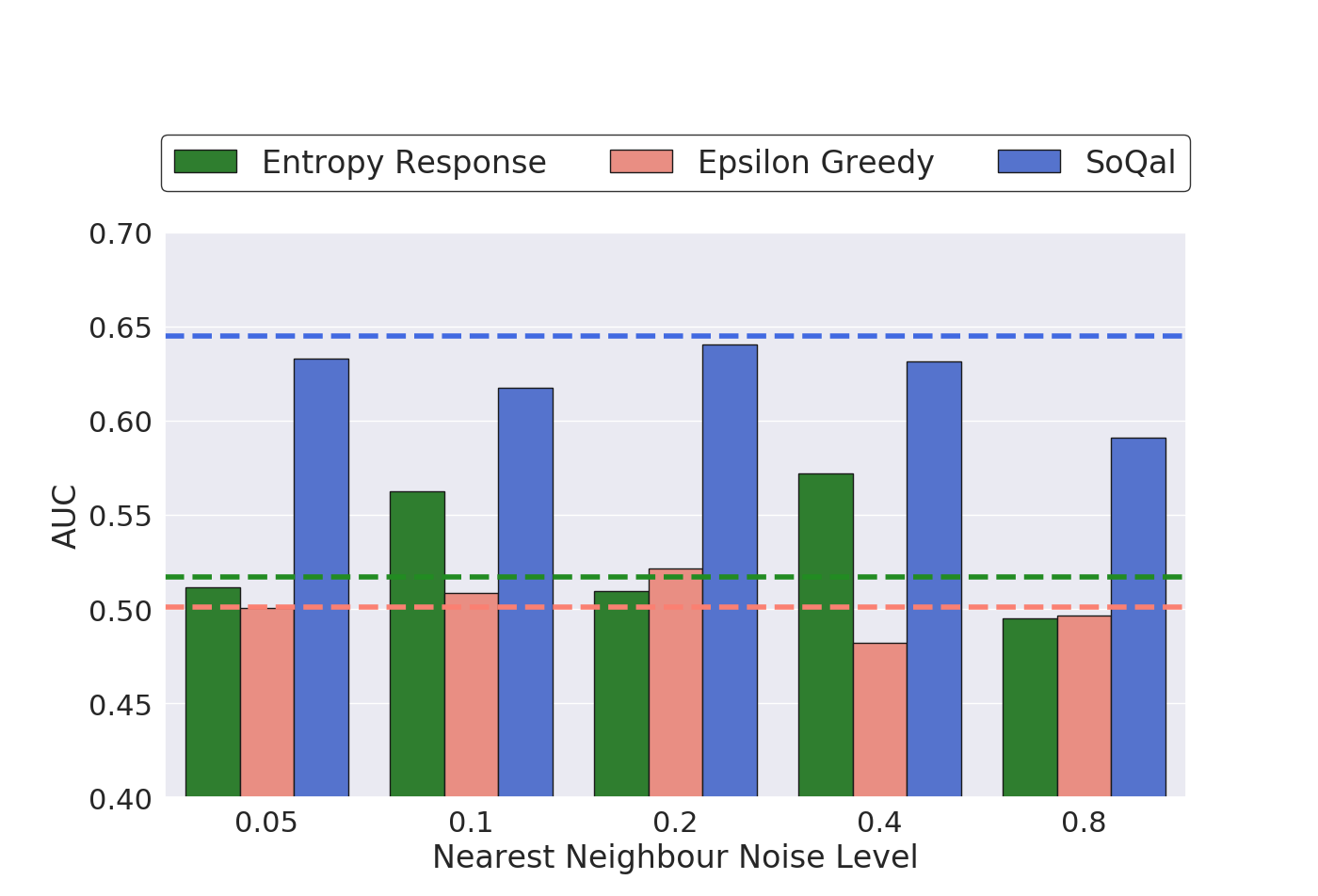}
    \end{subfigure}	
	\caption{\textbf{Test AUC of the oracle questioning methods with (left) random and (right) nearest neighbour label noise on $\mathcal{D}_{1}$ while using $\mathrm{BALD}_{\mathrm{MCP}}$}. Horizontal dashed lines indicate performance of network \textit{without} label noise. $\mathrm{SoQal}$, even with a noisy oracle, outperforms baseline methods with a \textit{noise-free} oracle.}
	\label{fig:average_auc_performance}
\end{figure*}

Building on the findings in the previous section, we now explore the performance of our oracle questioning methods with a \textit{noisy} oracle. In Fig.~\ref{fig:average_auc_performance}, we illustrate the test $\mathrm{AUC}$ on $\mathcal{D}_{1}$ as a function of various types and levels of noise. We also present the performance with a \textit{noise-free} oracle (horizontal dashed lines). 

We find that $\mathrm{SoQal}$ is more robust to a noisy oracle than $\epsilon$-greedy and $S$-response. This is evident by the $\uparrow \mathrm{AUC}$ of the former relative to the latter across different noise types and magnitudes (except at $\gamma=0.4$ random noise). For example, at $5\%$ random noise, $\mathrm{SoQal}$ achieves $\mathrm{AUC} \approx 0.66$ whereas $\epsilon$-greedy and $S$-response achieve $\mathrm{AUC} \approx 0.56$ and $0.53$, respectively. We also find that $\mathrm{SoQal}$, in the presence of noise, continues to outperform the baseline methods in the absence of noise. For example, at $80\%$ nearest neighbour noise, $\mathrm{SoQal}$ achieves $\mathrm{AUC} \approx 0.59$ whereas $\epsilon$-greedy and $S$-response \textit{without} label noise achieve $\mathrm{AUC} \approx 0.50$ and $0.52$, respectively. We arrive at similar conclusions for other datasets and acquisition functions (see Appendix~\ref{appendix:effect_of_label_noise}). One hypothesis for this improved performance is that $\mathrm{SoQal}$, by appropriately deciding when to not request a label from a noisy oracle, avoids an incorrect instance annotation, and thus allows the network to learn well.

\section{Discussion}

In this paper, we proposed a family of active learning frameworks which either perturb instances (Monte Carlo Perturbations) or both instances and network parameters (Bayesian Active Learning by Consistency) and observe changes in the network output distribution. We showed that $\mathrm{BALC}$ can outperform state-of-the-art methods such as $\mathrm{BALD}$. We also found that $\mathrm{MCP}$, when used alongside tracked acquisition functions, can be more favourable than $\mathrm{MCD}$ with static acquisition functions, particularly in low-data regimes. We also proposed $\mathrm{SoQal}$, a framework that dynamically determines whether, for each acquired unlabelled instance, to request a label from an oracle or to pseudo-label it instead. We demonstrated that $\mathrm{SoQal}$ outperforms several baseline methods, even with a noisy oracle, while reducing a network's dependence on the oracle. 

We now outline some of the limitations of our framework. In doing so, we hope to provide guidance on when it should, and should not, be used by machine learning practitioners. First, $\mathrm{SoQal}$ assumed that the zero-one loss incurred by the prediction network (see Fig.~\ref{fig:network_architecture}) is a reliable signal for the oracle selection network to learn from. However, the presence of class label noise hinders the reliability of this signal, introducing an error into the oracle selection process which could manifest as misplaced under- or over-dependence on an oracle. We leave it to future work to investigate the interplay between class label noise and oracle selection. 

$\mathrm{SoQal}$ also assumed that the annotations provided by the oracle, when requested, are \textit{consistently} reliable. However, an oracle (e.g., a physician) is likely to experience fatigue over time and exhibit undesired variability in annotation quality. Such oracle dynamics are not accounted for by our framework yet pose exciting opportunities for the future. Moreover, our framework assumed that, at most, a \textit{single} oracle was available throughout the learning process. However, clinical settings are often characterized by the presence of multiple oracles (e.g., radiologists, cardiologists, oncologists) with different areas and levels of expertise. We hope the community considers incorporating these elements into an active learning framework which would prove quite valuable given the realistic nature of such a scenario.

\section*{Acknowledgements}
We thank the anonymous reviewers for their insightful feedback. We also thank Wadih Al Safi for lending us his voice. David Clifton was supported by the EPSRC under Grants EP/P009824/1and EP/N020774/1, and by the National Institute for Health Research (NIHR) Oxford Biomedical Research Centre (BRC). The views expressed are those of the authors and not necessarily those of the NHS, the NIHR or the Department of Health. Tingting Zhu was supported by the Engineering for Development Research Fellowship provided by the Royal Academy of Engineering.

\bibliography{main}
\bibliographystyle{icml2022}

\newpage
\appendix
\onecolumn

\begin{subappendices}
\section{Acquisition Functions $\alpha$}
\label{appendix:aq_functions}

\begin{equation} \label{eq:variance}
\begin{split}
\mathrm{Variance} \: \mathrm{Ratio} = 1 - \frac{1}{T} \sum_{t=1}^{T} \left(\delta \left(\argmaxB_c p(y=c|x,\omega_{t}) = \hat{c} \right) \right) \\
\hat{c} = \argmaxB_c \left(\argmaxB_c p(y=c|x,\omega_{t}) \ \forall t \in (1,T) \right)
\end{split}
\end{equation}

where $\hat{c}$ is the most common class prediction across the \textit{T} MC samples and $\delta$ is the Dirac delta function that evaluates to 1 if its argument is true, and 0 otherwise. 

\begin{equation} \label{eq:entropy}
\mathrm{Entropy} \: \mathrm{H} = -\sum_{c=1}^{C}p(y=c|x) \log p(y=c|x) 
\end{equation}

\begin{equation} \label{eqB}
\begin{split}
\mathrm{BALD} &= \mathrm{JSD}(p_{1},p_{2},\ldots,p_{T}) \\
	&= \mathrm{H}(p(y|x)) - \mathbb{E}_{p(w|D_{train})} \left[\mathrm{H}(p(y|x,w)) \right]
\end{split}
\end{equation}

where \textit{C} is the number of classes in the task formulation and $p(y=c|x,\omega)$ is the probability assigned by a network parameterised by $\omega$ to a particular class \textit{c} when given an input \textit{x}. 

\end{subappendices}

\begin{subappendices}
\section{Derivation of Monte Carlo Perturbations}
\label{appendix:mcp_derivation}

\begin{equation} \label{eq5}
\begin{split}
\mathrm{BALD_{MCP}}
	&= \mathrm{JSD}(p_{1},p_{2},\ldots,p_{T}) \\
	&= \mathrm{H}(p(y|x)) - \mathbb{E}_{p(x'|D_{train})} \left[\mathrm{H}(p(y|x,x')) \right]
\end{split}
\end{equation}

\begin{equation} \label{eq6}
\begin{split}
\mathrm{H}(p(y|x)) 
	&= \mathrm{H} \left (\int p(y|x') p(x'|x) dx' \right) \\
	&= \mathrm{H} \left (\int p(y|x') q_{\phi}(x'|x) dx' \right) \\
	&\approx \mathrm{H} \left (\frac{1}{T} \sum_{t=1}^{T} p(y|\hat{x'}_{t}) \right)
\end{split}
\end{equation}

where \textit{x'} represents the perturbed input, \textit{T} is the number of Monte Carlo samples, and $\hat{x'}_{t} \sim q_{\phi}(x'|x)$ is a sample from some perturbation generator.

\begin{equation} \label{eq7}
\begin{split}
 \mathbb{E}_{p(x'|D_{train})} \left[\mathrm{H}(p(y|x,x')) \right]
	&= \mathbb{E}_{q_{\phi}(x'|x)} \left[\mathrm{H}(p(y|x,x')) \right] \\
	&\approx \frac{1}{T} \sum_{t=1}^{T} \left[\mathrm{H}(p(y|\hat{x'}_{t})) \right] \\
	& = \frac{1}{T} \sum_{t=1}^{T} \left[-\sum_{c=1}^{C}p(y=c|\hat{x'}_{t}) \log p(y=c|\hat{x'}_{t}) \right]
\end{split}
\end{equation}

\end{subappendices}

\begin{subappendices}
\section{Derivation of Bayesian Active Learning by Consistency}
\label{appendix:balc_derivation}

\begin{equation} \label{eq2}
\begin{split}
\mathrm{BALC_{JSD}} &= \mathbb{E}_{p(\omega|D_{train})} \left[\infdiv{p(y|x,\omega)}{p(y|x',\omega)} \right]  - \infdiv{p(y|x)}{p(y|x')}
\end{split}
\end{equation}

\noindent where \textit{x'} is the perturbed version of the input and $\mathcal{D}_{KL}$ is the Kullback-Leibler divergence. 

\begin{equation} \label{eq3}
\begin{split}
\mathbb{E}_{p(\omega|D_{train})} \left[\infdiv{p(y|x,\omega)}{p(y|x',\omega)} \right] 
	&= \mathbb{E}_{q_{\theta}(\omega)} \left[\infdiv{p(y|x,\omega)}{p(y|x',\omega)} \right] \\
	&\approx \frac{1}{T} \sum_{t=1}^{T} \left[\infdiv{p(y|x,\hat{\omega}_{t}}{p(y|\hat{x},\hat{\omega}_{t})} \right] \\
	&=  \frac{1}{T} \sum_{t=1}^{T} \left[ \sum_{c=1}^{C}  p(y=c|x,\hat{\omega}_{t}) \log \frac {p(y=c|x,\hat{\omega}_{t})} {p(y=c|x',\hat{\omega}_{t})} \right]
\end{split}
\end{equation}

\begin{equation} \label{eq4}
\begin{split}
 \mathcal{D}_{KL}(p(y|x)||p(y|x'))
	&= \infdiv [\bigg]{\int p(y|\omega,x) p(\omega) d\omega}{\int p(y|\omega,x') p(\omega) d\omega} \\
	&= \infdiv [\bigg]{\int p(y|\omega,x) q_{\theta}(\omega) d\omega}{\int p(y|\omega,x') q_{\theta}(\omega) d\omega} \\
	&\approx \infdiv [\bigg]{\frac{1}{T} \sum_{t=1}^{T} p(y| \hat{\omega}_{t},x)}{\frac{1}{T} \sum_{t=1}^{T} p(y| \hat{\omega}_{t}, x')} \\
	&=\frac{1}{C} \sum_{c=1}^{C} \left [\frac{1}{T} \sum_{t=1}^{T} p(y=c| \hat{\omega}_{t},x) \log \frac{\frac{1}{T} \sum_{t=1}^{T} p(y=c| \hat{\omega}_{t},x)}{\frac{1}{T} \sum_{t=1}^{T} p(y=c| \hat{\omega}_{t}, \hat{x})} \right]
\end{split}
\end{equation}

where the integral is approximated by \textit{T} Monte Carlo samples, $\hat{\omega} \sim q_{\theta}(w)$ represents the parameters sampled from the Monte Carlo distribution, and \textit{C} represents the number of classes in the task formulation.

\end{subappendices}

\clearpage

\begin{subappendices}
\section{Datasets}
\renewcommand{\thesubsection}{\Alph{section}.\arabic{subsection}}
\label{appendix:datasets}

\subsection{Data Preprocessing}

Each dataset consists of cardiac time-series waveforms alongside their corresponding cardiac arrhythmia label. Each waveform was split into non-overlapping frames of 2500 samples. 

\textbf{PhysioNet 2015 PPG, $\mathcal{D}_{1}$} \cite{Clifford2015}. This dataset consists of photoplethysmogram (PPG) time-series waveforms sampled at 250Hz and five cardiac arrhythmia labels: Asystole, Extreme Bradycardia, Extreme Tachycardia, Ventricular Tachycardia, and Ventricular Fibrillation. Only patients with a True Positive Alarm are considered. The PPG frames were normalized in amplitude between the values of 0 and 1.

\textbf{PhysioNet 2015 ECG, $\mathcal{D}_{2}$} \cite{Clifford2015}. This dataset consists of electrocardiogram (ECG) time-series waveforms sampled at 250Hz and five cardiac arrhythmia labels: Asystole, Extreme Bradycardia, Extreme Tachycardia, Ventricular Tachycardia, and Ventricular Fibrillation. Only patients with a True Positive Alarm are considered. The ECG frames were normalized in amplitude between the values of 0 and 1.

\textbf{PhysioNet 2017 ECG, $\mathcal{D}_{3}$} \cite{Clifford2017}. This dataset consists of ECG time-series waveforms sampled at 300Hz and four labels: Normal, Atrial Fibrillation, Other, and Noisy. The ECG frames were not normalized.

\textbf{Cardiology ECG, $\mathcal{D}_{4}$} \cite{Hannun2019}. This dataset consists of ECG time-series waveforms sampled at 200Hz and twelve cardiac arrhythmia labels: Atrial Fibrillation, Atrio-ventricular Block, Bigeminy, Ectopic Atrial Rhythm, Idioventricular Rhythm, Junctional Rhythm, Noise, Sinus Rhythm, Supraventricular Tachycardia, Trigeminy, Ventricular Tachycardia, and Wenckebach. Sudden bradycardia cases were excluded from the data as they were not included in the original formulation by the authors. The ECG frames were not normalized. 

\clearpage

\subsection{Data Samples}

All datasets were split into training, validation, and test sets according to patient ID using a 60, 20, 20 configuration. In other words, patients appeared in only one of the sets. Samples in the training set were further split into a labelled and an unlabelled subset, also according to patient ID. In Tables~\ref{table:data_splits} and \ref{table:data_splits_soqal}, we show the number of samples and patients used in each of these sets. 

\subsubsection{Consistency-based Active Learning Experiments}

\begin{table}[h]
\small
\centering
\caption{Sample sizes (number of patients for cardiac datasets) of train/val/test splits. Datasets $\mathcal{D}_{1}$ to $\mathcal{D}_{4}$ are defined in the main manuscript.}
%%\vskip 0.1in
\label{table:data_splits}
\begin{tabular}{c c | c c c c}
%\hhline{======}
\toprule
Dataset&Fraction $F$&Train Labelled&Train Unlabelled&Val&Test\\
%\hhline{======}
\midrule
\multirow{5}{*}{$\mathcal{D}_{1}$}&0.1&401 (18)&4,233 (171)&\multirow{5}{*}{1,124 (47)}&\multirow{5}{*}{1,435 (58)}\\
	&0.3&1,285 (55)&3,349 (134)&&\\
	&0.5&2,187 (92)&2,447 (97)&&\\
	&0.7&3,132 (129)&1,502 (60)&&\\
	&0.9&4,184 (166)&450 (23)&&\\
\midrule
\multirow{5}{*}{$\mathcal{D}_{2}$}&0.1&401 (18)&4,233 (171)&\multirow{5}{*}{1,124 (47)}&\multirow{5}{*}{1,435 (58)}\\
	&0.3&1,285 (55)&3,349 (134)&&\\
	&0.5&2,187 (92)&2,447 (97)&&\\
	&0.7&3,132 (129)&1,502 (60)&&\\
	&0.9&4,184 (166)&450 (23)&&\\
\midrule
\multirow{5}{*}{$\mathcal{D}_{3}$}&0.1&1,776 (545)&16,479 (4,914)&\multirow{5}{*}{4,582 (1,364)}&\multirow{5}{*}{5,824 (1705)}\\
	&0.3&5,399 (1636)&12,856 (3,823)&&\\
	&0.5&9,054 (2727)&9,201 (2,732)&&\\
	&0.7&12,733 (3818)&5,522 (1,641)&&\\
	&0.9&16,365 (4909)&1,890 (550)&&\\
\midrule
\multirow{5}{*}{$\mathcal{D}_{4}$}&0.1&452 (20)&4,110 (181)&\multirow{5}{*}{1,131 (50)}&\multirow{5}{*}{1,386 (62)}\\
	&0.3&1,368 (60)&3,194 (141)&&\\
	&0.5&2,280 (101)&2,282 (100)&&\\
	&0.7&3,200 (140)&1,362 (61)&&\\
	&0.9&4,079 (180)&483 (21)&&\\
% \midrule
% \multirow{5}{*}{$\mathcal{D}_{5}$}&0.1&676 (12)&5,880 (116)&\multirow{5}{*}{1,566 (32)}&\multirow{5}{*}{1,971 (40)}\\
% 	&0.3&1,962 (37)&4,594 (91)&&\\
% 	&0.5&3,152 (62)&3,404 (66)&&\\
% 	&0.7&4,280 (87)&2,276 (41)&&\\
% 	&0.9&5,512 (112)&1,044 (16)&&\\
% \midrule
% \multirow{3}{*}{$\mathcal{D}_{5}$}&0.5& 20,000& 20,000&\multirow{3}{*}{10,000}&\multirow{3}{*}{10,000}\\
% 	&0.7& 28,000 & 12,000 &&\\
% 	&0.9& 36,000 & 4,000 &&\\
%\hhline{======}
\bottomrule
\end{tabular}
\end{table}

\subsubsection{Selective Oracle Questioning Experiments}

\begin{table}[h]
\small
\centering
\caption{Sample sizes (number of patients) of training, validation, and test sets.}
%%\vskip 0.1in
\label{table:data_splits_soqal}
\begin{tabular}{c | c c c c}
%\hhline{=====}\toprule
\toprule
Dataset & Training Labelled & Training Unlabelled & Validation & Test\\
%\hhline{=====}
\midrule
\multirow{1}{*}{$\mathcal{D}_{1}$}&401 (18)&4233 (171)&\multirow{1}{*}{1124 (47)}&\multirow{1}{*}{1435 (58)}\\
\multirow{1}{*}{$\mathcal{D}_{2}$}&401 (18)&4233 (171)&\multirow{1}{*}{1124 (47)}&\multirow{1}{*}{1435 (58)}\\
\multirow{1}{*}{$\mathcal{D}_{3}$}&1,776 (545)&16,479 (4,914)&\multirow{1}{*}{4,582 (1,364)}&\multirow{1}{*}{5,824 (1,705)}\\
\multirow{1}{*}{$\mathcal{D}_{4}$}&452 (20)&4,110 (181)&\multirow{1}{*}{1,131 (50)}&\multirow{1}{*}{1,386 (62)}\\
% \multirow{1}{*}{$\mathcal{D}_{5}$}&676 (12)&5,880 (116)&\multirow{1}{*}{1,566 (32)}&\multirow{1}{*}{1,971 (40)}\\
%\hhline{=====}%
\bottomrule
\end{tabular}
\end{table}

\end{subappendices}

\clearpage

\begin{subappendices}
\section{Implementation Details}
\renewcommand{\thesubsection}{\Alph{section}.\arabic{subsection}}
\label{appendix:implementation_details}

In this section, we outline the network architecture used for all experiments conducted in the main manuscript. We also outline the batchsize and learning rate associated with training on each of the datasets. 

\subsection{Network Architecture}

\begin{table}[h]
\small
\centering
\caption{Network architecture used for time-series experiments. \textit{K}, \textit{C}\textsubscript{in}, and \textit{C}\textsubscript{out} represent the kernel size, number of input channels, and number of output channels, respectively. A stride of 3 was used for Conv1D operators, respectively.}
\label{table:network_architecture}
\begin{subtable}{\textwidth}
\centering
\caption{Network for time-series datasets}
\begin{tabular}{c c c}
\toprule
Layer Number &Layer Components&Kernel Dimension\\
\midrule
	\multirow{5}{*}{1}&Conv 1D & 7 x 1 x 4 (\textit{K} x \textit{C}\textsubscript{in} x \textit{C}\textsubscript{out})\\
										& BatchNorm &\\
										& ReLU& \\
										& MaxPool(2)& \\
										& Dropout(0.1) &\\
	\midrule
	\multirow{5}{*}{2}&Conv 1D & 7 x 4 x 16\\
										& BatchNorm& \\
										& ReLU &\\
										& MaxPool(2) &\\
										& Dropout(0.1)& \\
	\midrule
	\multirow{5}{*}{3}&Conv 1D & 7 x 16 x 32 \\
										& BatchNorm &\\
										& ReLU &\\
										& MaxPool(2) &\\
										& Dropout(0.1) &\\
	\midrule
	\multirow{2}{*}{4}&Linear&320 x 100 \\
										& ReLU &\\
	\midrule
	\multirow{1}{*}{5}&Linear &100 x C (classes) \\
\bottomrule
\end{tabular}
\end{subtable}
\end{table}

% \vspace*{1 cm}

% \begin{subtable}{\textwidth}
% \centering
% \caption{Network for CIFAR10}
% \begin{tabular}{c c c}
% \toprule
% Layer Number &Layer Components&Kernel Dimension\\
% \midrule
% \multirow{3}{*}{1}&Conv 2D & 5 x 3 x 6 \\
% 										& ReLU &\\
% 										& MaxPool(2)& \\
% 	\midrule
% 	\multirow{3}{*}{2}&Conv 2D & 5 x 6 x 16\\
% 										& ReLU &\\
% 										& MaxPool(2) &\\
% 	\midrule
% 	\multirow{3}{*}{3}&Linear&160 x 120 \\
% 										& ReLU &\\
% 										& Dropout(0.1) & \\
% 	\midrule
% 	\multirow{3}{*}{4}&Linear&120 x 84 \\
% 										& ReLU &\\
% 										& Dropout(0.1) & \\
% 	\midrule
% 	\multirow{1}{*}{5}&Linear &84 x C (classes) \\
% \bottomrule
% \end{tabular}
% \end{subtable}
% \end{table}

\clearpage

\subsection{Experiment Details}

\begin{table}[h]
\small
\centering
\caption{Batchsize and learning rates used for training with different datasets. The Adam optimizer was used for all experiments.}
%%\vskip 0.1in
\label{table:batch_size}
\begin{tabular}{c | c c }
%\hhline{===}%
\toprule
Dataset & Batchsize & Learning Rate\\
%\hhline{===}%
\midrule
$\mathcal{D}_{1}$&256&10\textsuperscript{-4}\\
$\mathcal{D}_{2}$&256&10\textsuperscript{-4}\\
$\mathcal{D}_{3}$&256&10\textsuperscript{-4}\\
$\mathcal{D}_{4}$&16&10\textsuperscript{-4}\\
% $\mathcal{D}_{5}$&64&5x10\textsuperscript{-5}\\
%\hhline{===}%
\bottomrule
\end{tabular}
\end{table}

\subsection{Perturbation Details}

When conducting the MCP and BALC experiments, we perturbed each of the time-series frames with additive Gaussian noise, $\epsilon \sim \mathcal{N}(0,\sigma)$ where we chose $\sigma$ based on the specific dataset to avoid introducing too much noise. The details of these perturbations can be found in Table~\ref{table:perturbations}. We applied all perturbations to the input data before normalization. 

\begin{table}[h]
\small
\centering
\caption{Perturbations applied to different datasets during MCP and BALC implementations. \textit{p} represents the probability of applying a particular augmentation method.}
\label{table:perturbations}
\begin{tabular}{c | c }
\toprule
Dataset&Perturbation\\
\midrule
$\mathcal{D}_{1}$&$\epsilon\sim\mathcal{N}(0,100)$\\
$\mathcal{D}_{2}$&$\epsilon\sim\mathcal{N}(0,100)$\\
$\mathcal{D}_{3}$&$\epsilon\sim\mathcal{N}(0,100)$\\
$\mathcal{D}_{4}$&$\epsilon\sim\mathcal{N}(0,100)$\\
% \multirow{3}{*}{$\mathcal{D}_{5}$}&\multicolumn{1}{l}{1) $\mathrm{RandomResizedCrop}$($\mathrm{scale}=(0.8,1.0)$)}\\
%                 &\multicolumn{1}{l}{2) $\mathrm{RandomApply}$($\mathrm{ColorJitter}$(0.8,0.8,0.8,0.2),$p=0.2$)}\\
%                 &\multicolumn{1}{l}{3) $\mathrm{RandomGrayscale}$($p=0.2$)}\\
\bottomrule
\end{tabular}
\end{table}

\clearpage

\subsection{Baseline Implementations}

In this section, we outline our implementation of the baseline methods used in the selective oracle questioning experiments. 

\subsubsection{Entropy Response, ($S$-response)}

This approach is anchored around the idea that network outputs that exhibit high entropy (i.e., close to a uniform distribution) are likely to correspond to instances that the network is uncertain of. Consequently, we exploited this idea to determine whether a label is requested from an oracle or if a pseudo-label should be generated instead. More specifically, we introduced a threshold, $\text{S}_{Entropy}=w \times \text{S}_{Max}$, which is a fraction of the maximum entropy possible for a particular classification problem. As mentioned, $\text{S}_{Max}=\log C$, where $C$ is the number of classes. We chose $w=0.9$ to balance between oracle dependence and pseudo-label accuracy. This value was kept fixed during training. In our implementation, we take the mean of the network outputs as a result of the perturbations, calculate its entropy, and determine whether it exceeds the aforementioned threshold. If it does, then the uncertainty is deemed high and a label is requested from an oracle.

\subsubsection{Epsilon Greedy, ($\epsilon$-greedy)}

This approach is inspired by the reinforcement learning literature and is used to decay the dependence of network on the oracle. More specifically, we define $\epsilon=e^{\frac{-\text{epoch}}{k \times \tau}}$ where epoch represents the training epoch number and $\tau$ is the epoch interval at which acquisitions are performed. $\epsilon$ decays from $1 \rightarrow 0$ as training progresses. We chose $k=\tau=5$ in order to balance between oracle dependence and pseudo-label accuracy. To determine whether a label is requested from an oracle, we generate a random number, $R \sim \mathcal{U}(0,1)$, for a uniform distribution and check whether it is below $\epsilon$. If this is satisfied, then an oracle is requested for a label, and a pseudo-label is generated otherwise. As designed, this approach starts off with 100\% dependence on an oracle and decays towards minimal dependence as training progresses. 

\end{subappendices}

\clearpage

\begin{subappendices}
\section{Pseudo-code}
\label{appendix:pseudo_code}
\renewcommand{\thesubsection}{\Alph{section}.\arabic{subsection}}

We elucidate the entire active learning framework in Algorithms~\ref{alg:BALC} and \ref{algo:soqal} where the coloured lines indicate the components associated with the (optional) tracking of acquisition functions. It is worthwhile to note that our oracle selection framework is independent of the acquisition of unlabelled instances. As such, it is flexible enough to be used alongside other acquisition functions.

\begin{algorithm}[H]
\small
\caption{Bayesian Active Learning by Consistency}
\label{alg:BALC}
\begin{algorithmic}[1]
\Statex {\bfseries Input:} acquisition epochs $\tau$, temporal period $\Delta t$, labelled data $\mathcal{D}_{L}$, unlabelled data $\boldsymbol{X_{U}}$, network parameters $\boldsymbol{\theta}$ $\boldsymbol{\omega}$ $\boldsymbol{\phi}$, MC samples $T$, acquisition fraction $b$
	\While{training}
			\If{epoch in $\Delta t$}
				\For{$\boldsymbol{x} \sim X_{U}$}
					\State $\boldsymbol{x'} = \boldsymbol{x} + \boldsymbol{\epsilon}, \quad \epsilon\sim\mathcal{N}(0,\sigma^2)$ 
					\For{$t$-th MC sample in \textit{T}}
						\State obtain $ p(y|\boldsymbol{x}, \boldsymbol{\theta}_{t}) $ \Comment{ original input}
						\State obtain $ p(y|\boldsymbol{x'}, \boldsymbol{\theta}_{t}) $ \Comment{ perturbed input}
					\EndFor
					\State calculate $\alpha(\boldsymbol{x})$ using (\ref{eq3_main}) or (\ref{eq4_main})
					\State \textcolor{royalblue}{$\alpha(\boldsymbol{x},t) = \alpha$}
				\EndFor
			\EndIf
			\If{epoch in $\tau$}\Comment{acquire unlabelled instances}
				\State \textcolor{royalblue}{calculate $\alpha$ using (\ref{eq1_main}})
				\State SortDescending($\alpha$)
				\State $\boldsymbol{x_{b}} \subset \boldsymbol{X_{U}}$
				\State $y_{b} = \text{SoQal}(\boldsymbol{x_{b}})$  \Comment{ selective oracle questioning}
				\State $\boldsymbol{X_{U}} \in (\boldsymbol{X_{U}}\setminus (\boldsymbol{x_{b}},y_{b}))$
				\State $\mathcal{D}_{L} \in (\mathcal{D}_{L}\cup (\boldsymbol{x_{b}},y_{b}))$
			\EndIf
	\EndWhile
\end{algorithmic}
\end{algorithm}

\begin{algorithm}[H]
\small
\caption{SoQal}
\label{algo:soqal}
\begin{algorithmic}[1]
\Statex {\bfseries Input:} unlabelled instances $\boldsymbol{x_{b}}$, Hellinger distance $D_{H}$, Hellinger threshold $S$
		\For{$\boldsymbol{x}_{u} \sim \boldsymbol{x_{b}}$}
			\State $r_{u} = g_{\phi}(\boldsymbol{x}_{u})$
			\If{$D_{H} > S$}
				 \State calculate $p(\mathrm{A})$ from (\ref{eq:oracle_strategy})
				 \If{$p(\mathrm{A}) = 1$}
				 	\State $y_{b} \subset \boldsymbol{Y_{U}}$ \Comment{request label from physician}
				\Else
					\State $y_{b} = \underset{c} {\mathrm{argmax}} \ p_{\omega}(y=c|\boldsymbol{x}_{u})$ \Comment{pseudo-label}
				\EndIf
			\EndIf
		\EndFor
\end{algorithmic}
\end{algorithm}

\end{subappendices}

\clearpage

\begin{subappendices}
\section{Validation Set AUC with Non-Temporal Acquisition Functions in the Absence of Oracle }
\label{appendix:effect_of_perturbations}

\renewcommand{\thesubsection}{\Alph{section}.\arabic{subsection}}

\begin{figure}[!h]
        \centering
        \begin{subfigure}[h]{\textwidth}
        \centering
        \includegraphics[width=0.7\textwidth]{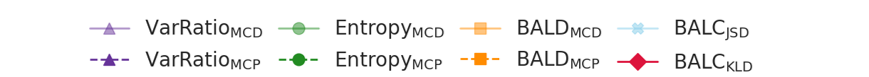}
        \end{subfigure}
        \begin{subfigure}[h]{0.30\textwidth}
        \centering
        \includegraphics[width=\textwidth]{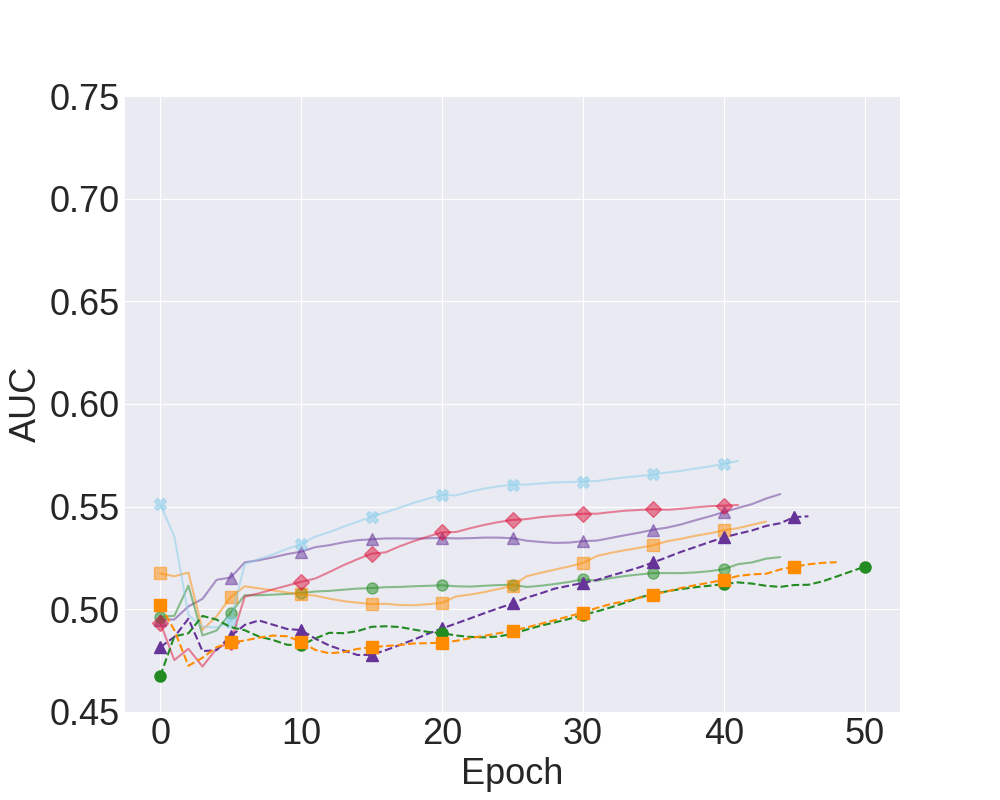}
        \caption{$\beta=0.1$}
        \end{subfigure}
         \begin{subfigure}[h]{0.30\textwidth}
        \includegraphics[width=\textwidth]{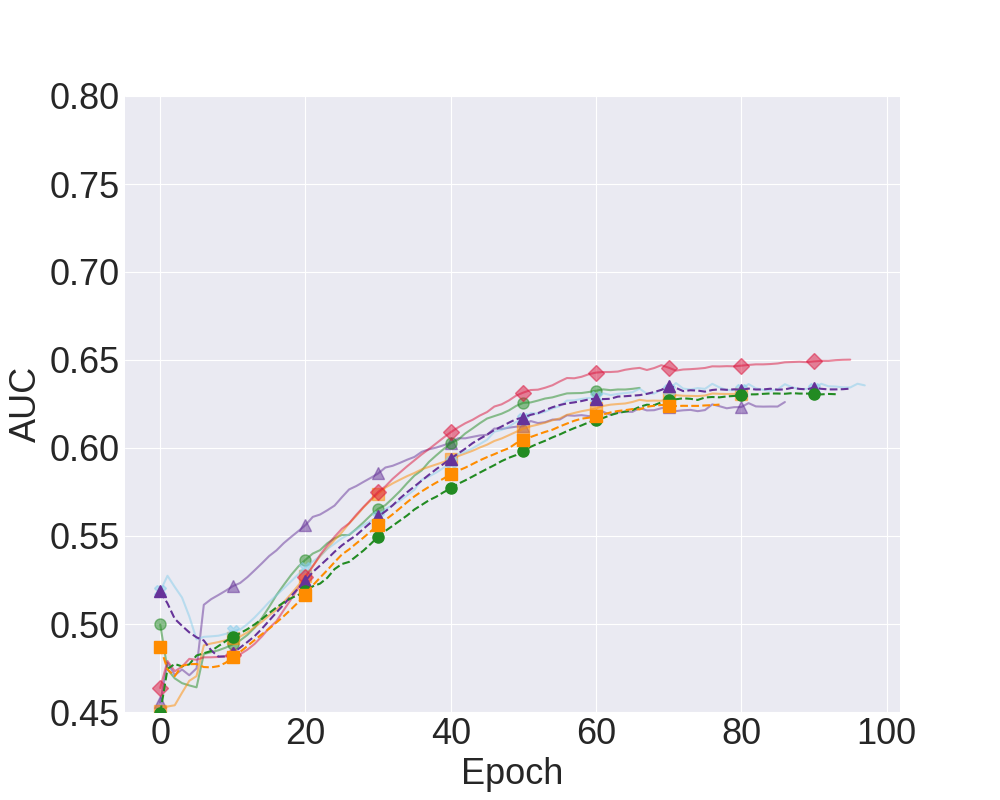}
         \caption{$\beta$ = 0.3}
        \end{subfigure}
         \begin{subfigure}[h]{0.30\textwidth}
        \includegraphics[width=\textwidth]{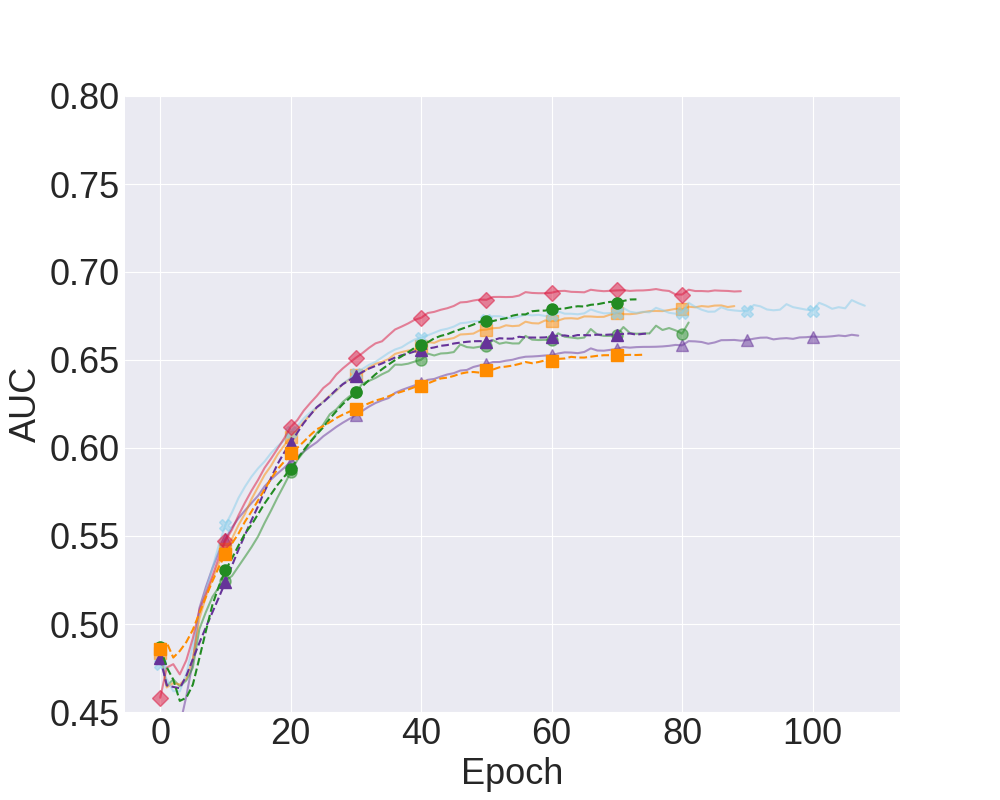}
         \caption{$\beta$ = 0.5}
        \end{subfigure}
         \begin{subfigure}[h]{0.30\textwidth}
        \includegraphics[width=\textwidth]{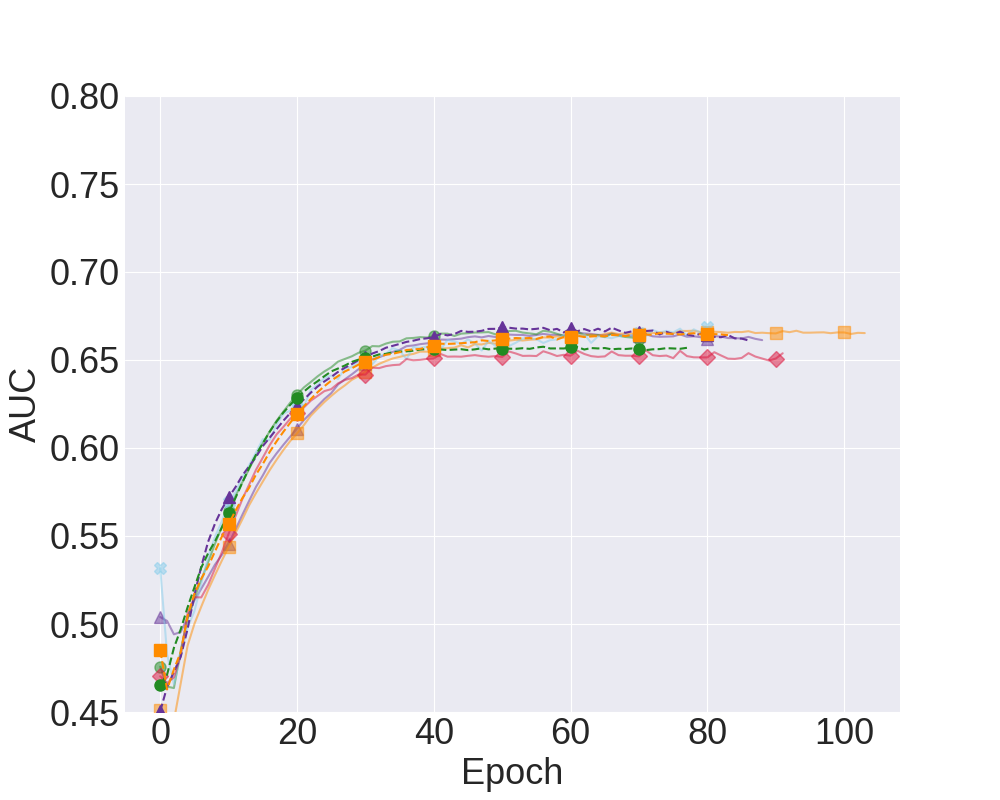}
         \caption{$\beta$ = 0.7}
        \end{subfigure}
        \begin{subfigure}[h]{0.30\textwidth}
        \includegraphics[width=\textwidth]{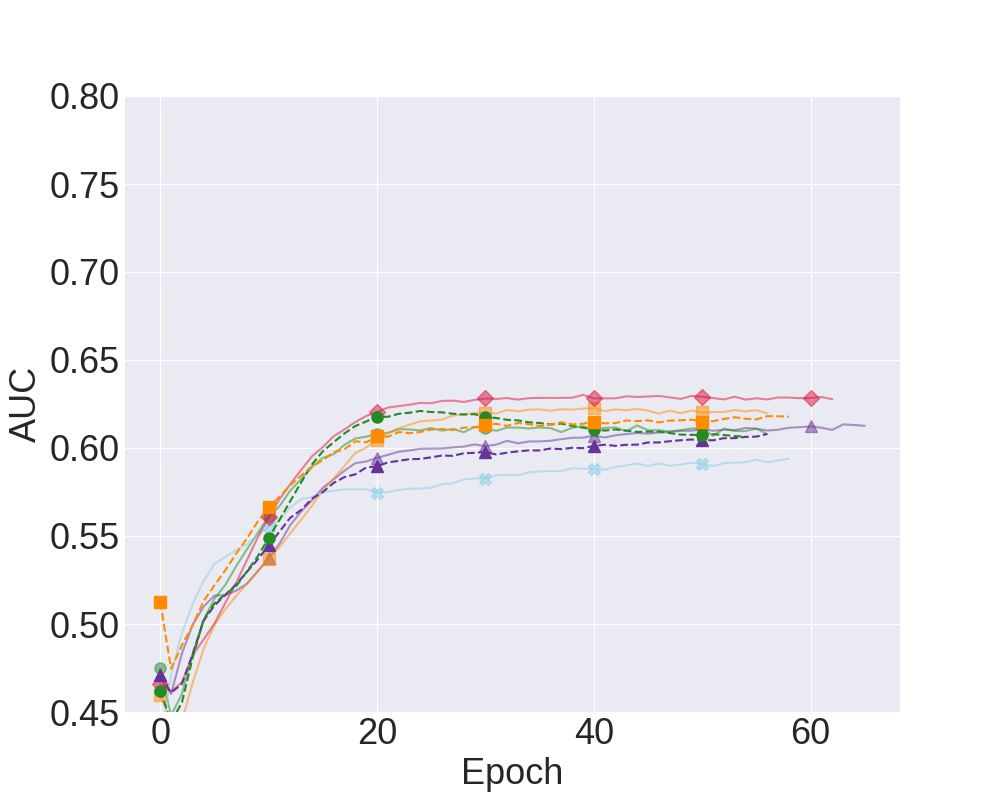}
        \caption{$\beta$ = 0.9}
        \end{subfigure}
        \caption{Mean validation set AUC for the various methodologies and acquisition functions on $\mathcal{D}_{1}$ at increasing fraction levels $\beta = (0.1, 0.3, 0.5, 0.7, 0.9)$. The no-oracle cold-start problem is observed at $\beta=0.1$ where active learning approaches fail due to few available labelled training instances. Clear benefits of our methods can be seen at $\beta=0.5, 0.7$. Results are averaged across 5 seeds.}
\label{fig:impact_of_perturbations_physionet_ppg}
\end{figure}

\begin{figure}[!h]
        \centering
        \begin{subfigure}[h]{\textwidth}
        \centering
        \includegraphics[width=0.7\textwidth]{impact_of_consistency_legend_resized.png}
        \end{subfigure}
        \begin{subfigure}[h]{0.30\textwidth}
        \centering
        \includegraphics[width=\textwidth]{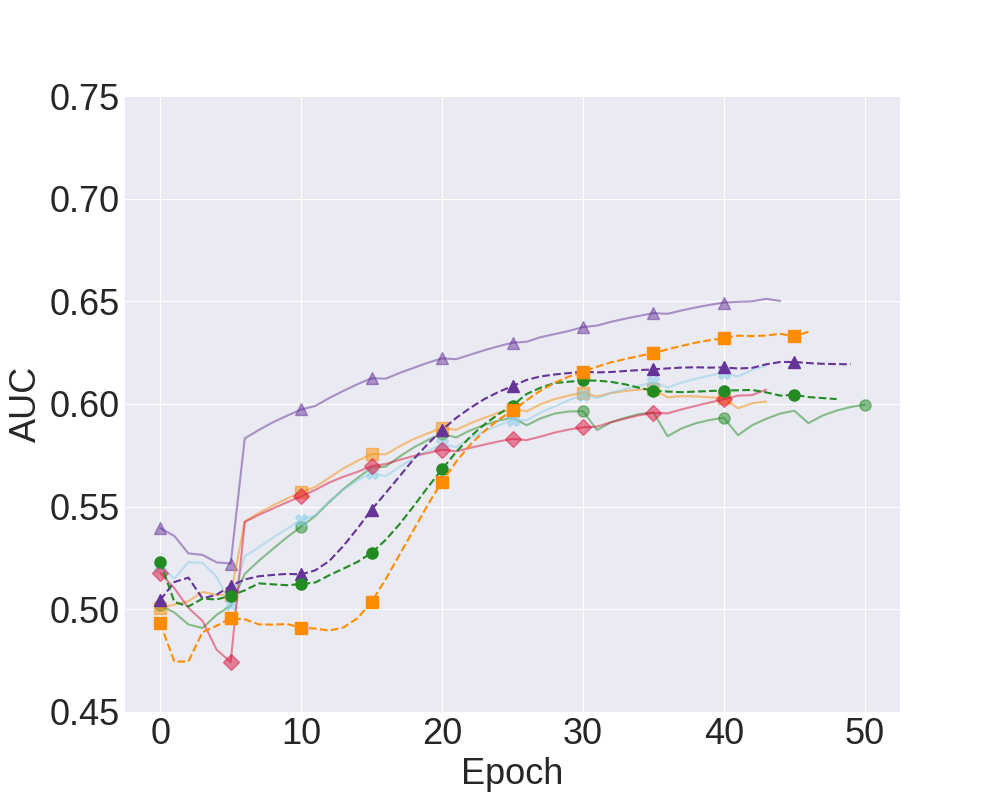}
        \caption{$\beta$ = 0.1}
        \end{subfigure}       
        \begin{subfigure}[h]{0.30\textwidth}
        \centering
        \includegraphics[width=\textwidth]{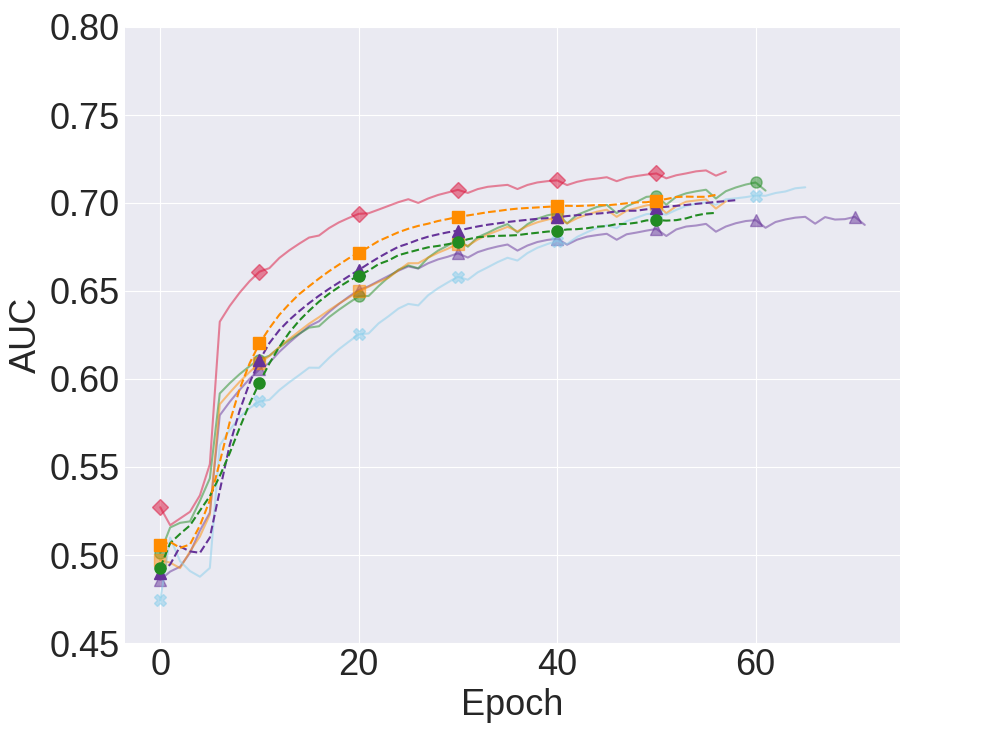}
                        \caption{$\beta$ = 0.3}
        \end{subfigure}
                \begin{subfigure}[h]{0.30\textwidth}
        \centering
        \includegraphics[width=\textwidth]{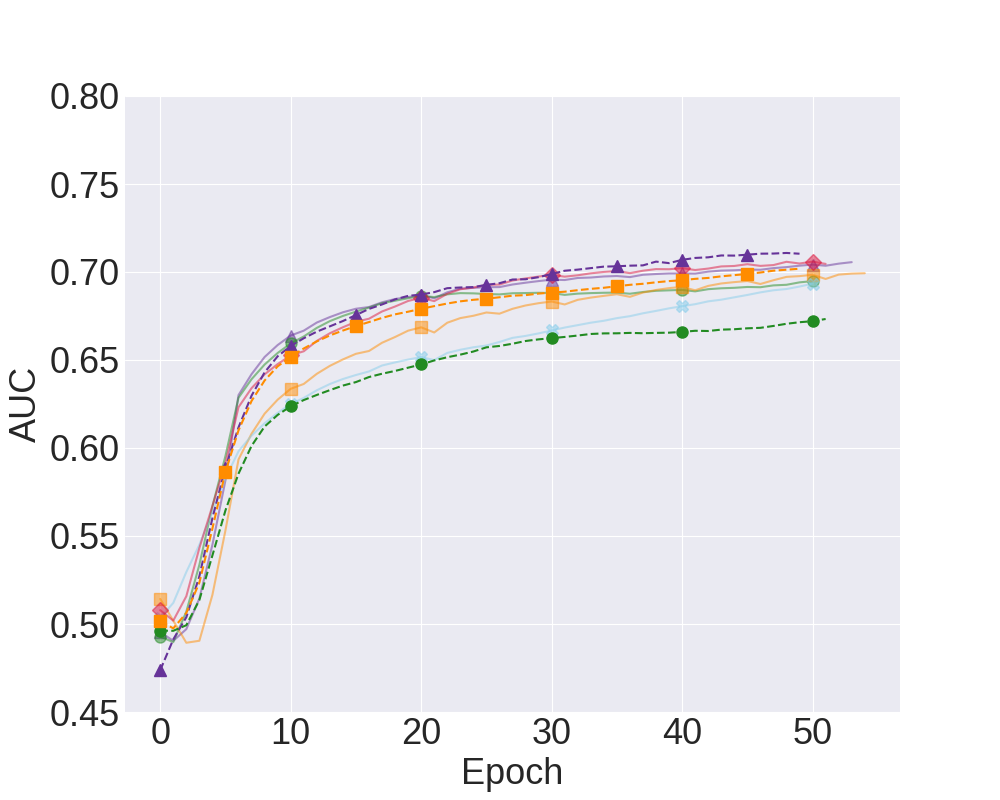}
                        \caption{$\beta$ = 0.5}
        \end{subfigure}
                \begin{subfigure}[h]{0.30\textwidth}
        \centering
        \includegraphics[width=\textwidth]{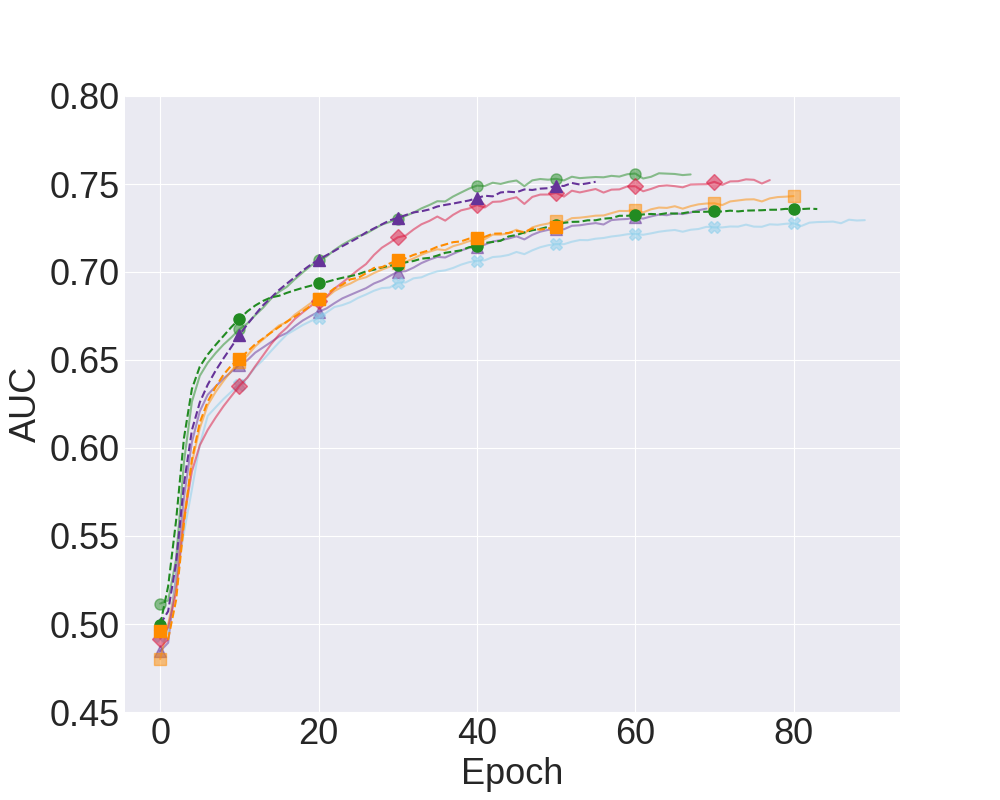}
                        \caption{$\beta$ = 0.7}
        \end{subfigure}
                \begin{subfigure}[h]{0.30\textwidth}
        \centering
        \includegraphics[width=\textwidth]{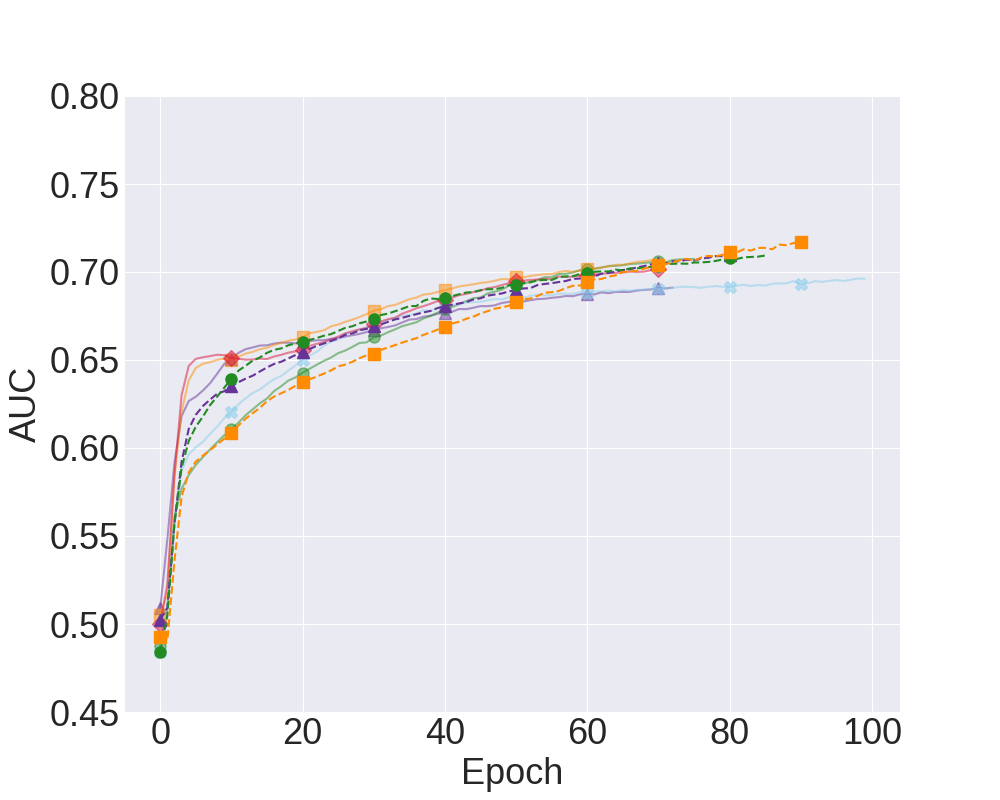}
        \caption{$\beta$ = 0.9}
        \end{subfigure}
        \caption{Mean validation set AUC for the various methodologies and acquisition functions on $\mathcal{D}_{2}$ at increasing fraction levels $\beta = (0.1, 0.3, 0.5, 0.7, 0.9)$. Our methods include MCP and BALC methods. The no-oracle cold-start problem is observed at $\beta=0.1$ where active learning approaches fail due to few available labelled training instances. However, our approaches outperform all others at $\beta=0.3,0.5,0.7,0.9$. Results are averaged across 5 seeds.}
\label{fig:impact_of_perturbations_physionet_ecg}
\end{figure}

\begin{figure}[!h]
        \centering
        \begin{subfigure}[h]{\textwidth}
        \centering
        \includegraphics[width=0.7\textwidth]{impact_of_consistency_legend_resized.png}
        \end{subfigure}
        \begin{subfigure}[h]{0.30\textwidth}
        \centering
        \includegraphics[width=\textwidth]{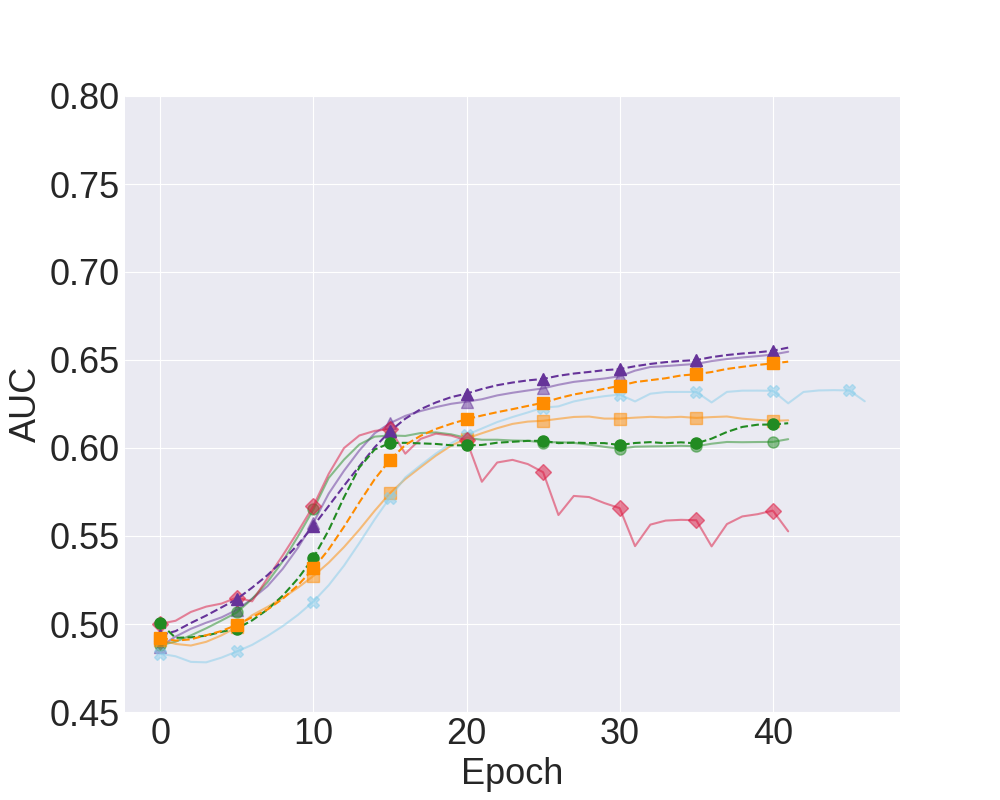}
        \caption{$\beta$ = 0.1}
        \end{subfigure}   
                        \begin{subfigure}[h]{0.30\textwidth}
        \centering
        \includegraphics[width=\textwidth]{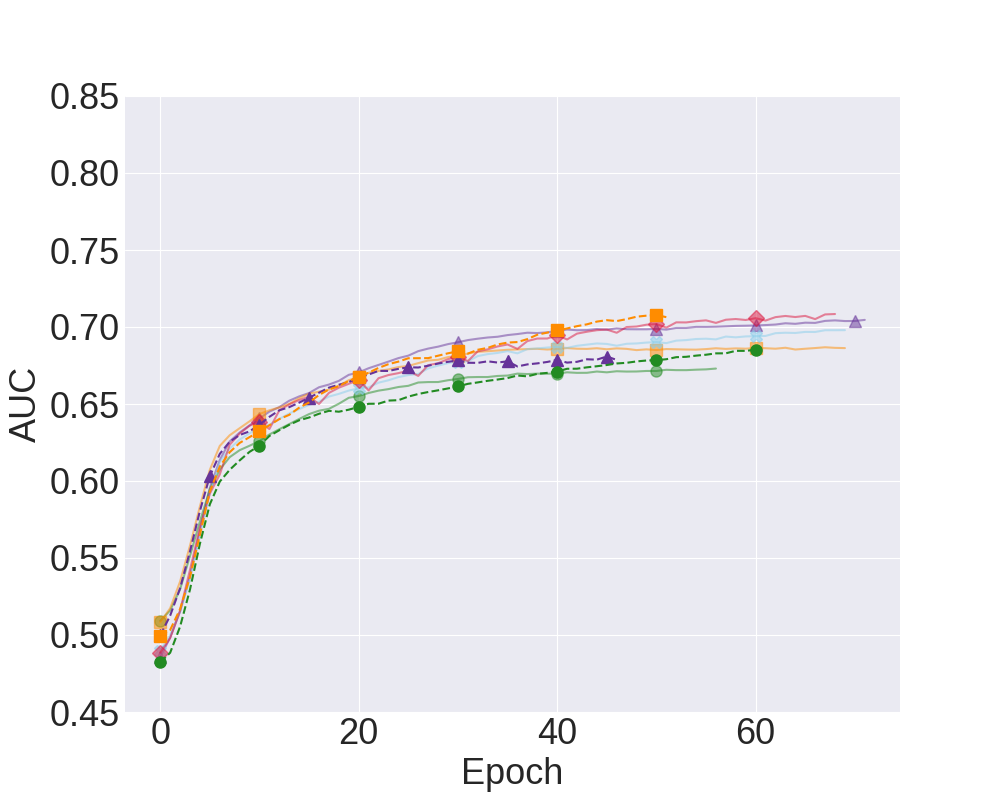}
                                \caption{$\beta$ = 0.3}
        \end{subfigure}
                        \begin{subfigure}[h]{0.30\textwidth}
        \centering
        \includegraphics[width=\textwidth]{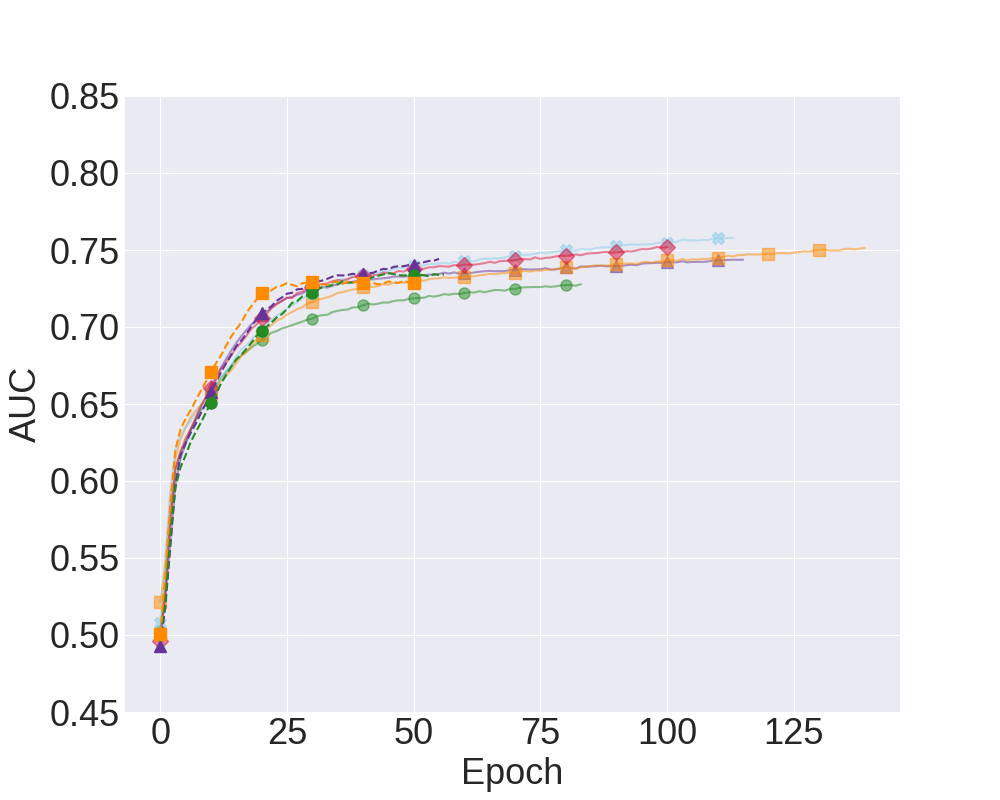}
                                \caption{$\beta$ = 0.5}
        \end{subfigure}
                        \begin{subfigure}[h]{0.30\textwidth}
        \centering
        \includegraphics[width=\textwidth]{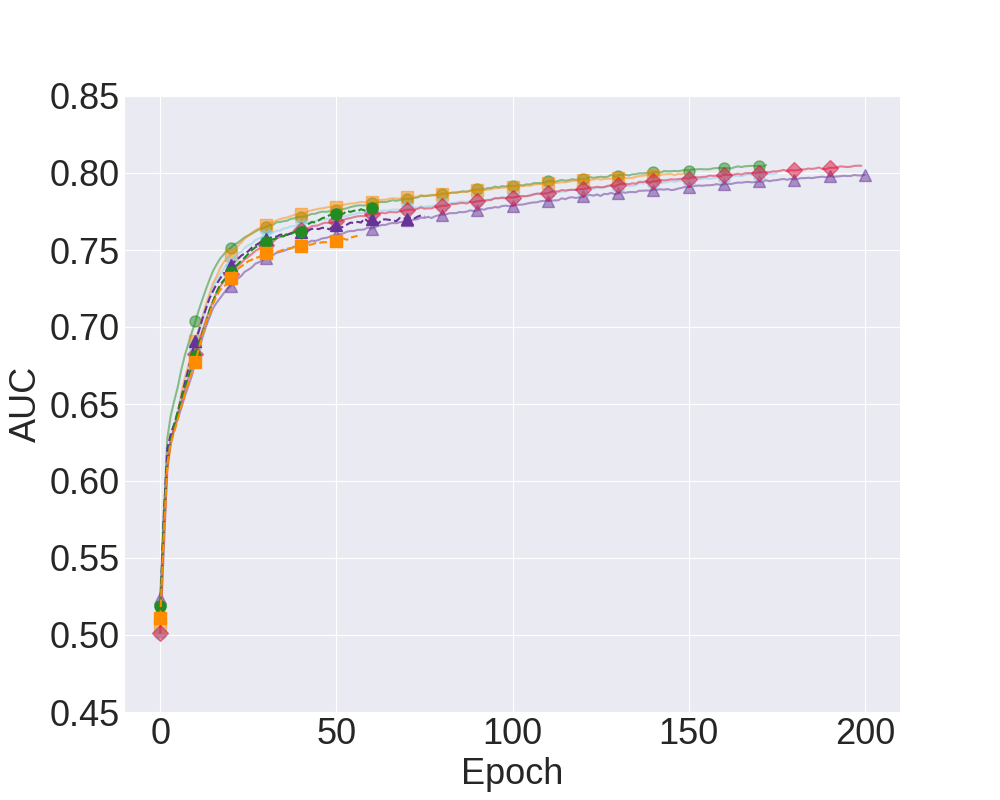}
                                \caption{$\beta$ = 0.7}
        \end{subfigure}
                        \begin{subfigure}[h]{0.30\textwidth}
        \centering
        \includegraphics[width=\textwidth]{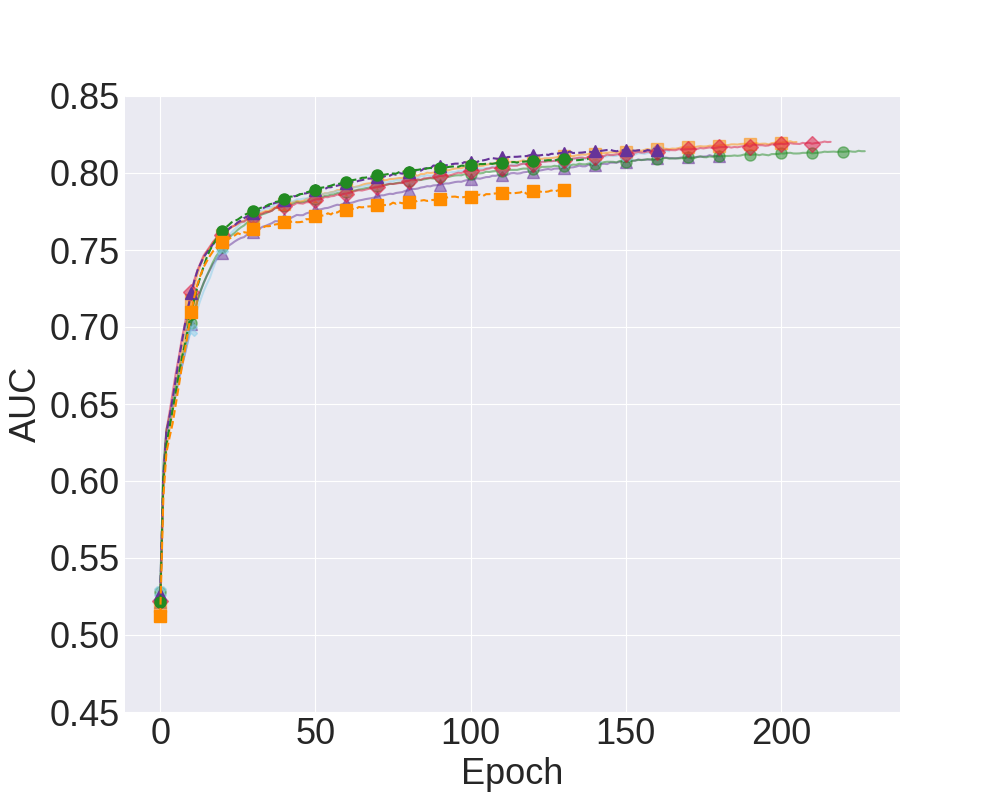}
        \caption{$\beta$ = 0.9}
        \end{subfigure}
        \caption{Mean validation set AUC for the various methodologies and acquisition functions on $\mathcal{D}_{3}$ at increasing fraction levels $\beta = (0.1, 0.3, 0.5, 0.7, 0.9)$. Our methods include MCP and BALC methods. The no-oracle cold-start problem is observed at $\beta=0.1$ where active learning approaches fail due to few available labelled training instances. Most methods perform on par with the no active learning strategy for this particular dataset. Results are averaged across 5 seeds.}
\label{fig:impact_of_perturbations_physionet2017_ecg}
\end{figure}

\begin{figure}[!h]
        \centering
        \begin{subfigure}[h]{\textwidth}
        \centering
        \includegraphics[width=0.7\textwidth]{impact_of_consistency_legend_resized.png}
        \end{subfigure}
        \begin{subfigure}[h]{0.30\textwidth}
        \centering
        \includegraphics[width=\textwidth]{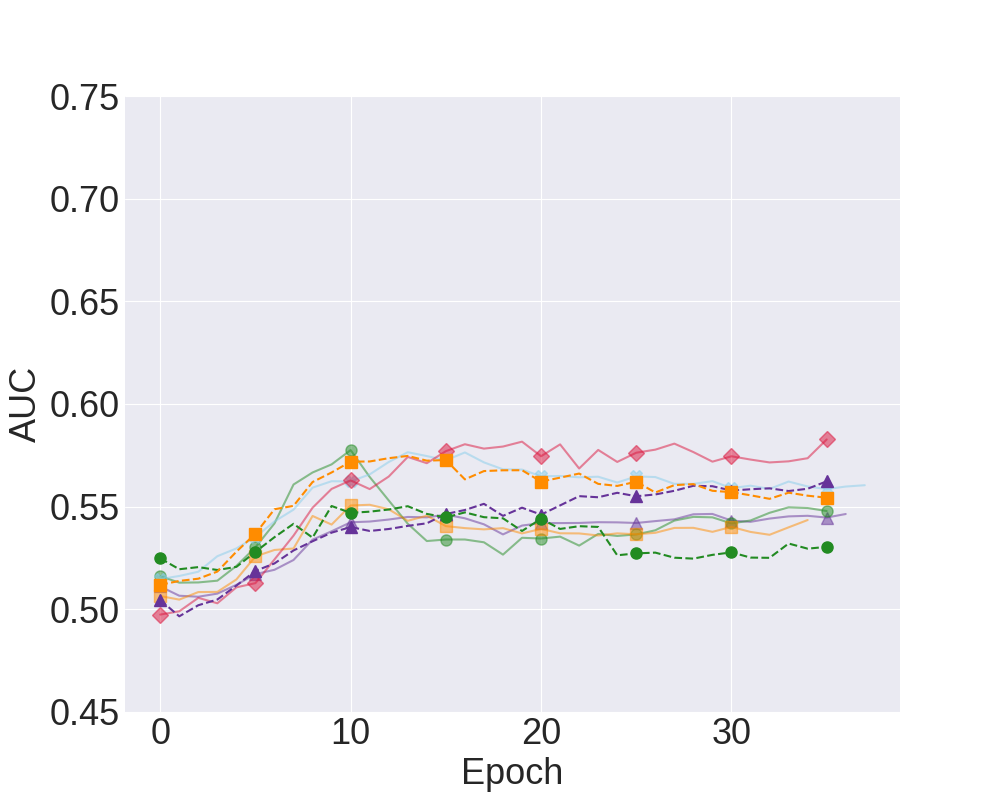}
                                        \caption{$\beta$ = 0.1}
        \end{subfigure}
                                \begin{subfigure}[h]{0.30\textwidth}
        \centering
        \includegraphics[width=\textwidth]{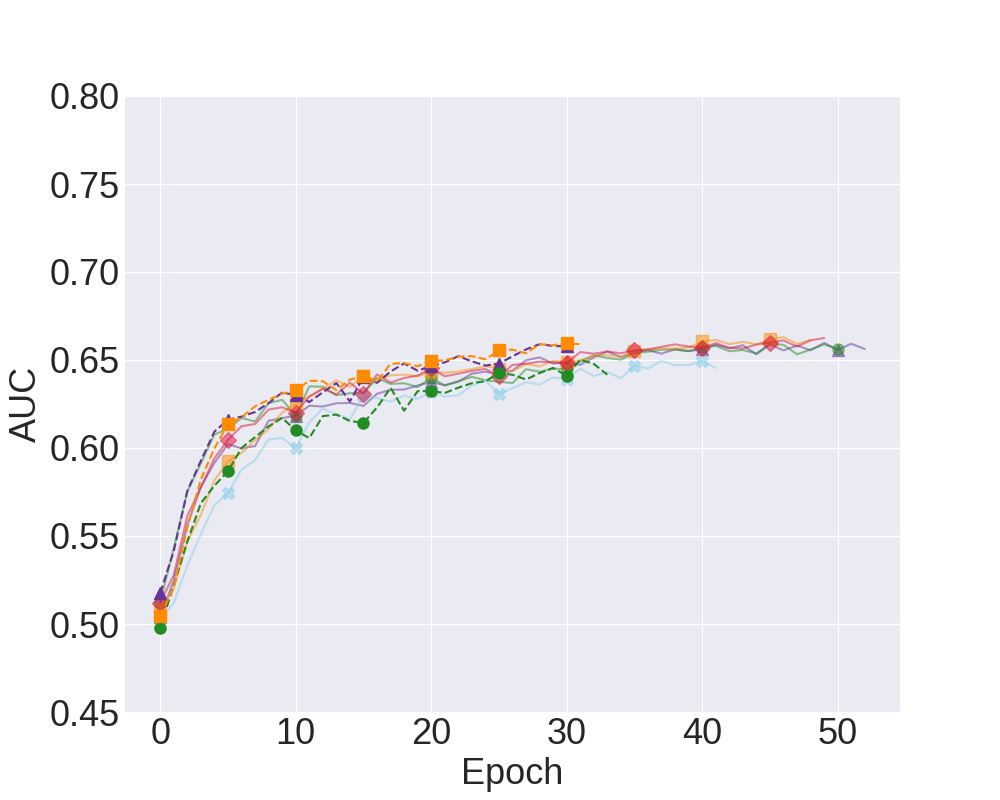}
                                        \caption{$\beta$ = 0.3}
        \end{subfigure}
                                \begin{subfigure}[h]{0.30\textwidth}
        \centering
        \includegraphics[width=\textwidth]{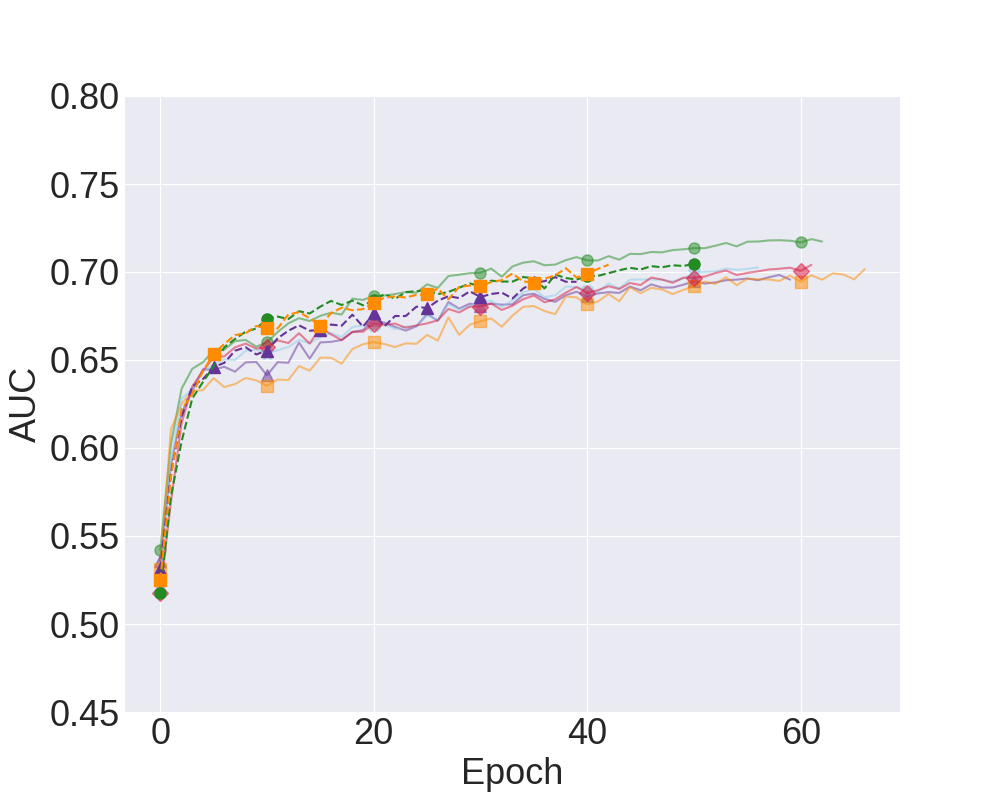}
                                        \caption{$\beta$ = 0.5}
        \end{subfigure}
                                \begin{subfigure}[h]{0.30\textwidth}
        \centering
        \includegraphics[width=\textwidth]{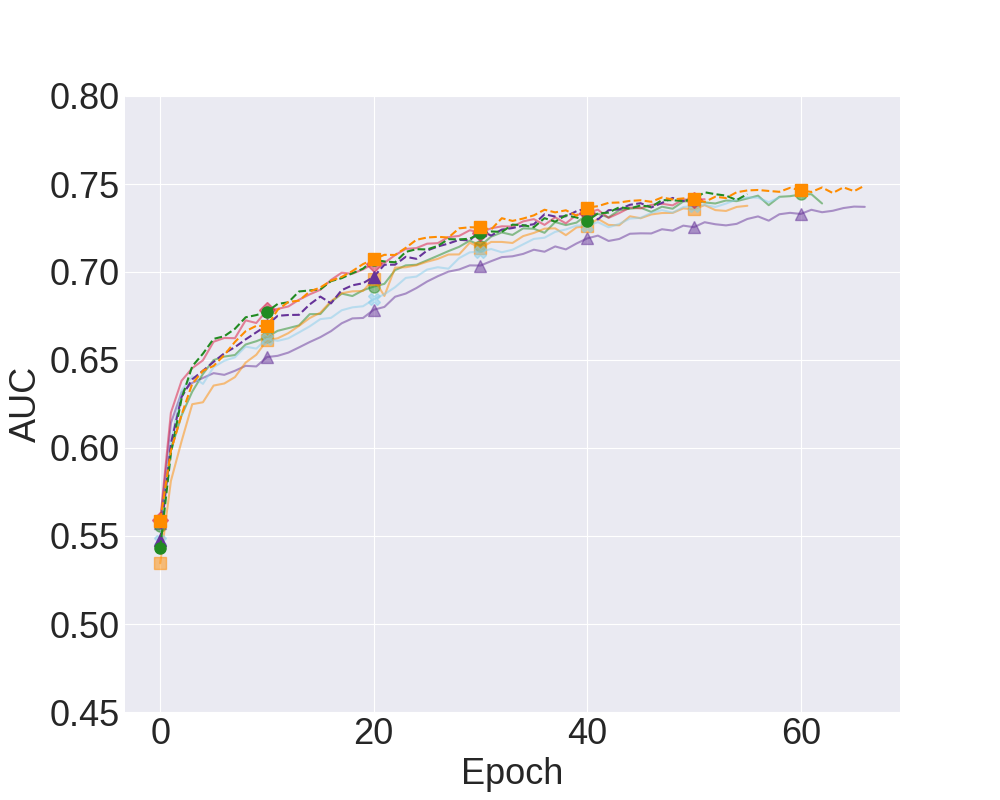}
                                        \caption{$\beta$ = 0.7}
        \end{subfigure}
                                \begin{subfigure}[h]{0.30\textwidth}
        \centering
        \includegraphics[width=\textwidth]{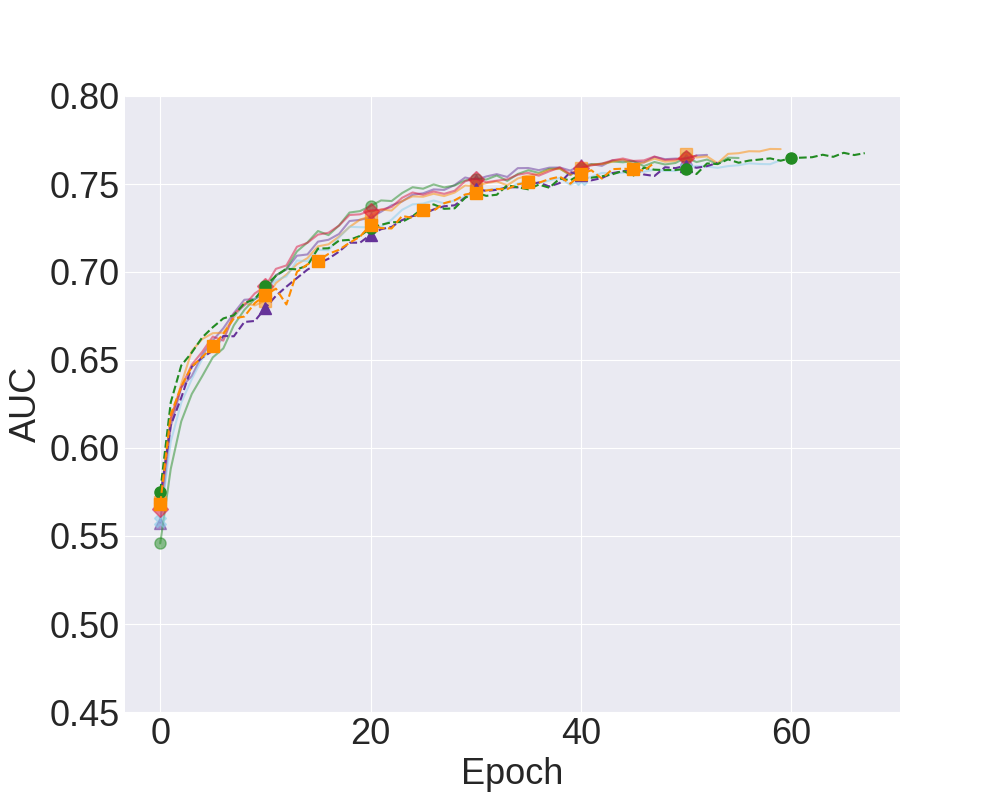}
        \caption{$\beta$ = 0.9}
        \end{subfigure}
        \caption{Mean validation set AUC for the various methodologies and acquisition functions on $\mathcal{D}_{4}$ at increasing fraction levels $\beta = (0.1, 0.3, 0.5, 0.7, 0.9)$. Our methods include MCP and BALC methods. The no-oracle cold-start problem is \textit{not} observed for this dataset. Most methods perform comparably to one another at high values of $\beta$. Results are averaged across 5 seeds.}
\label{fig:impact_of_perturbations_cardiology_ecg}
\end{figure}

In the main manuscript, we presented a subset of results for experiments in which oracles are absent and thus unavailable to provide annotations. Instead, unlabelled instances are pseudo-labelled based on network-generated predictions. In this section, we include an exhaustive set of results for all those experiments. More specifically, we illustrate in Figs.~\ref{fig:impact_of_perturbations_physionet_ppg} - \ref{fig:impact_of_perturbations_cardiology_ecg} the validation AUC of the various AL methods for datasets $\mathcal{D}_{1}$ - $\mathcal{D}_{5}$. At a high level and across datasets, we find that the cold-start problem is likely to occur at low fraction values ($\beta=0.1$). We include more details in the respective sections.

\subsection{PhysioNet 2015 PPG, $\mathcal{D}_{1}$}

\clearpage

\subsection{PhysioNet 2015 ECG, $\mathcal{D}_{2}$}

\clearpage

\subsection{PhysioNet 2017 ECG, $\mathcal{D}_{3}$}

\clearpage

\subsection{Cardiology ECG, $\mathcal{D}_{4}$}

\end{subappendices}

\clearpage

\begin{subappendices}
\section{Effect of MCP with Tracked Acquisition Functions on Performance}
\label{appendix:effect_of_temporal_functions}

\begin{figure}[!h]
    \centering
    \begin{subfigure}[h]{0.48\textwidth}
    \centering
    \includegraphics[width=1\textwidth]{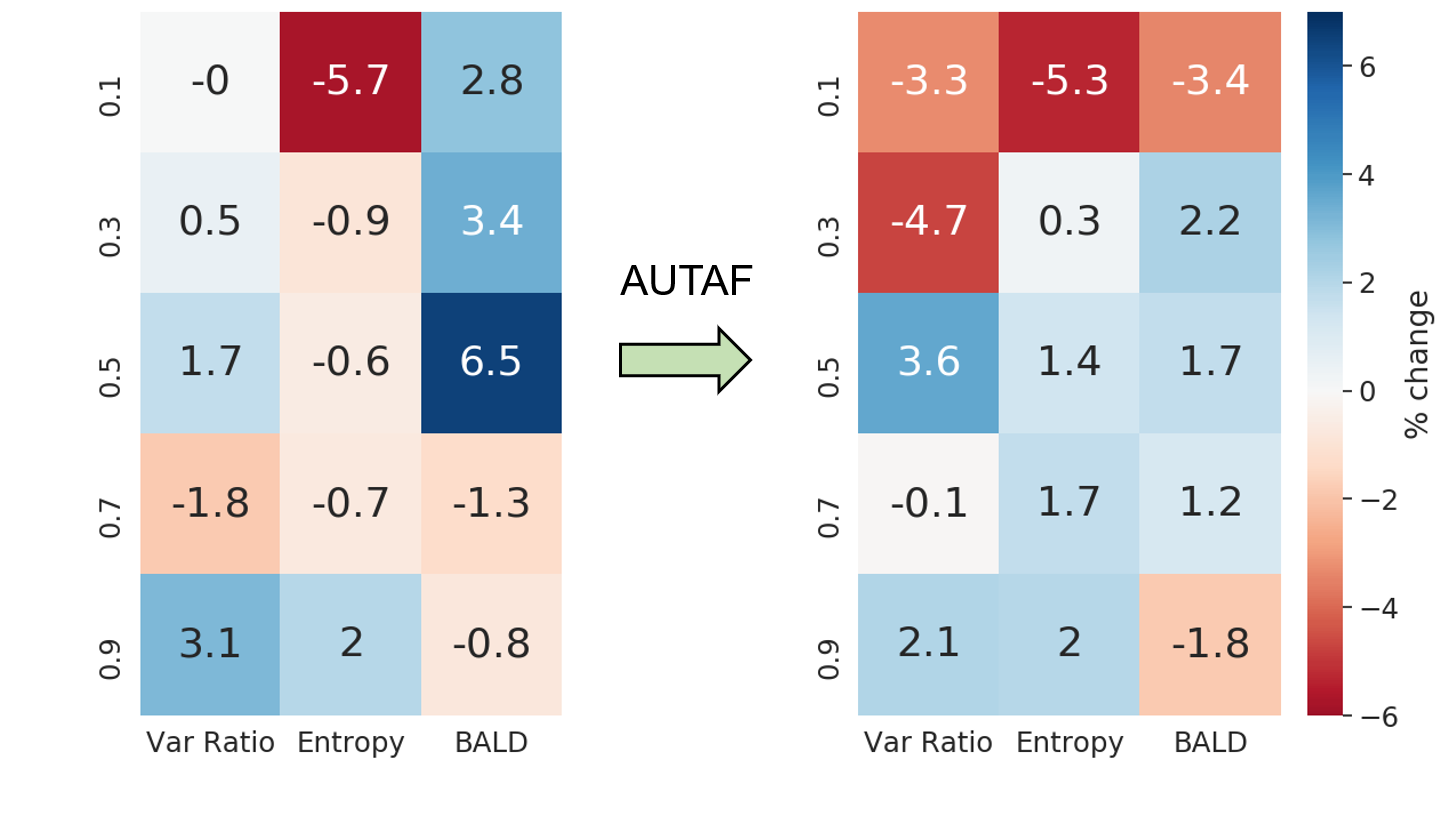}
    \caption{$\mathcal{D}_{2}$}
    \end{subfigure}
    ~
    \begin{subfigure}[h]{0.48\textwidth}
    \centering
    \includegraphics[width=1\textwidth]{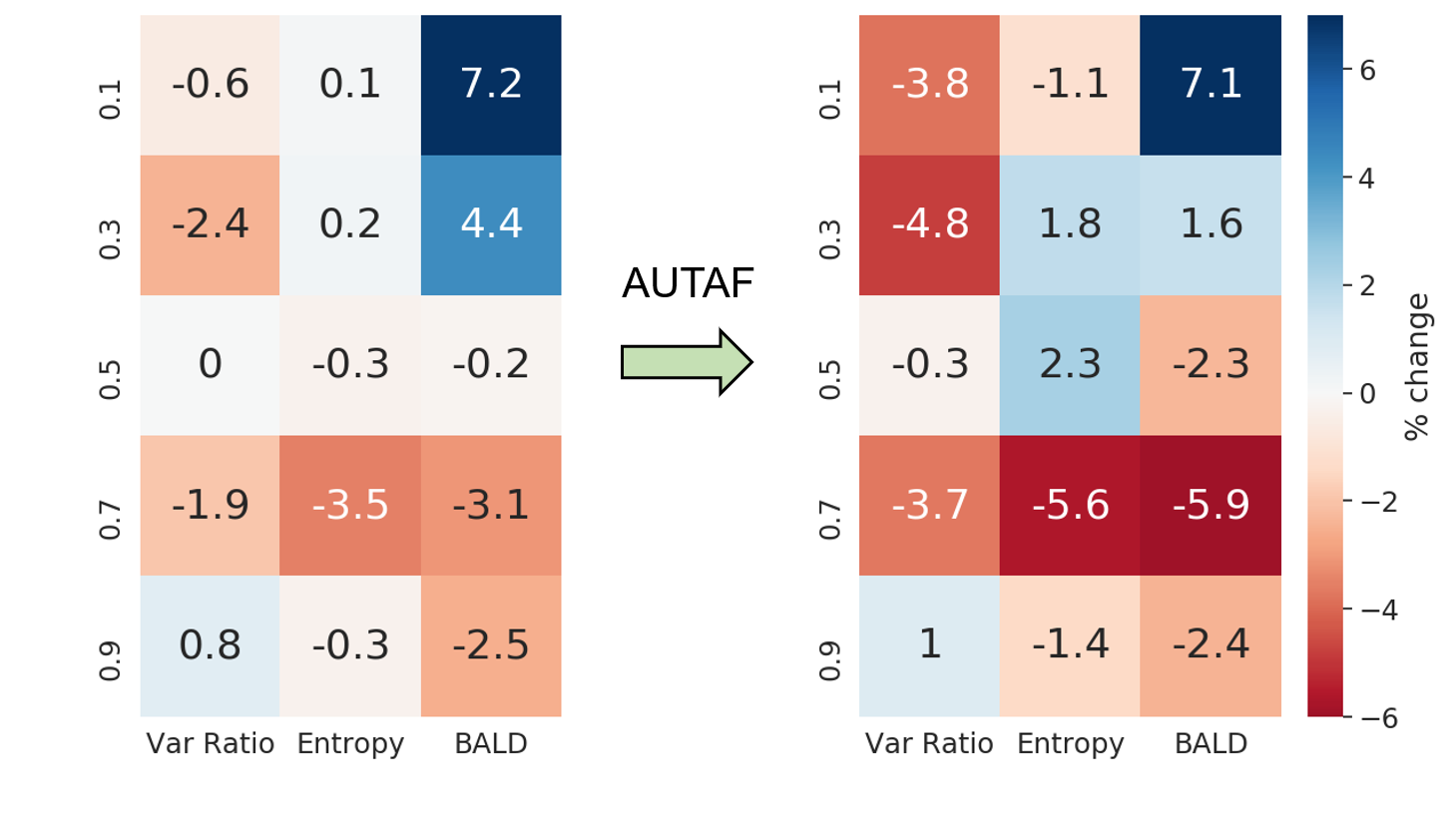}
    \caption{$\mathcal{D}_{3}$}
    \end{subfigure}
    ~
    \begin{subfigure}[h]{0.48\textwidth}
    \centering
    \includegraphics[width=1\textwidth]{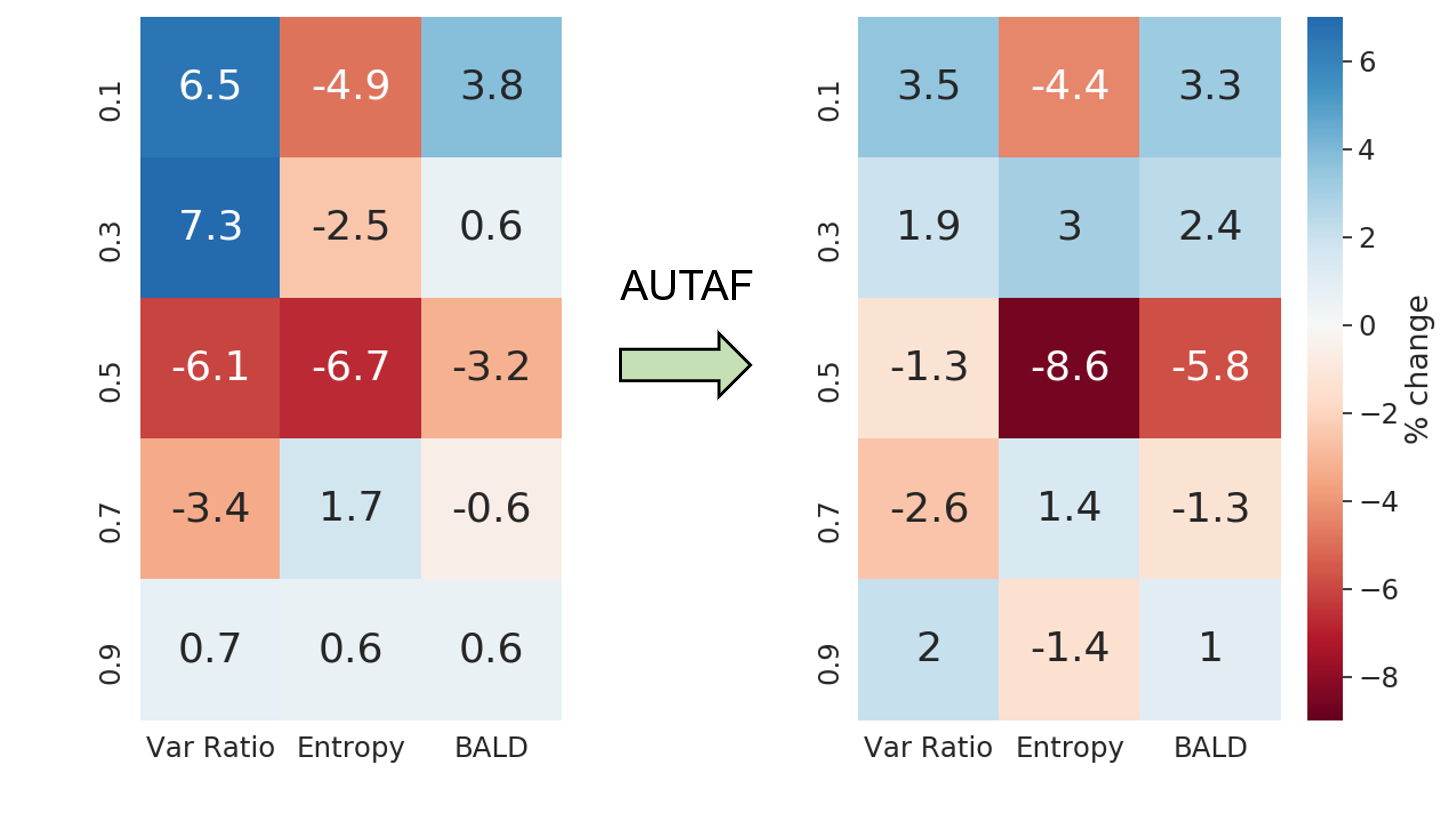}
    \caption{$\mathcal{D}_{4}$}
    \end{subfigure}
    \caption{Mean percent change in test AUC when comparing MCP with static and tracked acquisition functions to MCD with their static counterparts on \textbf{(a)} $\mathcal{D}_{2}$ and \textbf{(b)} $\mathcal{D}_{3}$ and \textbf{(c)} $\mathcal{D}_{4}$ . We show results for Var Ratio, Entropy, and BALD, at all fractions, $\beta \in [0.1,0.3,0.5,0.7,0.9]$.}
    \label{fig:effect_of_tracked_aq}
\end{figure}

In this section, we are interested in quantifying the effect of implementing a temporal acquisition function in conjunction with MCP on performance. In Fig.~\ref{fig:effect_of_tracked_aq}, we illustrate two columns of matrices. The first column reflects the percent change in generalization performance between implementing MCP and MCD with static temporal functions (i.e., without tracking) for three different datasets. 

We find that there are mixed results. For example, on $mathcal{D}_{2}$ at $\beta=0.5$, $\mathrm{BALD_{MCP}}$ outperforms $\mathrm{BALD_{MCD}}$ by $6.5\%$. However, on $mathcal{D}_{4}$ at $\beta=0.5$, $\mathrm{Entropy_{MCP}}$ performs worse than $\mathrm{Entropy_{MCD}}$ by $6.7\%$. Furthermore, upon applying tracked acquisition functions, we also obtain mixed results. In many cases, there are notable improvements. For example, on  $mathcal{D}_{3}$ at $\beta=0.5$, $\mathrm{Temporal \ Entropy_{MCP}}$ improves performance by an additional $0.3 + 2.3 = 2.6\%$. On the other hand, at $\beta=0.7$, $\mathrm{Temporal \ Entropy_{MCP}}$ worsens performance by $2.1\%$. Based on these findings, we would recommend that the utility of temporal acquisition functions be determined on a case-by-case basis.

\end{subappendices}

\clearpage

\begin{subappendices}
\section{Dependence of SoQal on Oracle}
\renewcommand{\thesubsection}{\Alph{section}.\arabic{subsection}}
\label{appendix:soqal_oracle_dependence}

\begin{figure}[!b]
    \centering
    \begin{subfigure}[h]{0.75\textwidth}
    \centering
    \includegraphics[width=\textwidth]{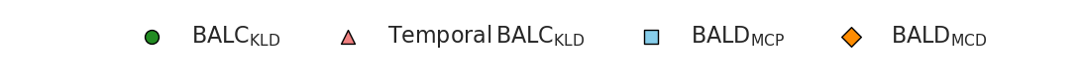}
    \end{subfigure}
    ~
    \begin{subfigure}[h]{1\textwidth}
    \centering
    \includegraphics[width=\textwidth]{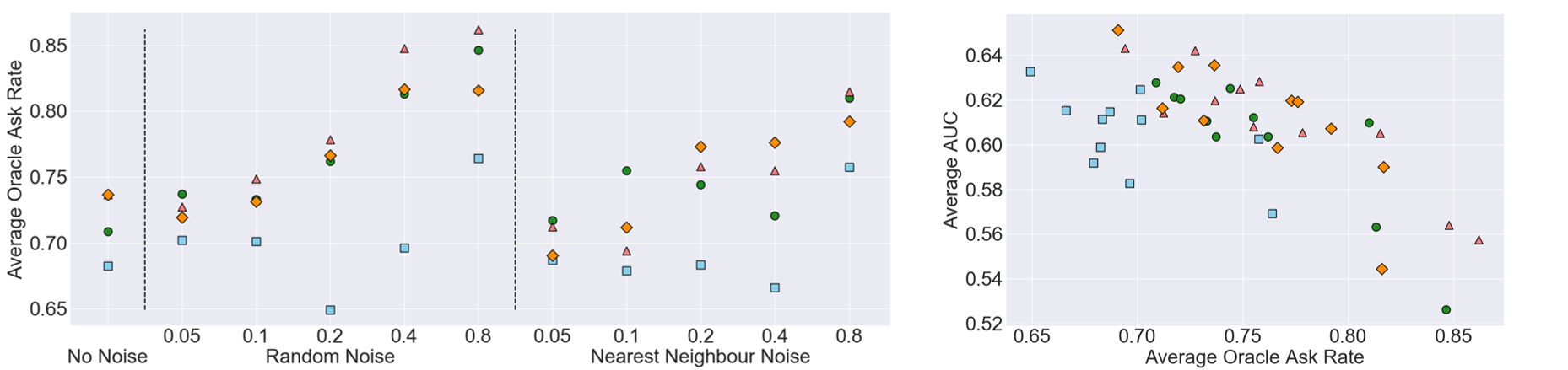}
    \caption{}
    \label{fig:average_oracle_ask_rate}
	\end{subfigure}
	~
    \begin{subfigure}{0.48\textwidth}
    \includegraphics[width=1\textwidth]{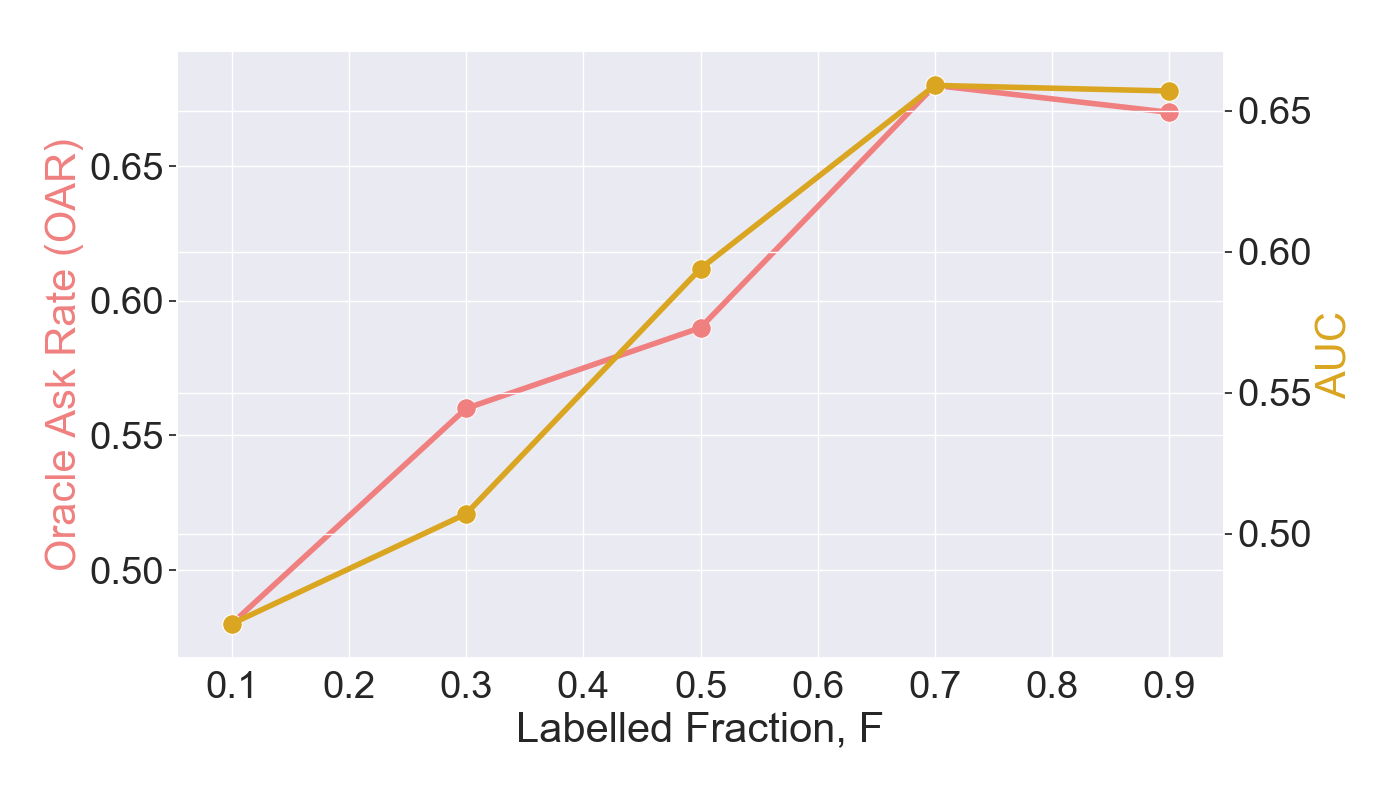}
    \caption{}
    \label{fig:oracle_dependence_data}
    \end{subfigure}
    ~
    \begin{subfigure}{0.48\textwidth}
    \includegraphics[width=1\textwidth]{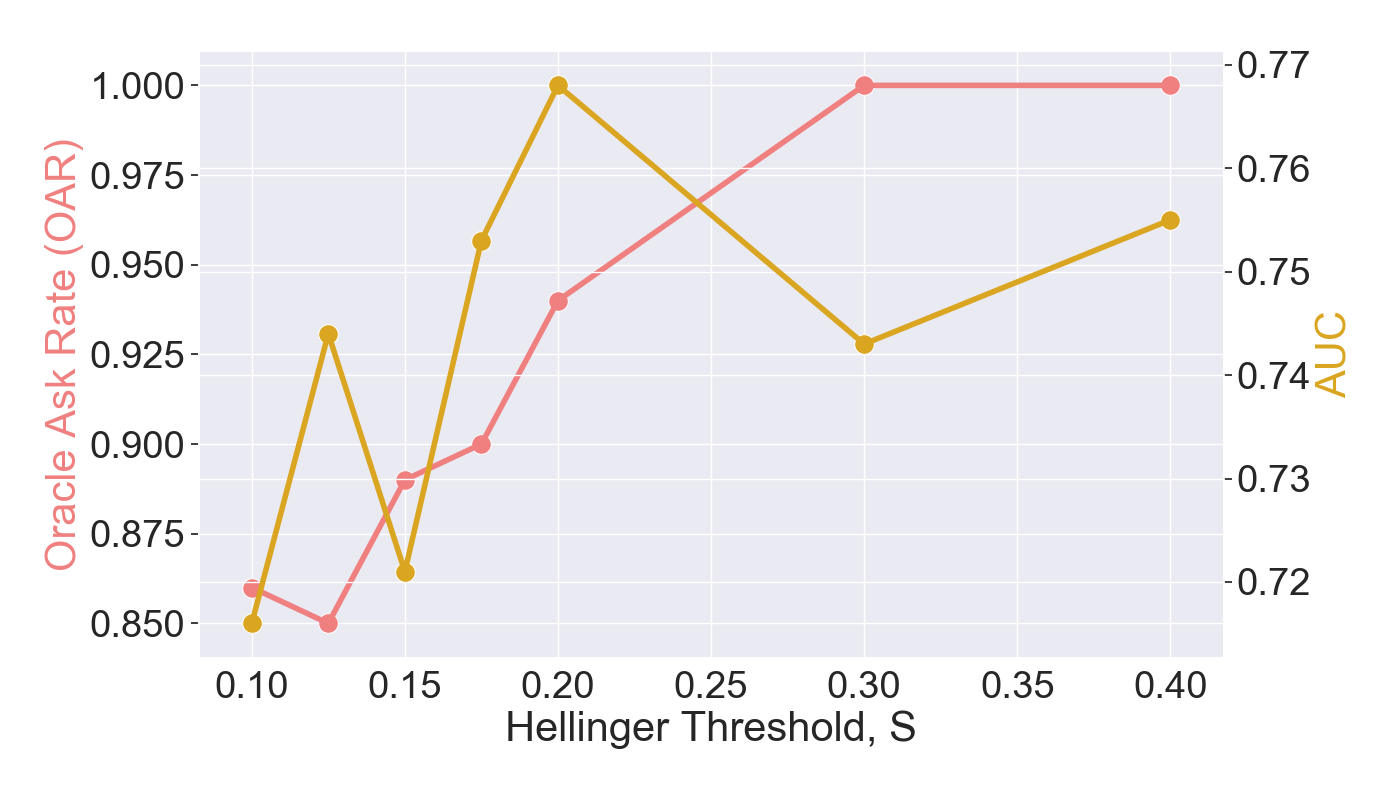}
    \caption{}
    \label{fig:oracle_dependence_threshold}
    \end{subfigure}
    \caption{\textbf{Dependence of SoQal on oracle as a function label noise, data availability, and the Hellinger threshold.} \textbf{(a)} (left) Average oracle ask rate (OAR) as a function of various types and levels of label noise, shown for a subset of acquisition functions. (right) Correlation between oracle ask rate and generalization performance. We also present the OAR and corresponding AUC while using $\mathrm{BALD}_{\mathrm{MCD}}$ as a function of \textbf{(b)} the labelled fraction, $F$, of labelled data and \textbf{(c)} the Hellinger threshold, $S$. All results are averaged across five random seeds. In (a) (right), we show that the reduced dependence of a network on an oracle can be advantageous. This is particularly true if the oracle exhibits label noise. In (b), we show that increasing the amount of labelled data increases the network's dependence on an oracle. In (c), we show that tuning $S$ provides researchers with flexibility over how often to request a label from an oracle.}
    \label{fig:oracle_dependence_extra}
\end{figure}

Recall that one of our original goals is to reduce the labelling burden that is placed on an oracle (e.g., physician). In this section, we look to quantify the degree to which our active learning framework is dependent on an oracle. To do so, we define a metric, which we refer to as the oracle ask rate (OAR), that quantifies the proportion of all acquired instances whose labels are requested from an oracle. For example, an $\mathrm{OAR}=50\%$ implies that $50\%$ of the acquired instances had labels provided by an oracle, with the remaining $50\%$ pseudo-labelled by the network. In Fig.~\ref{fig:average_oracle_ask_rate} (left), we present the oracle ask rate of $\mathrm{SoQal}$ as a function of the type and magnitude of label noise. 

\subsection{Label noise and dependence on oracle} 

In Fig.~\ref{fig:average_oracle_ask_rate} (left), we find that SoQal alleviates the labelling burden placed on an oracle. For example, SoQal when used in conjunction with $\mathrm{BALD}_{\mathrm{MCP}}$ and without label noise, achieves $\mathrm{OAR} \approx 0.68 < 1.0$. This suggests that $32\%$ fewer label requests were sent to an oracle compared to an active learning framework that requests $100\%$ of its labels from an oracle. However, these fewer label requests do result in slightly lower performance compared to the $100\%$ method (see earlier Table~\ref{table:effect_of_oracle_strategies}). This is expected since noise-free oracle-provided labels are likely to be of higher quality than those generated by a network (via pseudo-labels). 

We also find that $\mathrm{SoQal}$ can become more dependent on an oracle with increased levels of noise. For example, with random noise $\gamma=0.05 \rightarrow 0.8$, the average $\mathrm{OAR} \approx 0.72 \rightarrow 0.80$. This increased dependence can be problematic, particularly when the oracle is highly likely to annotate unlabelled instances incorrectly. Nonetheless, $\mathrm{SoQal}$, even with extreme label noise, achieves $\mathrm{OAR} < 1.0$ (fewer requests) while continuing to outperform the remaining questioning methods (see earlier Fig.~\ref{fig:average_auc_performance}). 

In Fig.~\ref{fig:average_oracle_ask_rate} (right), we further explore the relationship between the oracle ask rate and generalization performance. We find that, with label noise, the decreased dependence of $\mathrm{SoQal}$ on an oracle is associated with improved generalization performance. This is shown by the negative correlation between the average $\mathrm{OAR}$ and the $\mathrm{AUC}$. For example, with $\mathrm{OAR} \approx 0.85 \rightarrow 0.65$, the $\mathrm{AUC} \approx 0.53 \rightarrow 0.63$. This suggests that $\mathrm{SoQal}$ performs better while requesting \textit{fewer} labels. Such a finding reaffirms the observation that $\mathrm{SoQal}$ appropriately learns \textit{when} to request a label from an oracle or to pseudo-label instead.

\subsection{Data availability and dependence on oracle} 
The previous oracle questioning results we presented were based on experiments conducted on limited, labelled datasets (i.e., $F=0.1$). In this section, we explore the dependence of SoQal on an oracle when provided with more labelled data (i.e., $F>0.1$). In Fig.~\ref{fig:oracle_dependence_data}, we present the $\mathrm{OAR}$ and corresponding $\mathrm{AUC}$ as a function of the amount of available labelled data, $F$. As expected, we find that performance improves with more data. For example, as $F=0.1 \rightarrow 0.9$, $\mathrm{AUC} \approx 0.50 \rightarrow 0.65$. We also find that $\mathrm{SoQal}$ continues to alleviate the labelling burden placed on an oracle even when more labelled data are available. For example, at $F=0.9$, $\mathrm{SoQal}$ exhibits $\mathrm{OAR} \approx 0.67 < 1.0$, reflecting a $33\%$ reduction in labelling burden. 

\subsection{Hellinger threshold and dependence on oracle}
We claimed that the Hellinger threshold, $S$, can be tuned according to the relative trust one has in the network and the oracle. A higher threshold ($\uparrow S$) implies that less trust is placed in the network than in the oracle. In Fig.~\ref{fig:oracle_dependence_threshold}, we present the $\mathrm{OAR}$ and corresponding $\mathrm{AUC}$ as a function of the Hellinger threshold, $S$. 

Consistent with expectations, we find that the dependence of $\mathrm{SoQal}$ on the oracle increases with the threshold. For example, as $S=0.10 \rightarrow 0.40$, $\mathrm{OAR} \approx 0.86 \rightarrow 1.0$. Such a finding suggests that researchers can set the threshold, $S$, a priori based on the extent to which they would like to request labels from an oracle. We also find that this increased dependence sometimes results in worse performance. For example, although transitioning from $S=0.20 \rightarrow 0.30$ leads to $\mathrm{OAR} \approx 0.95 \rightarrow 1.0$, the $\mathrm{AUC} \approx 0.77 \rightarrow 0.74$. We hypothesize that this worse performance is due to inherent label noise in the datasets. Therefore, a strategy which depends on an oracle $100\%$ of the time is worse than one which delegates some of the labelling to the network instead. Such a finding provides further evidence in support of a dynamic oracle selection strategy.

\end{subappendices}

\clearpage

\begin{subappendices}
\renewcommand{\thesubsection}{\Alph{section}.\arabic{subsection}}
\section{Performance of Oracle Selection Strategies with Noisy Oracle}
\label{appendix:effect_of_label_noise}

Over-reliance on an oracle could be detrimental for an active learning algorithm if that oracle is unable to label instances accurately. In the case of physicians, this inability could arise due to poor training, fatigue, or the difficulty of a particular case being diagnosed. We simulate these scenarios by injecting label noise of various magnitude into the datasets. In this section, we illustrate the performance of three oracle selection strategies, SoQal, Epsilon Greedy, and Entropy Response, in response to label noise. The results are shown for random and nearest neighbour noise in Secs.~\ref{appendix:effect_of_label_noise_random} and \ref{appendix:effect_of_label_noise_nearest_neighbour}, respectively. 

\subsection{Label Noise - Random}
\label{appendix:effect_of_label_noise_random}

Although random noise can be considered an extreme case, it is nonetheless plausible in certain scenarios where labellers are poorly trained or the task at hand is too difficult. In this section, we illustrate, in Tables~\ref{table:noise_random_soqal} - \ref{table:noise_random_ER}, the degree to which the test AUC is affected by the introduction of random label noise during the active learning procedure. 

As expected, extreme levels of noise negatively affect performance. For instance, this can be seen in Table~\ref{table:noise_random_soqal} at $\mathcal{D}_{2}$ using BALD\textsubscript{MCD} where increasing the level of random noise from $5\% \xrightarrow{} 80\%$ leads to a reduction of $\mathrm{AUC} = 0.679 \xrightarrow{} 0.556$. Across most noise levels, SoQal continues to outperform Epsilon Greedy and Entropy Response. This finding is consistent with that presented in the main manuscript and illustrates the relative robustness of SoQal to label noise. 

\begin{table}[h]
\centering
\caption{Mean test AUC of oracle questioning strategies as a function of increasing levels of \textit{random} label noise by the oracle. Results are shown for datasets $\mathcal{D}_{1} - \mathcal{D}_{5}$ and all acquisition functions. Mean and standard deviation values are shown across five seeds.}
\label{table:effect_of_random_noise}
%\vskip 0.1in 
\begin{subtable}{\textwidth}
\caption{SoQal}
\label{table:noise_random_soqal}
\resizebox{\linewidth}{!}{%
\begin{tabular}{c c | c c c c c }
%\hhline{=======}
\toprule
\multirow{2}{*}{Dataset} & \multirow{2}{*}{Ac. Function $\alpha$}
& \multicolumn{5}{c}{Random Noise Level} \\
 &  & 0.05 & 0.10 & 0.20 & 0.40 & 0.80 \\
%\hhline{=======}
\midrule
\multirow{4}{*}{$\mathcal{D}_{1}$} & $\mathrm{BALD_{MCD}}$ & 0.595 $\pm$   0.053 &  0.554 $\pm$  0.028 &  0.558 $\pm$ 0.042 &  0.600 $\pm$  0.029 &  0.511 $\pm$  0.055 \\
& $\mathrm{BALD_{MCP}}$ & 0.659 $\pm$   0.014 &  0.650 $\pm$  0.027 &  0.636 $\pm$  0.029 &  0.549 $\pm$  0.058 &  0.528 $\pm$  0.029 \\
& $\mathrm{BALC_{KLD}}$ & 0.564 $\pm$  0.058 &  0.570 $\pm$  0.045 &  0.562 $\pm$  0.067 &  0.498 $\pm$  0.038 &  0.477 $\pm$  0.011 \\
& Temporal $\mathrm{BALC_{KLD}}$ &  0.634 $\pm$   0.026 &  0.597 $\pm$  0.035 &  0.611 $\pm$  0.040 &  0.494 $\pm$ 0.034 &  0.490 $\pm$  0.026 \\
\hline

\multirow{4}{*}{$\mathcal{D}_{2}$} & $\mathrm{BALD_{MCD}}$ & 0.679 $\pm$   0.017 &  0.659 $\pm$  0.042 &  0.646 $\pm$ 0.044 &  0.602 $\pm$  0.047 &  0.556 $\pm$  0.065 \\
& $\mathrm{BALD_{MCP}}$ & 0.643 $\pm$   0.020 &  0.677 $\pm$  0.053 &  0.637 $\pm$  0.042 &  0.619 $\pm$  0.033 &  0.602 $\pm$  0.041 \\
& $\mathrm{BALC_{KLD}}$ & 0.652 $\pm$   0.037 &  0.659 $\pm$  0.056 &  0.649 $\pm$  0.054 &  0.614 $\pm$  0.016 &  0.522 $\pm$  0.032 \\
& Temporal $\mathrm{BALC_{KLD}}$ & 0.655 $\pm$   0.048 &  0.701 $\pm$  0.029 &  0.628 $\pm$ 0.074 &  0.594 $\pm$  0.041 &  0.581 $\pm$  0.032\\
\hline

\multirow{4}{*}{$\mathcal{D}_{3}$} & $\mathrm{BALD_{MCD}}$ & 0.750 $\pm$   0.017 &  0.742 $\pm$  0.031 &  0.718 $\pm$  0.037 &  0.646 $\pm$  0.023 &  0.584 $\pm$  0.017 \\
& $\mathrm{BALD_{MCP}}$ & 0.724 $\pm$  0.022 &  0.707 $\pm$  0.021 &  0.682 $\pm$  0.038 &  0.629 $\pm$  0.029 &  0.537 $\pm$ 0.025 \\
& $\mathrm{BALC_{KLD}}$ & 0.724 $\pm$   0.032 &  0.725 $\pm$  0.028 &  0.702 $\pm$  0.024 &  0.651 $\pm$  0.046 &  0.564 $\pm$  0.040 \\
& Temporal $\mathrm{BALC_{KLD}}$ & 0.725 $\pm$   0.031 &  0.738 $\pm$  0.013 &  0.705 $\pm$  0.017 &  0.596 $\pm$  0.071 &  0.546 $\pm$  0.041\\
\hline

\multirow{4}{*}{$\mathcal{D}_{4}$} & $\mathrm{BALD_{MCD}}$ & 0.506 $\pm$   0.019 &  0.479 $\pm$  0.022 &  0.496 $\pm$  0.023 &  0.490 $\pm$  0.010 &  0.518 $\pm$  0.029 \\
& $\mathrm{BALD_{MCP}}$ & 0.499 $\pm$   0.037 &  0.508 $\pm$  0.022 &  0.523 $\pm$  0.027 &  0.495 $\pm$  0.023 &  0.503 $\pm$  0.021 \\
& $\mathrm{BALC_{KLD}}$ & 0.491 $\pm$   0.026 &  0.481 $\pm$ 0.023 &  0.496 $\pm$  0.031 &  0.518 $\pm$  0.012 &  0.525 $\pm$  0.011 \\
& Temporal $\mathrm{BALC_{KLD}}$ & 0.522 $\pm$   0.016 &  0.505 $\pm$ 0.027 &  0.501 $\pm$  0.021 &  0.511 $\pm$ 0.025 &  0.515 $\pm$  0.031 \\
% \hline

% \multirow{4}{*}{$\mathcal{D}_{5}$} & $\mathrm{BALD_{MCD}}$ &0.644 $\pm$   0.115 &  0.619 $\pm$  0.083 &  0.576 $\pm$  0.122 &  0.613 $\pm$  0.164 &  0.552 $\pm$  0.143\\
% & $\mathrm{BALD_{MCP}}$ & 0.531 $\pm$   0.128 &  0.583 $\pm$  0.113 &  0.684 $\pm$  0.042 &  0.623 $\pm$  0.046 &  0.675 $\pm$  0.094   \\
% & $\mathrm{BALC_{KLD}}$ & 0.588 $\pm$   0.111 &  0.618 $\pm$  0.120 &  0.608 $\pm$  0.111 &  0.535 $\pm$  0.139 &  0.543 $\pm$  0.134   \\
% & Temporal $\mathrm{BALC_{KLD}}$ & 0.675 $\pm$   0.042 &  0.583 $\pm$  0.143 &  0.583 $\pm$ 0.144 &  0.625 $\pm$  0.108 &  0.655 $\pm$ 0.070 \\
%\hhline{=======}
\bottomrule
\end{tabular}}
\end{subtable}
\end{table}

% \newline
\vspace*{1 cm}
% \newline

\begin{table}[!t]
\ContinuedFloat
\centering
\begin{subtable}[!t]{\textwidth}
\caption{Epsilon Greedy}
\label{table:noise_random_greedy}
\centering
\resizebox{\linewidth}{!}{%
\begin{tabular}{c c | c c c c c }
%\hhline{=======}
\toprule
\multirow{2}{*}{Dataset} & \multirow{2}{*}{Ac. Function $\alpha$}
& \multicolumn{5}{c}{Random Noise Level} \\
 &  & 0.05 & 0.10 & 0.20 & 0.40 & 0.80 \\
%\hhline{=======}
\midrule
\multirow{4}{*}{$\mathcal{D}_{1}$} & $\mathrm{BALD_{MCD}}$ & 0.496$\pm$   0.058 &  0.494 $\pm$  0.029 &  0.476 $\pm$  0.030 &  0.507 $\pm$  0.044 &  0.501 $\pm$ 0.056 \\
& $\mathrm{BALD_{MCP}}$ & 0.557 $\pm$   0.018 &  0.549 $\pm$  0.036 &  0.508 $\pm$  0.032 &  0.511 $\pm$  0.033 &  0.497 $\pm$  0.057 \\
& $\mathrm{BALC_{KLD}}$ & 0.517 $\pm$ 0.014 &  0.518 $\pm$  0.035 &  0.504 $\pm$  0.028 &  0.506 $\pm$  0.034 &  0.498 $\pm$  0.017 \\
& Temporal $\mathrm{BALC_{KLD}}$ & 0.525 $\pm$   0.037 &  0.512 $\pm$  0.040 &  0.501 $\pm$ 0.025 &  0.493 $\pm$  0.019 &  0.497 $\pm$  0.039 \\
\hline
\multirow{4}{*}{$\mathcal{D}_{2}$} & $\mathrm{BALD_{MCD}}$ & 0.600$\pm$   0.053 &  0.628 $\pm$  0.053 &  0.589 $\pm$  0.037 &  0.612 $\pm$  0.022 &  0.555 $\pm$  0.041 \\
& $\mathrm{BALD_{MCP}}$ & 0.629 $\pm$  0.023 &  0.614 $\pm$ 0.048 &  0.536 $\pm$  0.081 &  0.575 $\pm$  0.029 &  0.588 $\pm$  0.050 \\
& $\mathrm{BALC_{KLD}}$ & 0.619 $\pm$   0.038 &  0.586 $\pm$  0.054 &  0.629 $\pm$  0.061 &  0.613 $\pm$  0.045 &  0.582 $\pm$  0.067 \\
& Temporal $\mathrm{BALC_{KLD}}$ & 0.630 $\pm$   0.041 &  0.652 $\pm$  0.029 &  0.579 $\pm$  0.034 &  0.610 $\pm$  0.036 &  0.564 $\pm$  0.035 \\
\hline
\multirow{4}{*}{$\mathcal{D}_{3}$} & $\mathrm{BALD_{MCD}}$ & 0.663 $\pm$   0.018 &  0.661 $\pm$  0.011 &  0.632 $\pm$  0.021 &  0.626$\pm$  0.012 &  0.588 $\pm$  0.017 \\
& $\mathrm{BALD_{MCP}}$ & 0.671 $\pm$   0.017 &  0.670 $\pm$  0.016 &  0.639 $\pm$  0.024 &  0.623 $\pm$  0.019 &  0.574 $\pm$  0.041 \\
& $\mathrm{BALC_{KLD}}$ & 0.665 $\pm$   0.022 &  0.650 $\pm$  0.014 &  0.664 $\pm$  0.013 &  0.618 $\pm$  0.034 &  0.595 $\pm$  0.031 \\
& Temporal $\mathrm{BALC_{KLD}}$ & 0.661 $\pm$   0.012 &  0.651 $\pm$  0.018 &  0.651 $\pm$ 0.016 &  0.629 $\pm$  0.019 &  0.612 $\pm$  0.049 \\
\hline
\multirow{4}{*}{$\mathcal{D}_{4}$} & $\mathrm{BALD_{MCD}}$ & 0.473 $\pm$   0.030 &  0.480 $\pm$  0.033 &  0.469 $\pm$  0.024 &  0.468$\pm$  0.018 &  0.493 $\pm$  0.015 \\
& $\mathrm{BALD_{MCP}}$ & 0.508 $\pm$   0.016 &  0.495 $\pm$  0.019 &  0.498 $\pm$  0.043 &  0.494 $\pm$  0.032 &  0.497 $\pm$  0.015 \\
& $\mathrm{BALC_{KLD}}$ & 0.492 $\pm$   0.026 &  0.496 $\pm$  0.021 &  0.481 $\pm$  0.025 &  0.491 $\pm$  0.020 &  0.498 $\pm$  0.021 \\
& Temporal $\mathrm{BALC_{KLD}}$ & 0.514 $\pm$   0.017 &  0.528 $\pm$  0.017 &  0.500 $\pm$  0.008 &  0.498 $\pm$  0.033 &  0.504 $\pm$ 0.037 \\
% \hline
% \multirow{4}{*}{$\mathcal{D}_{5}$} & $\mathrm{BALD_{MCD}}$ & 0.722 $\pm$   0.008 &  0.717 $\pm$ 0.008 &  0.724 $\pm$  0.016 &  0.699 $\pm$  0.038 &  0.699 $\pm$  0.020 \\
% & $\mathrm{BALD_{MCP}}$ & 0.685 $\pm$   0.060 &  0.716 $\pm$  0.045 &  0.719 $\pm$  0.015 &  0.719 $\pm$  0.004 &  0.708 $\pm$  0.034 \\
% & $\mathrm{BALC_{KLD}}$ & 0.716 $\pm$  0.057 &  0.735 $\pm$  0.029 &  0.734 $\pm$  0.028 &  0.717 $\pm$  0.019 &  0.699 $\pm$  0.017 \\
% & Temporal $\mathrm{BALC_{KLD}}$ & 0.747 $\pm$   0.012 &  0.740 $\pm$  0.014 &  0.679 $\pm$ 0.069 &  0.673 $\pm$  0.103 &  0.651 $\pm$  0.123 \\
%\hhline{=======}
\bottomrule
\end{tabular}}
\end{subtable}

% \newline
\vspace*{1 cm}
% \newline

\begin{subtable}[!t]{\textwidth}
\caption{Entropy Response}
\label{table:noise_random_ER}
\resizebox{\linewidth}{!}{%
\begin{tabular}{c c | c c c c c }
%\hhline{=======}
\toprule
\multirow{2}{*}{Dataset} & \multirow{2}{*}{Ac. Function $\alpha$}
& \multicolumn{5}{c}{Random Noise Level} \\
 &  & 0.05 & 0.10 & 0.20 & 0.40 & 0.80 \\
%\hhline{=======}
\midrule
\multirow{4}{*}{$\mathcal{D}_{1}$} & $\mathrm{BALD_{MCD}}$ & 0.495 $\pm$  0.038 &  0.497 $\pm$  0.057 &  0.498 $\pm$  0.057 &  0.486 $\pm$  0.044 &  0.512 $\pm$  0.057 \\
& $\mathrm{BALD_{MCP}}$ & 0.534 $\pm$   0.018 &  0.584 $\pm$  0.073 &  0.565 $\pm$  0.033 &  0.619 $\pm$  0.022 &  0.518 $\pm$  0.028 \\ 
& $\mathrm{BALC_{KLD}}$ & 0.535 $\pm$   0.038 &  0.521 $\pm$  0.042 &  0.514 $\pm$  0.053 &  0.511 $\pm$  0.027 &  0.525 $\pm$  0.037 \\ 
& Temporal $\mathrm{BALC_{KLD}}$ & 0.526 $\pm$   0.040 &  0.538 $\pm$  0.036 &  0.504 $\pm$  0.028 &  0.501 $\pm$  0.036 &  0.500 $\pm$  0.004\\
\hline
\multirow{4}{*}{$\mathcal{D}_{2}$} & $\mathrm{BALD_{MCD}}$ & 0.587 $\pm$   0.044 &  0.564 $\pm$  0.058 &  0.586 $\pm$  0.047 &  0.613 $\pm$  0.083 &  0.551 $\pm$  0.031 \\
& $\mathrm{BALD_{MCP}}$ & 0.624 $\pm$   0.044 &  0.598 $\pm$  0.057 &  0.573 $\pm$  0.053 &  0.560 $\pm$  0.081 &  0.530 $\pm$  0.015\\
& $\mathrm{BALC_{KLD}}$ & 0.616 $\pm$  0.043 &  0.653 $\pm$  0.049 &  0.624 $\pm$  0.051 &  0.565 $\pm$  0.055 &  0.579 $\pm$  0.019\\ 
& Temporal $\mathrm{BALC_{KLD}}$ & 0.635 $\pm$   0.045 &  0.603 $\pm$  0.050 &  0.590 $\pm$ 0.042 &  0.602 $\pm$  0.046 &  0.579 $\pm$  0.041\\
\hline
\multirow{4}{*}{$\mathcal{D}_{3}$} & $\mathrm{BALD_{MCD}}$ & 0.592 $\pm$   0.015 &  0.604 $\pm$  0.017 &  0.603 $\pm$  0.016 &  0.603 $\pm$  0.016 &  0.605 $\pm$  0.018 \\
& $\mathrm{BALD_{MCP}}$ &0.694 $\pm$  0.047 &  0.730 $\pm$  0.029 &  0.666 $\pm$  0.034 &  0.639 $\pm$  0.031 &  0.599 $\pm$  0.035 \\
& $\mathrm{BALC_{KLD}}$ &0.631 $\pm$   0.006 &  0.631 $\pm$  0.009 &  0.622 $\pm$ 0.011 &  0.631 $\pm$  0.025 &  0.564 $\pm$  0.046\\ 
& Temporal $\mathrm{BALC_{KLD}}$ &0.602 $\pm$   0.011 &  0.622 $\pm$  0.018 &  0.630 $\pm$ 0.014 &  0.618 $\pm$  0.040 &  0.565 $\pm$  0.050\\
\hline
\multirow{4}{*}{$\mathcal{D}_{4}$} & $\mathrm{BALD_{MCD}}$ & 0.472 $\pm$   0.029 &  0.472 $\pm$ 0.030 &  0.486 $\pm$  0.008 &  0.476 $\pm$  0.038 &  0.481 $\pm$  0.038 \\
& $\mathrm{BALD_{MCP}}$ &0.511 $\pm$   0.021 &  0.510 $\pm$  0.023 &  0.525 $\pm$  0.033 &  0.498 $\pm$  0.041 &  0.497 $\pm$  0.017\\
& $\mathrm{BALC_{KLD}}$ &0.468 $\pm$   0.022 &  0.472 $\pm$  0.029 &  0.477 $\pm$ 0.029 &  0.483 $\pm$  0.018 &  0.475 $\pm$  0.032\\ 
& Temporal $\mathrm{BALC_{KLD}}$ &0.482 $\pm$   0.023 &  0.491 $\pm$  0.013 &  0.490 $\pm$ 0.021 &  0.487 $\pm$  0.031 &  0.515 $\pm$ 0.022\\
% \hline
% \multirow{4}{*}{$\mathcal{D}_{5}$} & $\mathrm{BALD_{MCD}}$ & 0.711 $\pm$   0.003 &  0.716 $\pm$  0.011 &  0.716 $\pm$  0.011 &  0.716 $\pm$  0.014 &  0.711 $\pm$  0.003 \\
% & $\mathrm{BALD_{MCP}}$ &0.699 $\pm$   0.014 &  0.679 $\pm$  0.053 &  0.704 $\pm$  0.023 &  0.678 $\pm$  0.061 &  0.704 $\pm$  0.019\\
% & $\mathrm{BALC_{KLD}}$ & 0.687 $\pm$   0.055 &  0.718 $\pm$  0.012 &  0.719 $\pm$  0.017 &  0.691 $\pm$  0.058 &  0.693 $\pm$  0.048\\ 
% & Temporal $\mathrm{BALC_{KLD}}$ & 0.722 $\pm$   0.009 &  0.643 $\pm$  0.091 &  0.682 $\pm$  0.068 &  0.721 $\pm$  0.009 &  0.681 $\pm$  0.067\\
%\hhline{=======}
\bottomrule
\end{tabular}}
\end{subtable}
\end{table}

\clearpage

\subsection{Label Noise - Nearest Neighbour}
\label{appendix:effect_of_label_noise_nearest_neighbour}

Nearest neighbour noise is more realistic than that which is random as it may simulate uncertainty in diagnoses made by physicians. In this section, we illustrate, in Tables~\ref{table:noise_nn_soqal} - \ref{table:noise_nn_ER}, the degree to which the test AUC is affected by the introduction of nearest neighbour label noise during the active learning procedure. 

As expected, extreme levels of noise negatively affect performance. For instance, this can be seen in Table~\ref{table:noise_nn_soqal} at $\mathcal{D}_{3}$ using BALD\textsubscript{MCD} where increasing the level of nearest neighbour noise from $5\% \xrightarrow{} 80\%$ leads to a reduction of the $\mathrm{AUC} = 0.744 \xrightarrow{} 0.694$. SoQal continues to outperform Epsilon Greedy and Entropy Response across most of the noise levels. Building on the previous example, with 80\% nearest neighbour noise, SoQal achieves an $\mathrm{AUC} = 0.694$ whereas Epsilon Greedy and Entropy Response achieve an $\mathrm{AUC} = 0.632$ and $0.587$, respectively. Such a finding is similar to that arrived at with Random Noise and implies that SoQal is relatively more robust to noisy oracles than these other methods.

\begin{table}[!h]
\centering
\caption{Mean test AUC of oracle questioning strategies as a function of increasing levels of \textit{nearest neighbour} label noise by the oracle. Results are shown for datasets $\mathcal{D}_{1} - \mathcal{D}_{5}$ and all acquisition functions. Mean and standard deviation values are shown across five seeds.}
\label{table:effect_of_nn_noise}
%\vskip 0.1in 
\begin{subtable}{\textwidth}
\caption{SoQal}
\label{table:noise_nn_soqal}
\resizebox{\linewidth}{!}{%
\begin{tabular}{c c | c c c c c }
%\hhline{=======}
\toprule
\multirow{2}{*}{Dataset} & \multirow{2}{*}{Ac. Function $\alpha$}
& \multicolumn{5}{c}{Nearest Neighbour Noise Level} \\
 &  & 0.05 & 0.10 & 0.20 & 0.40 & 0.80 \\
%\hhline{=======}
\midrule
\multirow{4}{*}{$\mathcal{D}_{1}$} & $\mathrm{BALD_{MCD}}$ & 0.614 $\pm$   0.043 &  0.571 $\pm$  0.037 &  0.618 $\pm$  0.015 &  0.557 $\pm$  0.042 &  0.540 $\pm$  0.052 \\
& $\mathrm{BALD_{MCP}}$ & 0.633 $\pm$   0.011 &  0.617 $\pm$  0.095 &  0.641 $\pm$  0.023 &  0.632 $\pm$  0.026 &  0.591 $\pm$  0.047 \\
& $\mathrm{BALC_{KLD}}$ & 0.628 $\pm$   0.049 &  0.586 $\pm$  0.032 &  0.616 $\pm$  0.024 &  0.604 $\pm$  0.022 &  0.558 $\pm$  0.069 \\
& Temporal $\mathrm{BALC_{KLD}}$ & 0.557 $\pm$   0.045 &  0.647 $\pm$  0.060 &  0.620 $\pm$  0.036 &  0.625 $\pm$  0.038 &  0.577 $\pm$  0.039 \\
\hline
\multirow{4}{*}{$\mathcal{D}_{2}$} & $\mathrm{BALD_{MCD}}$ &  0.694 $\pm$   0.022 &  0.631 $\pm$  0.020 &  0.682 $\pm$  0.036 &  0.658 $\pm$  0.038 &  0.647 $\pm$  0.039 \\
& $\mathrm{BALD_{MCP}}$ & 0.605 $\pm$   0.054 &  0.660 $\pm$ 0.067 &  0.656 $\pm$  0.029 &  0.618 $\pm$  0.058 &  0.605 $\pm$  0.081 \\
& $\mathrm{BALC_{KLD}}$ & 0.655 $\pm$   0.015 &  0.660 $\pm$ 0.037 &  0.671 $\pm$  0.078 & 0.649 $\pm$  0.024 &  0.678 $\pm$  0.023 \\
& Temporal $\mathrm{BALC_{KLD}}$ & 0.702 $\pm$   0.044 &  0.654 $\pm$  0.024 &  0.686 $\pm$ 0.038 &  0.638 $\pm$  0.042 &  0.631 $\pm$  0.020 \\
\hline
\multirow{4}{*}{$\mathcal{D}_{3}$} & $\mathrm{BALD_{MCD}}$ & 0.744 $\pm$   0.023 &  0.745 $\pm$  0.021 &  0.709 $\pm$  0.028 &  0.700 $\pm$  0.026 &  0.694 $\pm$  0.014 \\
& $\mathrm{BALD_{MCP}}$ & 0.706 $\pm$   0.029 &  0.736 $\pm$  0.036 &  0.727 $\pm$  0.023 &  0.712 $\pm$  0.018 &  0.682 $\pm$  0.017 \\
& $\mathrm{BALC_{KLD}}$ & 0.718 $\pm$   0.029 &  0.729 $\pm$  0.028 &  0.735 $\pm$  0.021 &  0.680 $\pm$  0.050 &  0.688 $\pm$  0.009 \\
& Temporal $\mathrm{BALC_{KLD}}$ & 0.727 $\pm$   0.033 &  0.725 $\pm$ 0.033 &  0.724 $\pm$  0.018 &  0.700 $\pm$  0.022 &  0.645 $\pm$  0.062 \\
\hline
\multirow{4}{*}{$\mathcal{D}_{4}$} & $\mathrm{BALD_{MCD}}$ & 0.517 $\pm$   0.034 &  0.477 $\pm$  0.027 &  0.493 $\pm$ 0.034 &  0.498 $\pm$  0.036 &  0.459 $\pm$  0.035 \\
& $\mathrm{BALD_{MCP}}$ & 0.492 $\pm$   0.027 &  0.491 $\pm$  0.027 &  0.502 $\pm$  0.036 &  0.532 $\pm$  0.040 &  0.507 $\pm$  0.042 \\
& $\mathrm{BALC_{KLD}}$ & 0.494 $\pm$   0.024 &  0.494 $\pm$  0.016 &  0.504 $\pm$  0.026 &  0.506 $\pm$  0.031 &  0.503 $\pm$  0.021 \\
& Temporal $\mathrm{BALC_{KLD}}$ & 0.504 $\pm$   0.018 &  0.515 $\pm$  0.013 &  0.529 $\pm$  0.027 &  0.507 $\pm$  0.014 &  0.508 $\pm$  0.026 \\
% \hline
% \multirow{4}{*}{$\mathcal{D}_{5}$} & $\mathrm{BALD_{MCD}}$ & 0.687 $\pm$   0.018 &  0.657 $\pm$  0.057 &  0.597 $\pm$  0.141 &  0.683 $\pm$ 0.044 &  0.696 $\pm$  0.013 \\
% & $\mathrm{BALD_{MCP}}$  & 0.638 $\pm$   0.034 &  0.454 $\pm$  0.136 &  0.531 $\pm$  0.128 &  0.583 $\pm$  0.113 &  0.629 $\pm$  0.152 \\
% & $\mathrm{BALC_{KLD}}$ & 0.610 $\pm$   0.113 &  0.592 $\pm$  0.110 &  0.599 $\pm$  0.084 &  0.663 $\pm$  0.108 &  0.621 $\pm$  0.104 \\
% & Temporal $\mathrm{BALC_{KLD}}$ & 0.582 $\pm$   0.142 &  0.675 $\pm$  0.042 &  0.583 $\pm$  0.143 &  0.570 $\pm$  0.159 &  0.664 $\pm$  0.096 \\
%\hhline{=======}
\bottomrule
\end{tabular}}
\end{subtable}
\end{table}

% \newline
\vspace*{1 cm}
% \newline

\begin{table}[!t]
\ContinuedFloat
\centering
\begin{subtable}{\textwidth}
\caption{Epsilon Greedy}
\label{table:noise_nn_greedy}
\centering
\resizebox{\linewidth}{!}{%
\begin{tabular}{c c | c c c c c }
%\hhline{=======}
\toprule
\multirow{2}{*}{Dataset} & \multirow{2}{*}{Ac. Function $\alpha$}
& \multicolumn{5}{c}{Nearest Neighbour Noise Level} \\
 &  & 0.05 & 0.10 & 0.20 & 0.40 & 0.80 \\
%\hhline{=======}
\midrule
\multirow{4}{*}{$\mathcal{D}_{1}$} & $\mathrm{BALD_{MCD}}$ & 0.503 $\pm$   0.040 &  0.480 $\pm$  0.023 &  0.514 $\pm$  0.050 &  0.481 $\pm$ 0.038 &  0.456 $\pm$  0.031  \\
& $\mathrm{BALD_{MCP}}$  & 0.501 $\pm$   0.014 &  0.508 $\pm$  0.036 &  0.522 $\pm$  0.052 &  0.482 $\pm$  0.023 &  0.496 $\pm$  0.041\\
& $\mathrm{BALC_{KLD}}$ &0.541 $\pm$   0.035 &  0.503 $\pm$  0.033 &  0.486 $\pm$  0.047 &  0.500 $\pm$  0.041 &  0.473 $\pm$  0.026\\
& Temporal $\mathrm{BALC_{KLD}}$ &0.516 $\pm$   0.024 &  0.523 $\pm$  0.044 &  0.495 $\pm$  0.046 &  0.491 $\pm$  0.013 &  0.483 $\pm$  0.041\\
\hline
\multirow{4}{*}{$\mathcal{D}_{2}$} & $\mathrm{BALD_{MCD}}$ & 0.584 $\pm$   0.066 &  0.610 $\pm$  0.042 &  0.597 $\pm$  0.036 &  0.616 $\pm$ 0.054 &  0.593 $\pm$  0.054\\
& $\mathrm{BALD_{MCP}}$  & 0.565 $\pm$   0.031 &  0.589 $\pm$  0.075 &  0.616 $\pm$  0.059 &  0.605 $\pm$  0.047 &  0.586 $\pm$  0.047 \\
& $\mathrm{BALC_{KLD}}$ &0.608 $\pm$  0.031 &  0.607 $\pm$ 0.040 &  0.590 $\pm$  0.055 &  0.538 $\pm$  0.037 &  0.585 $\pm$  0.053 \\
& Temporal $\mathrm{BALC_{KLD}}$ &0.647 $\pm$   0.044 &  0.591 $\pm$  0.033 &  0.640 $\pm$  0.044 &  0.576 $\pm$  0.031 &  0.589 $\pm$  0.030\\
\hline
\multirow{4}{*}{$\mathcal{D}_{3}$} & $\mathrm{BALD_{MCD}}$ & 0.656 $\pm$  0.021 &  0.655 $\pm$  0.014 &  0.665 $\pm$  0.010 &  0.643 $\pm$  0.021 &  0.632 $\pm$  0.010\\
& $\mathrm{BALD_{MCP}}$  & 0.660 $\pm$   0.022 &  0.657 $\pm$  0.023 &  0.659 $\pm$  0.003 &  0.664 $\pm$  0.023 &  0.634 $\pm$  0.013 \\
& $\mathrm{BALC_{KLD}}$ & 0.608 $\pm$  0.031 &  0.607 $\pm$ 0.040 &  0.590 $\pm$  0.055 &  0.538 $\pm$  0.037 &  0.585 $\pm$  0.053\\
& Temporal $\mathrm{BALC_{KLD}}$ &0.644 $\pm$   0.016 &  0.651 $\pm$  0.011 &  0.658 $\pm$  0.013 &  0.634 $\pm$  0.016 &  0.627 $\pm$  0.014\\
\hline
\multirow{4}{*}{$\mathcal{D}_{4}$} & $\mathrm{BALD_{MCD}}$ & 0.438 $\pm$   0.014 &  0.457 $\pm$ 0.022 &  0.442 $\pm$  0.018 &  0.456 $\pm$  0.028 &  0.428 $\pm$  0.024 \\
& $\mathrm{BALD_{MCP}}$  &  0.489 $\pm$   0.018 &  0.489 $\pm$  0.021 &  0.474 $\pm$  0.023 &  0.485 $\pm$ 0.019 &  0.486 $\pm$0.015 \\
& $\mathrm{BALC_{KLD}}$ &0.485 $\pm$   0.029 &  0.487 $\pm$  0.019 &  0.495 $\pm$  0.023 &  0.488 $\pm$  0.028 &  0.481 $\pm$  0.021 \\
& Temporal $\mathrm{BALC_{KLD}}$ &0.486 $\pm$   0.018 &  0.500 $\pm$  0.029 &  0.486 $\pm$ 0.027 &  0.468 $\pm$  0.017 &  0.487 $\pm$ 0.028 \\
% \hline
% \multirow{4}{*}{$\mathcal{D}_{5}$} & $\mathrm{BALD_{MCD}}$ &  0.728 $\pm$   0.021 &  0.712 $\pm$  0.037 &  0.732 $\pm$  0.017 &  0.732 $\pm$  0.017 &  0.720 $\pm$  0.015 \\
% & $\mathrm{BALD_{MCP}}$  & 0.711 $\pm$   0.017 &  0.759 $\pm$  0.023 &  0.738 $\pm$  0.039 &  0.736 $\pm$  0.028 &  0.694 $\pm$  0.021\\
% & $\mathrm{BALC_{KLD}}$ & 0.702 $\pm$   0.048 &  0.707 $\pm$  0.081 &  0.729 $\pm$  0.013 &  0.710 $\pm$  0.048 &  0.713 $\pm$  0.009 \\
% & Temporal $\mathrm{BALC_{KLD}}$ & 0.728 $\pm$  0.029 &  0.709 $\pm$  0.048 &  0.700 $\pm$  0.067 &  0.731 $\pm$  0.018 &  0.682 $\pm$  0.072\\
%\hhline{=======}
\bottomrule
\end{tabular}}
\end{subtable}

% \newline
\vspace*{1 cm}
% \newline

% \begin{table}[h]
% \ContinuedFloat
% \centering
\begin{subtable}{\textwidth}
\caption{Entropy Response}
\label{table:noise_nn_ER}
\centering
\resizebox{\linewidth}{!}{%
\begin{tabular}{c c | c c c c c }
%\hhline{=======}
\toprule
\multirow{2}{*}{Dataset} & \multirow{2}{*}{Ac. Function $\alpha$}
& \multicolumn{5}{c}{Nearest Neighbour Noise Level} \\
 &  & 0.05 & 0.10 & 0.20 & 0.40 & 0.80 \\
%\hhline{=======}
\midrule
\multirow{4}{*}{$\mathcal{D}_{1}$} & $\mathrm{BALD_{MCD}}$ & 0.494 $\pm$   0.037 &  0.474 $\pm$  0.027 &  0.492 $\pm$  0.051 &  0.482 $\pm$  0.033 &  0.444 $\pm$  0.006 \\
& $\mathrm{BALD_{MCP}}$  & 0.511 $\pm$   0.019 &  0.562 $\pm$  0.052 &  0.509 $\pm$  0.042 &  0.572 $\pm$  0.060 &  0.495 $\pm$  0.041\\
& $\mathrm{BALC_{KLD}}$ &0.513 $\pm$   0.020 &  0.517 $\pm$  0.035 &  0.504 $\pm$  0.034 &  0.498 $\pm$  0.023 &  0.487 $\pm$  0.023\\ 
& Temporal $\mathrm{BALC_{KLD}}$ &0.500 $\pm$   0.043 &  0.540 $\pm$  0.025 &  0.503 $\pm$  0.043 &  0.516 $\pm$  0.024 &  0.490 $\pm$  0.024\\
\hline
\multirow{4}{*}{$\mathcal{D}_{2}$} & $\mathrm{BALD_{MCD}}$ & 0.585 $\pm$   0.045 &  0.630 $\pm$  0.056 &  0.600 $\pm$  0.045 &  0.585 $\pm$  0.046 &  0.586 $\pm$  0.063 \\
& $\mathrm{BALD_{MCP}}$  & 0.633 $\pm$   0.060 &  0.626 $\pm$  0.064 &  0.618 $\pm$  0.055 &  0.647 $\pm$  0.077 &  0.619 $\pm$  0.055 \\
& $\mathrm{BALC_{KLD}}$ &0.605 $\pm$   0.049 &  0.572 $\pm$  0.032 &  0.630 $\pm$  0.081 &  0.581 $\pm$  0.031 &  0.589 $\pm$  0.061\\ 
& Temporal $\mathrm{BALC_{KLD}}$ &0.625 $\pm$   0.030 &  0.599 $\pm$  0.024 &  0.613 $\pm$  0.050 &  0.614 $\pm$  0.052 &  0.606 $\pm$  0.054\\
\hline
\multirow{4}{*}{$\mathcal{D}_{3}$} & $\mathrm{BALD_{MCD}}$ &0.604 $\pm$   0.017 &  0.589 $\pm$  0.013 &  0.592 $\pm$  0.014 &  0.592 $\pm$  0.014 &  0.587 $\pm$  0.012 \\
& $\mathrm{BALD_{MCP}}$  & 0.636 $\pm$   0.030 &  0.635 $\pm$  0.030 &  0.640 $\pm$  0.040 &  0.634 $\pm$  0.039 &  0.623 $\pm$  0.032\\
& $\mathrm{BALC_{KLD}}$ &0.632 $\pm$   0.008 &  0.633 $\pm$  0.008 &  0.630 $\pm$  0.005 &  0.629 $\pm$  0.004 &  0.625 $\pm$  0.008\\ 
& Temporal $\mathrm{BALC_{KLD}}$ & 0.631 $\pm$   0.013 &  0.630 $\pm$  0.013 &  0.637 $\pm$  0.013 &  0.630 $\pm$  0.014 &  0.629 $\pm$  0.009\\
\hline
\multirow{4}{*}{$\mathcal{D}_{4}$} & $\mathrm{BALD_{MCD}}$ &0.475 $\pm$   0.035 &  0.493 $\pm$ 0.025 &  0.471 $\pm$  0.031 &  0.468  $\pm$  0.027 &  0.481 $\pm$  0.035 \\
& $\mathrm{BALD_{MCP}}$  & 0.508 $\pm$   0.024 &  0.512 $\pm$  0.020 &  0.513 $\pm$  0.019 &  0.499 $\pm$  0.012 &  0.492 $\pm$  0.016\\
& $\mathrm{BALC_{KLD}}$ & 0.483 $\pm$   0.031 &  0.476 $\pm$  0.033 &  0.473 $\pm$  0.026 &  0.479 $\pm$  0.021 &  0.479 $\pm$  0.032\\ 
& Temporal $\mathrm{BALC_{KLD}}$ &0.490 $\pm$   0.012 &  0.497 $\pm$  0.030 &  0.466 $\pm$  0.013 &  0.485 $\pm$  0.016 &  0.500 $\pm$ 0.013\\
% \hline
% \multirow{4}{*}{$\mathcal{D}_{5}$} & $\mathrm{BALD_{MCD}}$ & 0.711 $\pm$   0.017 &  0.759 $\pm$  0.023 &  0.738 $\pm$  0.039 &  0.736 $\pm$  0.028 &  0.694 $\pm$  0.021 \\
% & $\mathrm{BALD_{MCP}}$  & 0.704 $\pm$  0.021 &  0.678 $\pm$  0.056 &  0.678 $\pm$  0.074 &  0.709 $\pm$  0.025 &  0.714 $\pm$  0.014\\
% & $\mathrm{BALC_{KLD}}$ & 0.719 $\pm$   0.017 & 0.714 $\pm$  0.012 &  0.685 $\pm$  0.044 &  0.712 $\pm$  0.006 &  0.717 $\pm$  0.012\\ 
% & Temporal $\mathrm{BALC_{KLD}}$ & 0.686 $\pm$   0.070 &  0.718 $\pm$  0.007 &  0.653 $\pm$  0.078 &  0.721 $\pm$  0.007 &  0.721 $\pm$  0.021\\
%\hhline{=======}
\bottomrule
\end{tabular}}
\end{subtable}
\end{table}

\end{subappendices}

\clearpage

\begin{subappendices}
\section{Effect of Number of Monte Carlo Samples, \textit{T}, on Performance}
\label{appendix:effect_of_mcsamples}

The number of MC samples, \textit{T}, within an AL framework can be associated with an improved approximation of the version space. This, in turn, should lead to improved AL results. To quantify the effect of the number of MC samples on performance, we illustrate in Fig.~\ref{fig:impact_of_mcsamples_sup}, the validation AUC for experiments conducted with $T=(5,20,40,100)$. We show that there does not exist a simple proportional relationship between the number of MC samples and performance. This can be seen by the relatively strong generalization performance of models when $T=100$ in Figs.~\ref{fig:entropy_mcsamples}, \ref{fig:temporal_balc_kld_mcsamples}, and \ref{fig:temporal_balc_jsd_mcsamples} and poorer performance when $T=100$. This suggests that our family of methods can perform well without being computationally expensive. 

% \graphicspath{{../impact_of_mcsamples/}}

\begin{figure}[!h]
    \centering
    \begin{subfigure}[h]{0.8\textwidth}
    \centering
    \includegraphics[width=\textwidth]{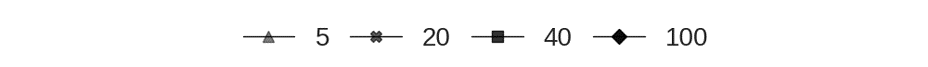}
    \end{subfigure}
    ~
    \begin{subfigure}[h]{0.40\textwidth}
        \centering
        \includegraphics[width=\textwidth]{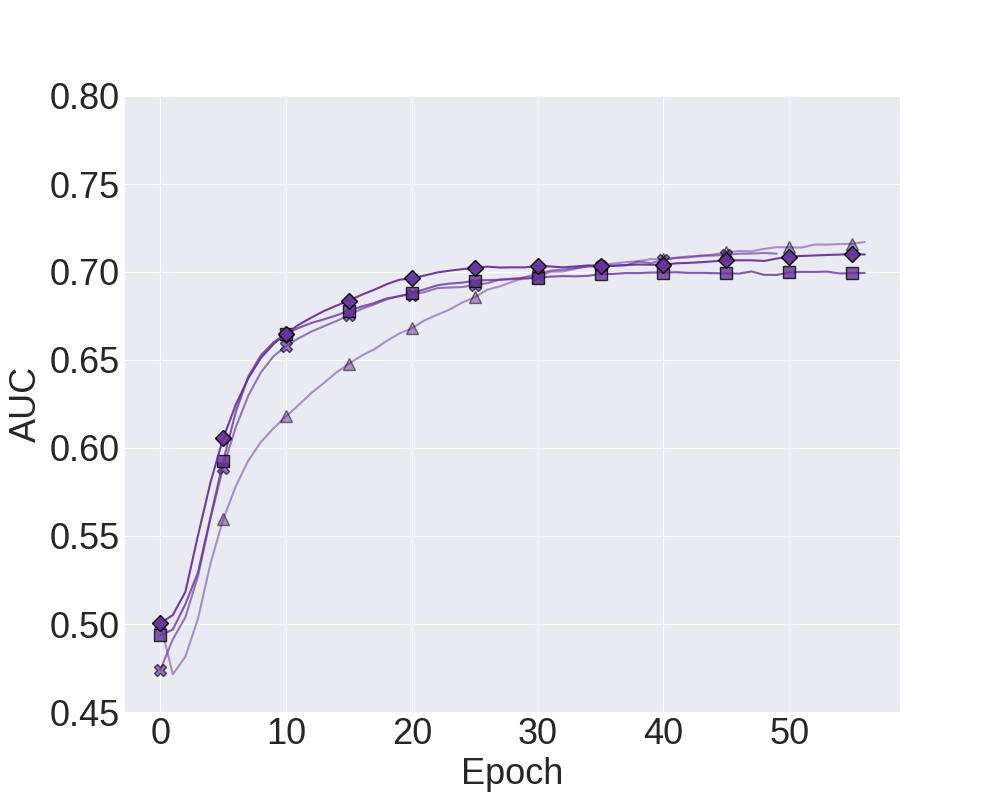}
        \caption{Var Ratio}
    \end{subfigure}
    ~ 
    \begin{subfigure}[h]{0.40\textwidth}
        \centering
        \includegraphics[width=\textwidth]{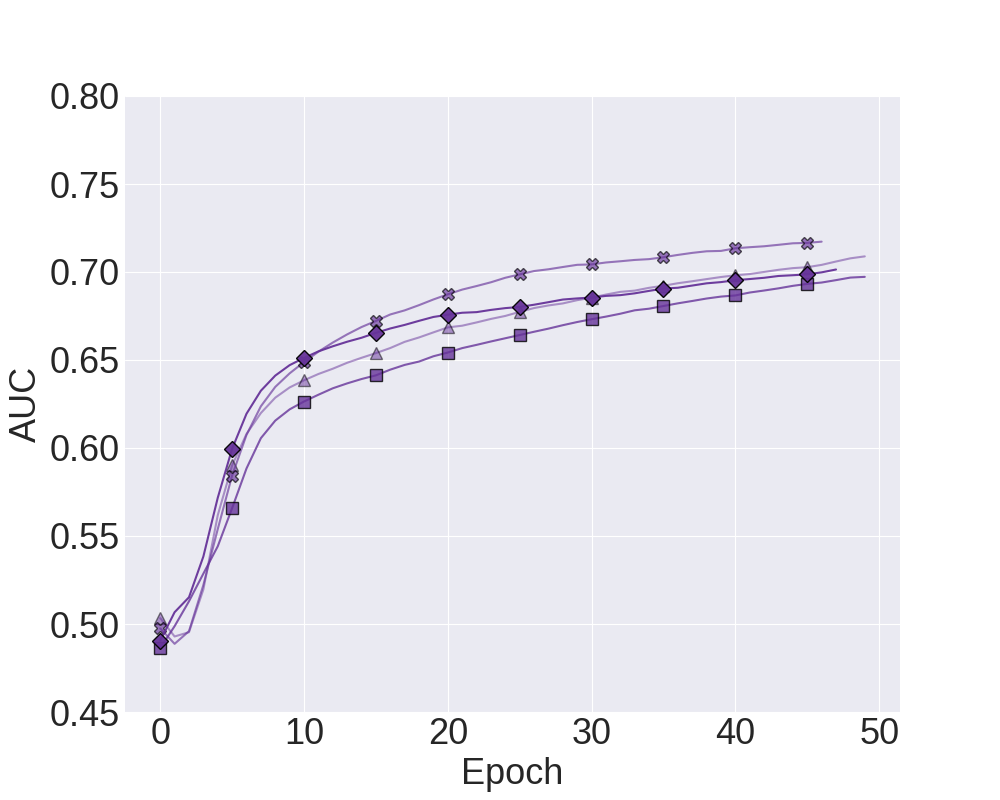}
        \caption{Temporal Var Ratio}
    \end{subfigure}
    ~ 
    \begin{subfigure}[h]{0.40\textwidth}
        \centering
        \includegraphics[width=\textwidth]{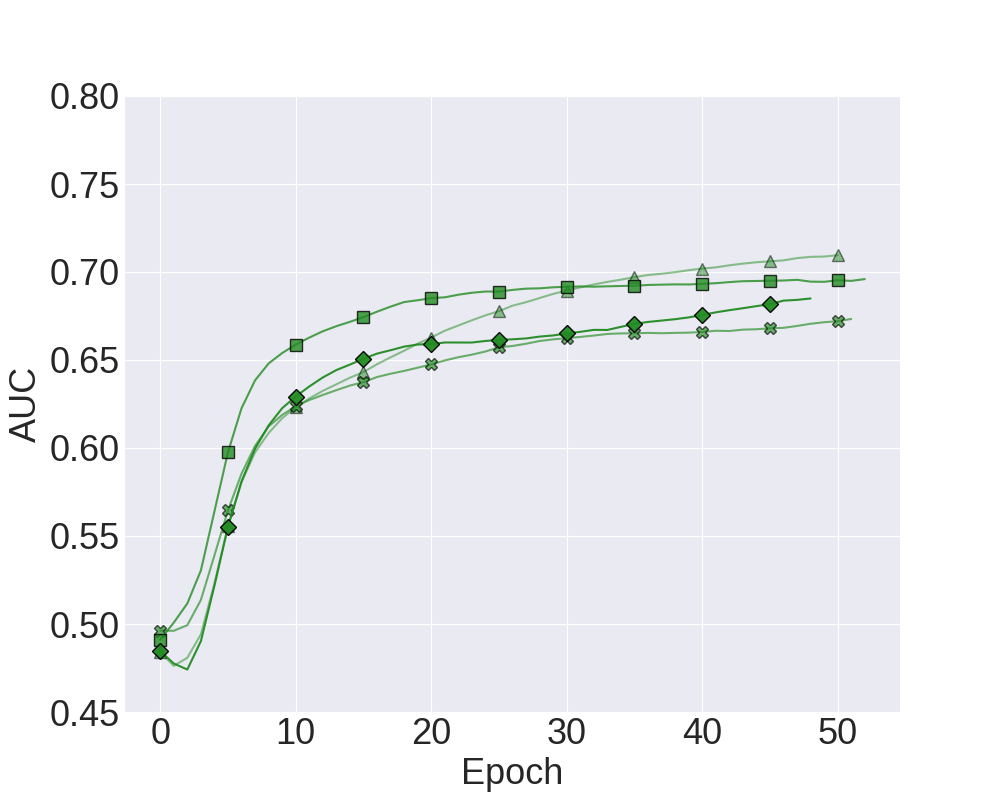}
        \caption{Entropy}
        \label{fig:entropy_mcsamples}
    \end{subfigure}
    ~ 
    \begin{subfigure}[h]{0.40\textwidth}
        \centering
        \includegraphics[width=\textwidth]{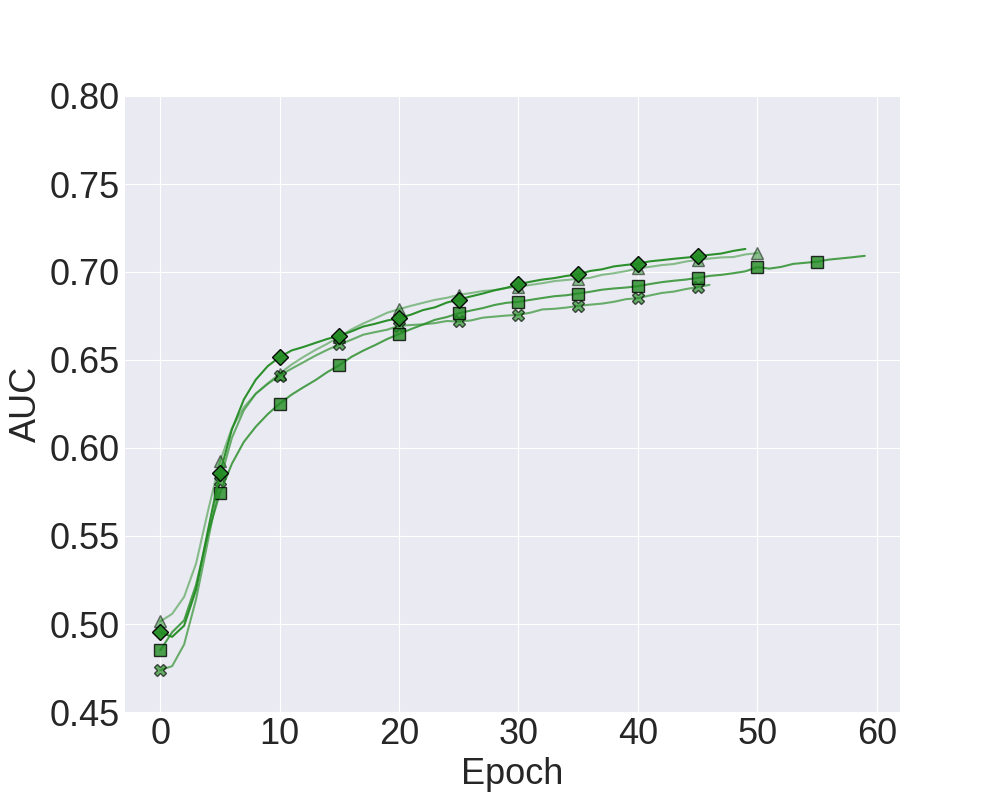}
        \caption{Temporal Entropy}
    \end{subfigure}
    ~ 
    \begin{subfigure}[h]{0.40\textwidth}
        \centering
        \includegraphics[width=\textwidth]{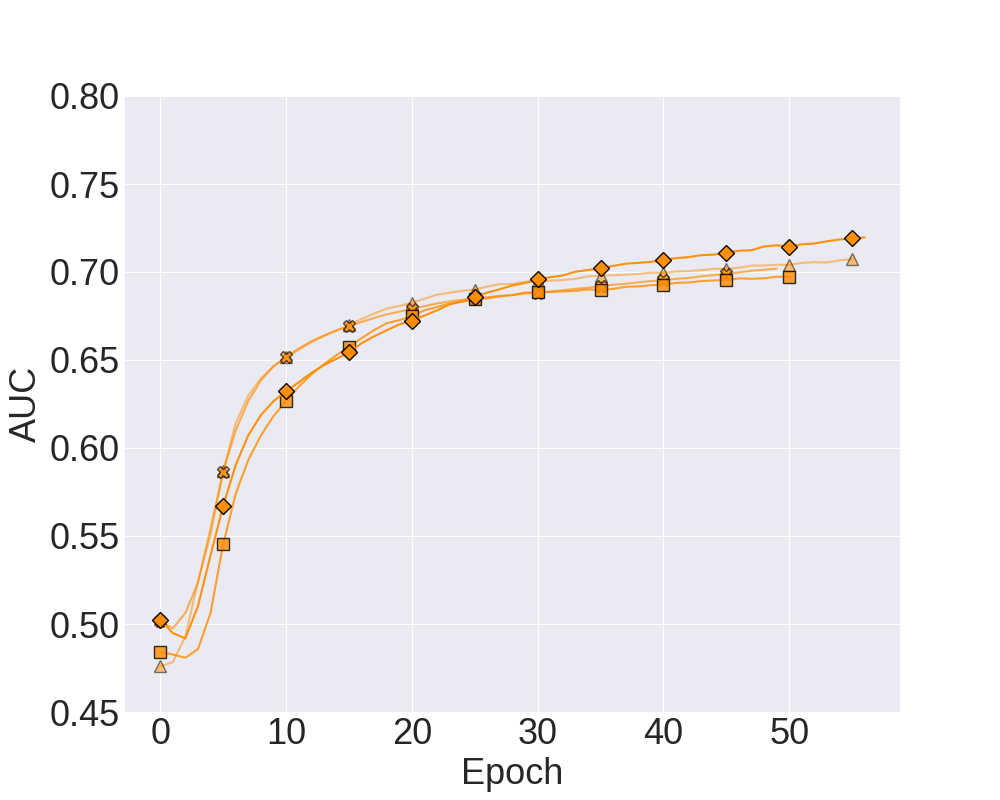}
        \caption{BALD}
    \end{subfigure}
    ~ 
    \begin{subfigure}[h]{0.40\textwidth}
        \centering
        \includegraphics[width=\textwidth]{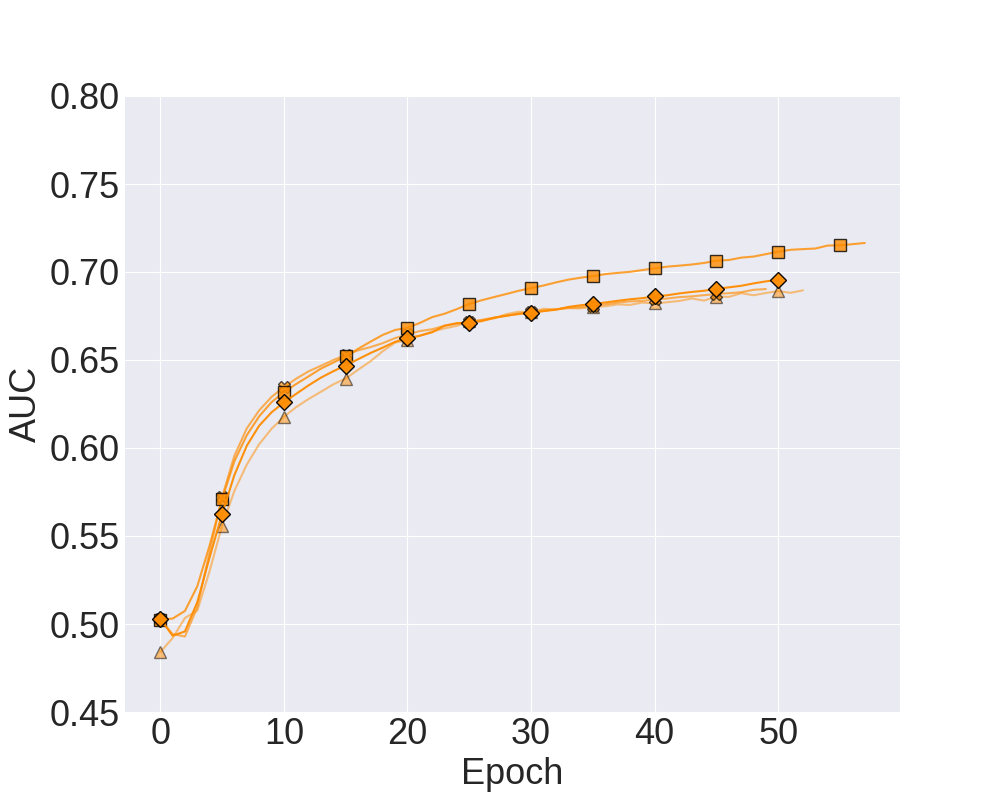}
        \caption{Temporal BALD}
    \end{subfigure}
\end{figure}

\begin{figure}[ht]\ContinuedFloat
    \centering
    \begin{subfigure}[h]{0.40\textwidth}
        \centering
        \includegraphics[width=\textwidth]{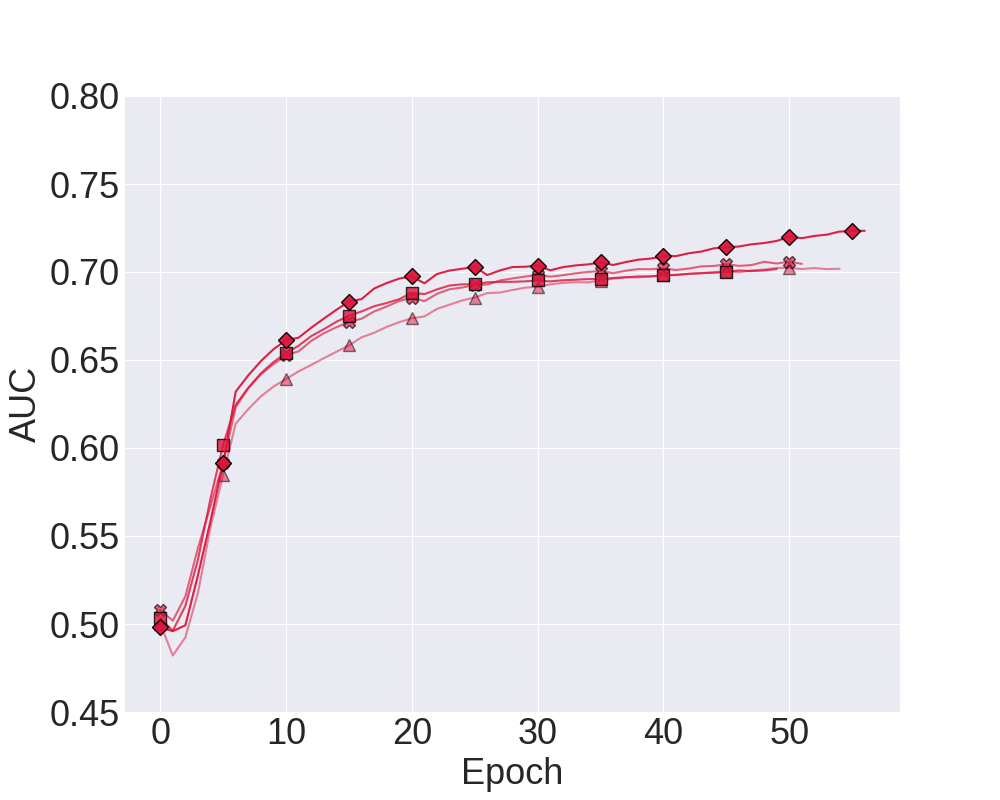}
        \caption{BALC\textsubscript{KLD}}
    \end{subfigure}
    ~ 
    \begin{subfigure}[h]{0.40\textwidth}
        \centering
        \includegraphics[width=\textwidth]{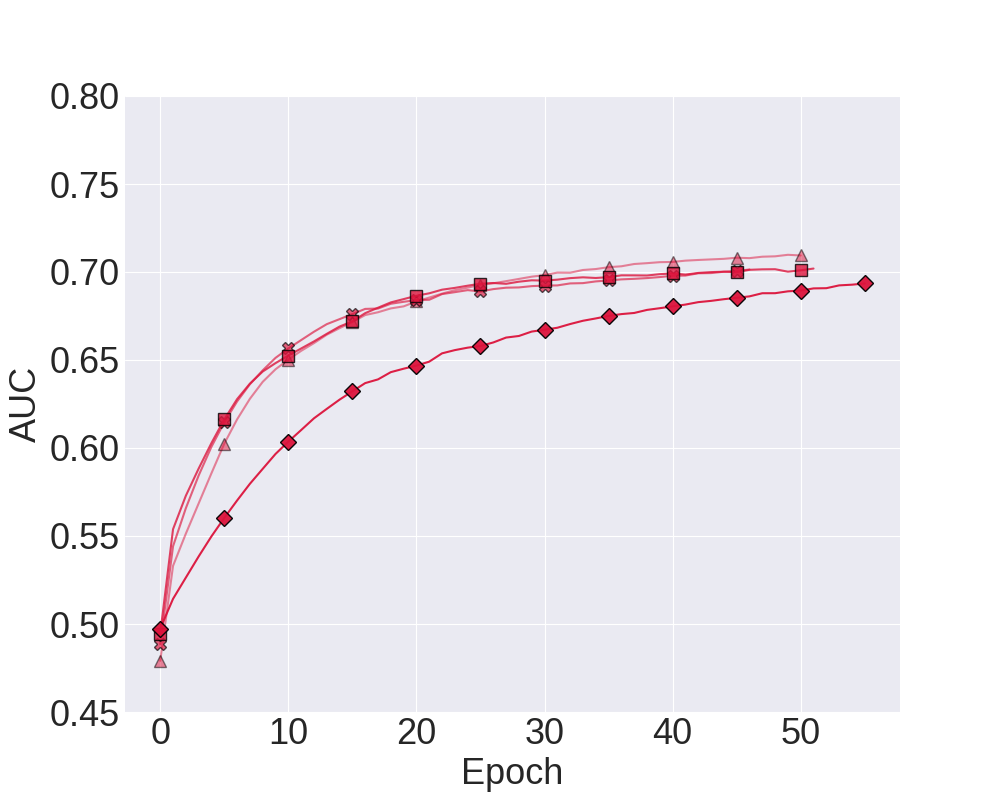}
        \caption{Temporal BALC\textsubscript{KLD}}
        \label{fig:temporal_balc_kld_mcsamples}
    \end{subfigure}
    ~ 
    \begin{subfigure}[h]{0.40\textwidth}
        \centering
        \includegraphics[width=\textwidth]{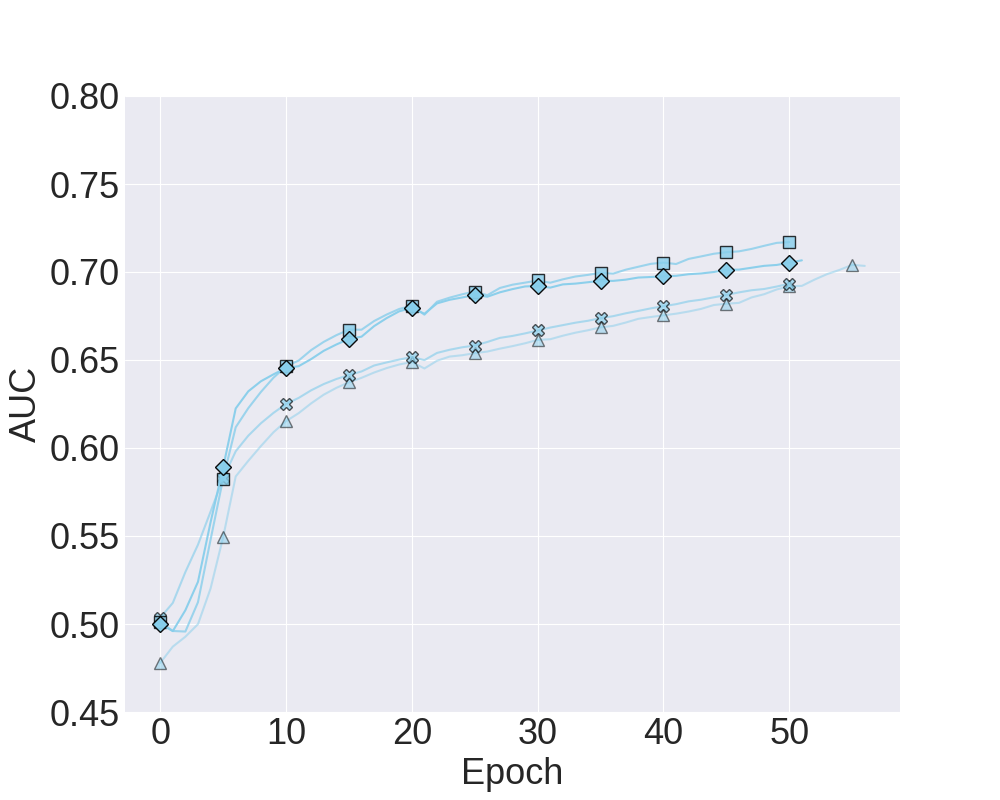}
        \caption{BALC\textsubscript{JSD}}
        \label{fig:temporal_balc_jsd_mcsamples}
    \end{subfigure}
    ~ 
    \begin{subfigure}[h]{0.40\textwidth}
        \centering
        \includegraphics[width=\textwidth]{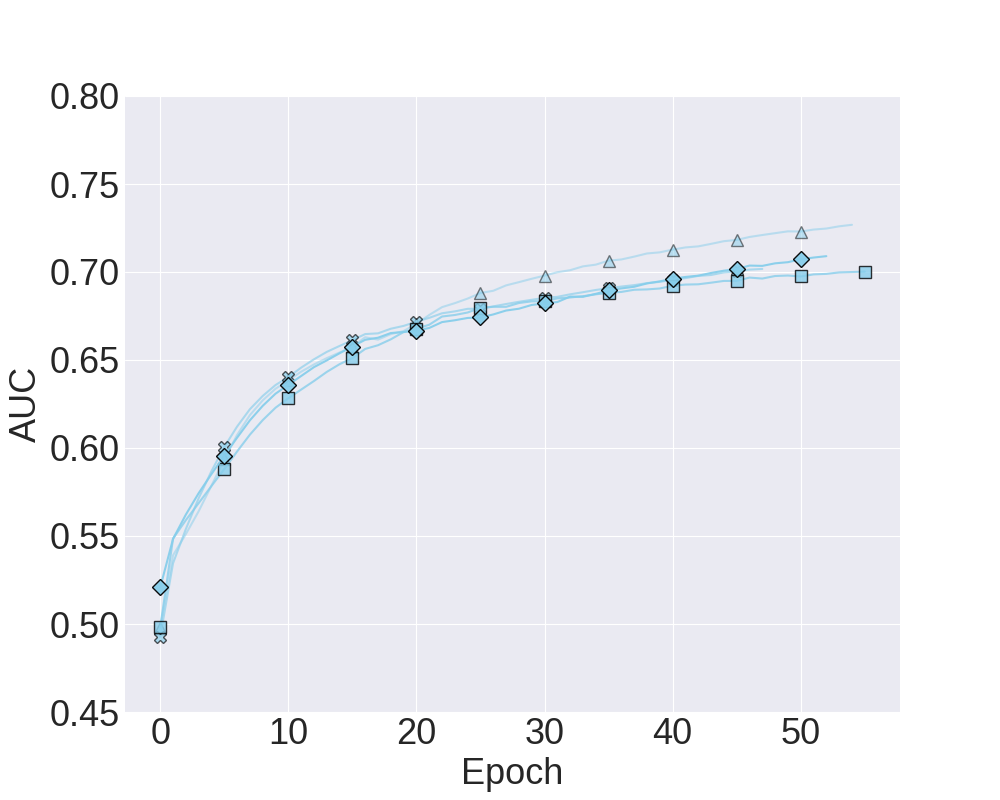}
        \caption{Temporal BALC\textsubscript{JSD}}
    \end{subfigure}
    ~ 
    \caption{Mean validation AUC as a function of number of Monte Carlo samples \textit{T} for the different acquisition functions using the MCP method. The acquisition percentage and acquisition epochs were fixed at \textit{b} = 2\% and $\tau=5$, respectively. These experiments are performed on $\mathcal{D}_{2}$ at a fraction of $\beta=0.5$. Results are averaged across 5 seeds.}
    \label{fig:impact_of_mcsamples_sup}
\end{figure}

\end{subappendices}

\clearpage

\begin{subappendices}
\section{Effect of Acquisition Percentage, \textit{b}, on Performance}
\label{appendix:effect_of_aqpercent}

The number of unlabelled instances acquired during the AL procedure can have a strong effect on the generalization performance of networks. We investigate the effect of this on our family of methods and illustrate the results in Fig.~\ref{fig:impact_of_aqpercent_sup} when conducting experiments for $b=(1\%,2\%,5\%,20\%)$. Contrary to expectations that more acquisition is better, we show that acquiring large amounts of data is actually detrimental. This can be seen by the poorer performance attributed to $b=20\%$ in, for instance, Figs.~\ref{fig:temporal_var_ratio_aqpercent}, \ref{fig:temporal_bald_aqpercent}, and \ref{fig:balc_kld_aqpercent}. We hypothesize that this is due to larger magnitude 1) distribution shifts and 2) label noise brought about by the absence of an oracle. 

% \graphicspath{{../impact_of_aqpercent/}}

\begin{figure}[!h]
    \centering
    \begin{subfigure}[h]{0.8\textwidth}
    \centering
    \includegraphics[width=\textwidth]{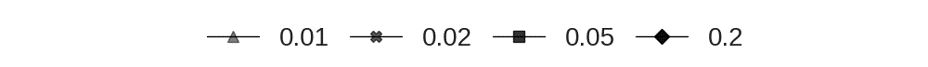}
    \end{subfigure}
    ~
    \begin{subfigure}[h]{0.40\textwidth}
        \centering
        \includegraphics[width=\textwidth]{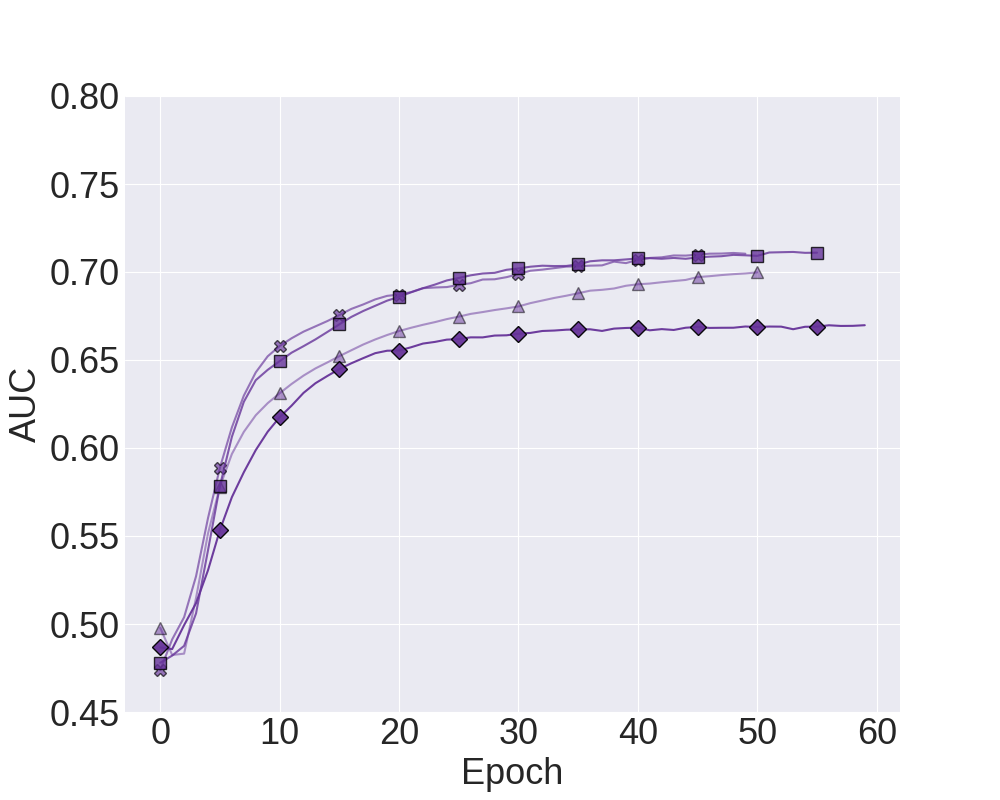}
        \caption{Var Ratio}
    \end{subfigure}
    ~ 
    \begin{subfigure}[h]{0.40\textwidth}
        \centering
        \includegraphics[width=\textwidth]{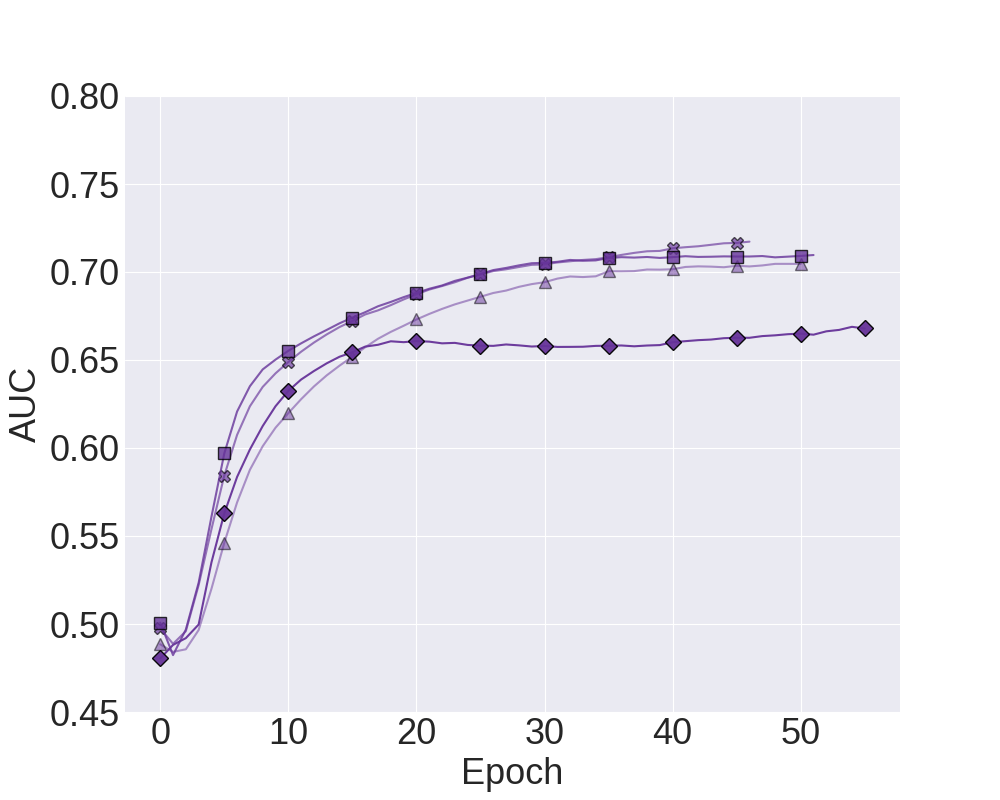}
        \caption{Temporal Var Ratio}
        \label{fig:temporal_var_ratio_aqpercent}
    \end{subfigure}
    ~ 
    \begin{subfigure}[h]{0.40\textwidth}
        \centering
        \includegraphics[width=\textwidth]{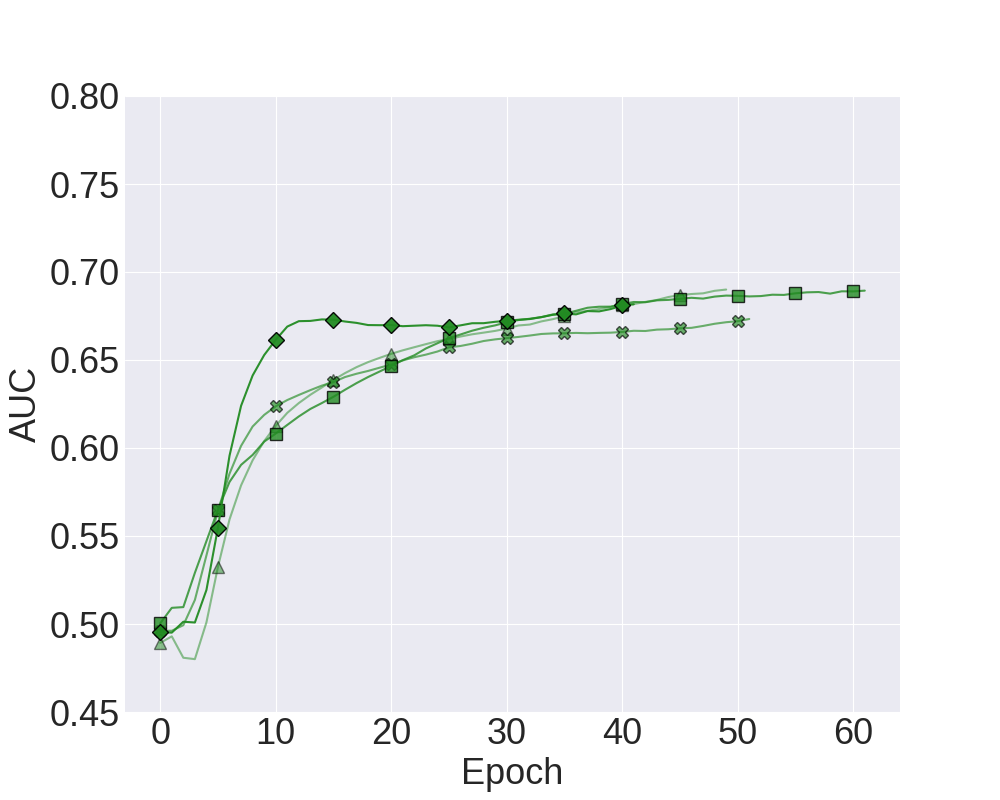}
        \caption{Entropy}
    \end{subfigure}
    ~ 
    \begin{subfigure}[h]{0.40\textwidth}
        \centering
        \includegraphics[width=\textwidth]{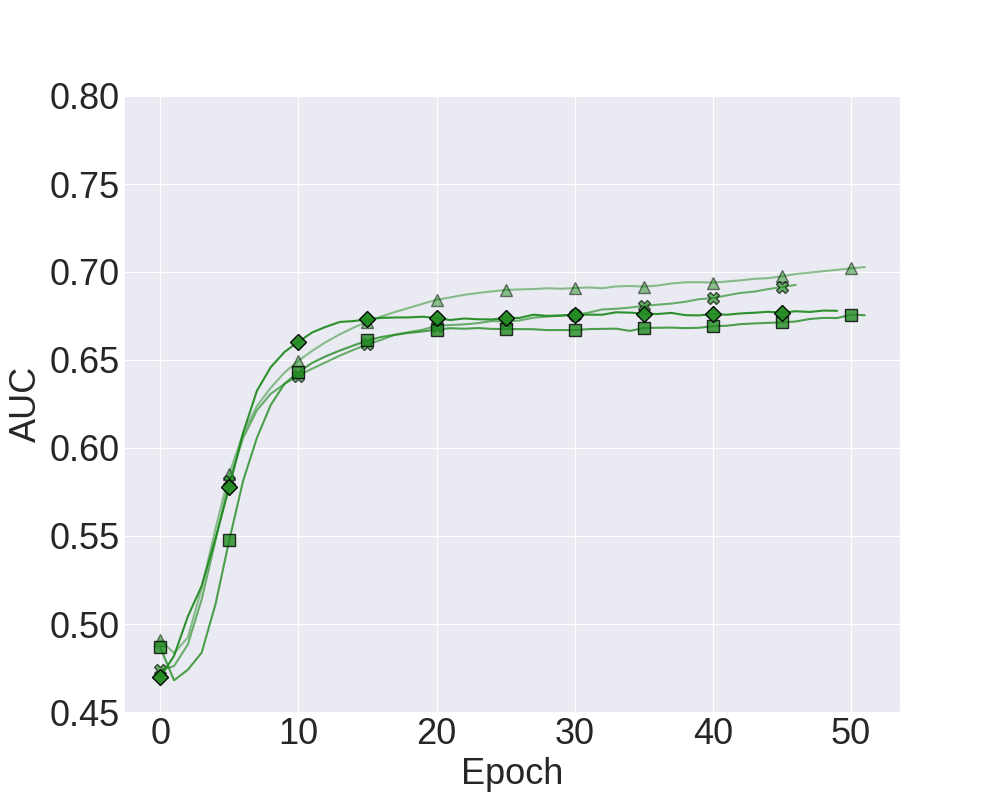}
        \caption{Temporal Entropy}
    \end{subfigure}
    ~ 
    \begin{subfigure}[h]{0.40\textwidth}
        \centering
        \includegraphics[width=\textwidth]{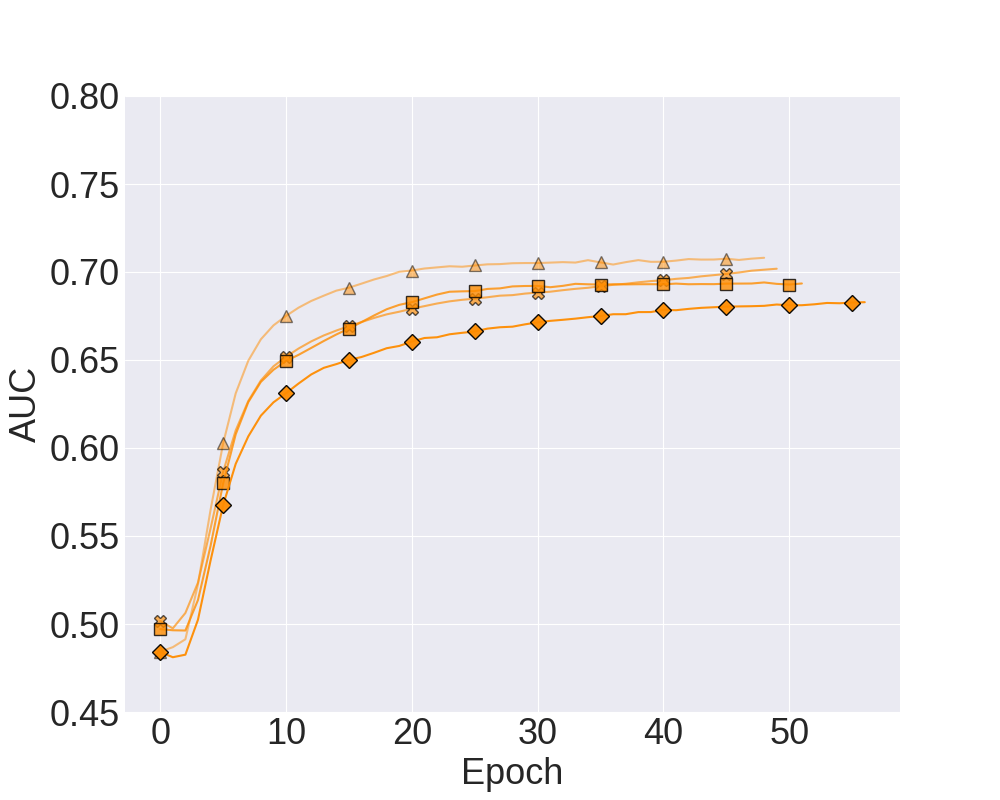}
        \caption{BALD}
    \end{subfigure}
    ~ 
    \begin{subfigure}[h]{0.40\textwidth}
        \centering
        \includegraphics[width=\textwidth]{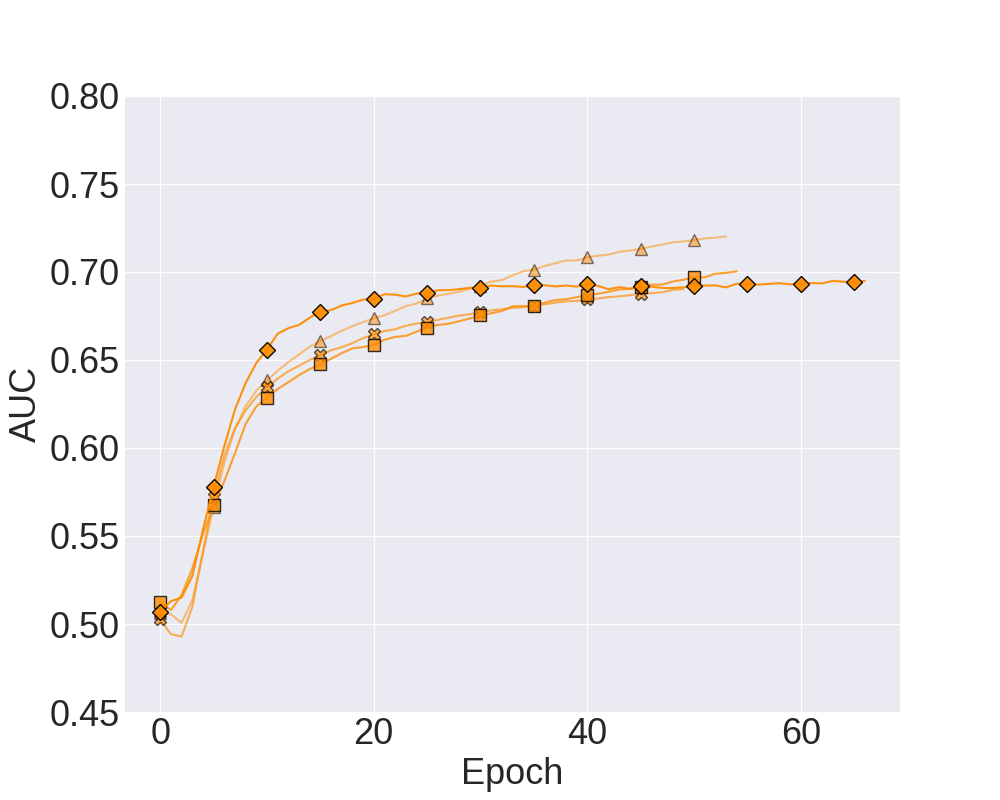}
        \caption{Temporal BALD}
        \label{fig:temporal_bald_aqpercent}
    \end{subfigure}
\end{figure}

\begin{figure}[ht]\ContinuedFloat
    \centering
    \begin{subfigure}[h]{0.40\textwidth}
        \centering
        \includegraphics[width=\textwidth]{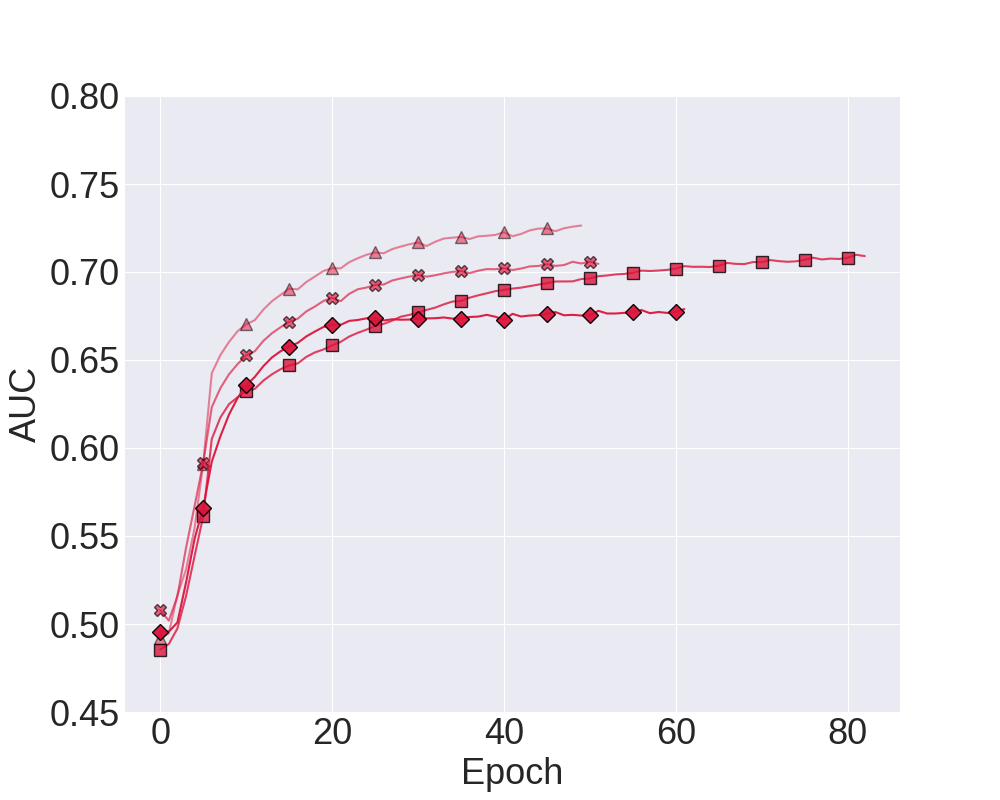}
        \caption{BALC\textsubscript{KLD}}
        \label{fig:balc_kld_aqpercent}
    \end{subfigure}
    ~ 
    \begin{subfigure}[h]{0.40\textwidth}
        \centering
        \includegraphics[width=\textwidth]{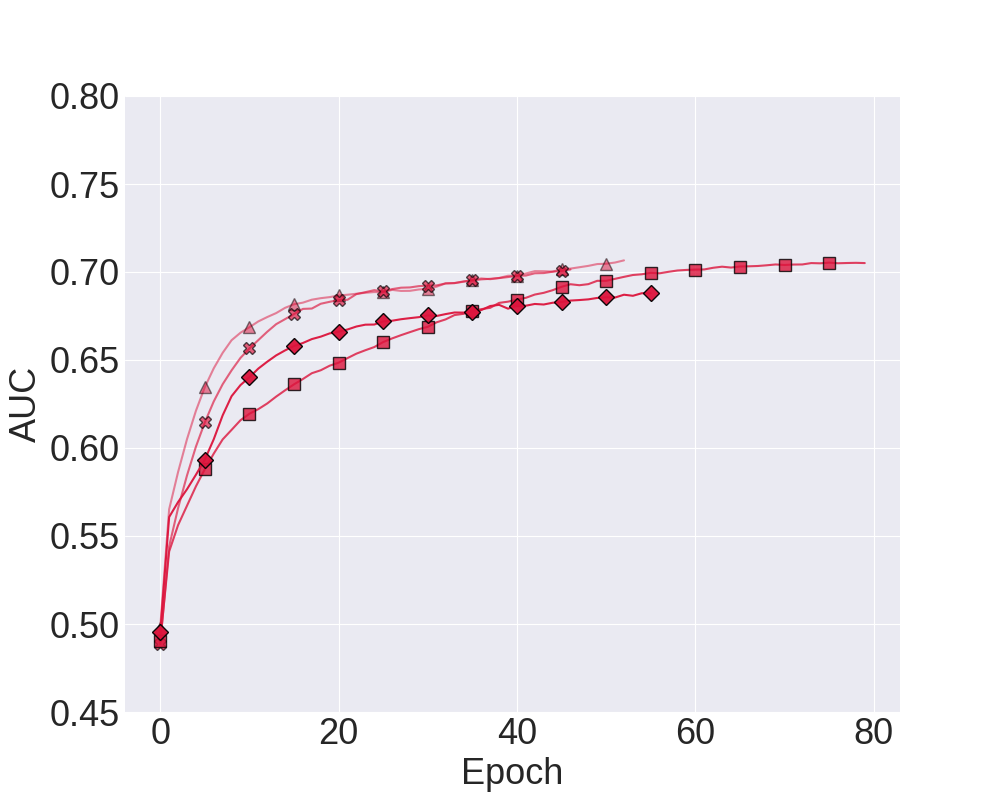}
        \caption{Temporal BALC\textsubscript{KLD}}
    \end{subfigure}
    ~ 
    \begin{subfigure}[h]{0.40\textwidth}
        \centering
        \includegraphics[width=\textwidth]{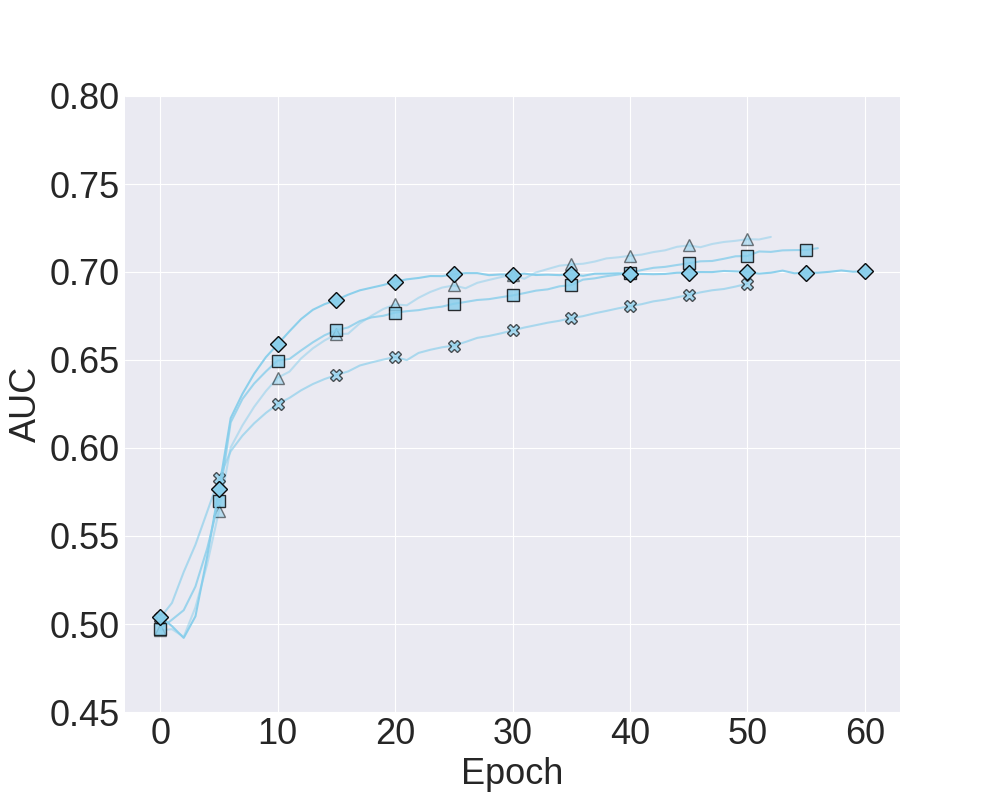}
        \caption{BALC\textsubscript{JSD}}
    \end{subfigure}
    ~ 
    \begin{subfigure}[h]{0.40\textwidth}
        \centering
        \includegraphics[width=\textwidth]{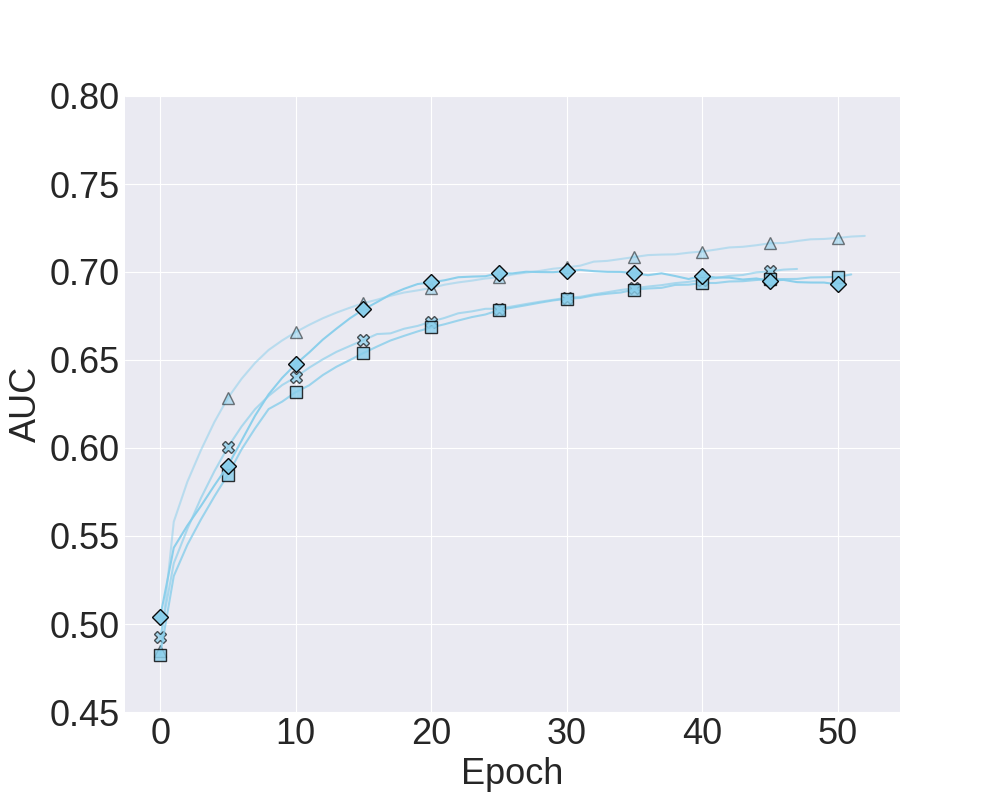}
        \caption{Temporal BALC\textsubscript{JSD}}
    \end{subfigure}
    ~ 
    \caption{Mean AUC of the validation set as a function of acquisition percentage \textit{b} for the different acquisition functions using the MCP method. These experiments are performed on $\mathcal{D}_{2}$ at a fraction $\beta= 0.5$. MC samples and acquisition epochs were fixed at \textit{T} = 20 and $\tau=5$, respectively. Results are averaged across 5 seeds. }
    \label{fig:impact_of_aqpercent_sup}
\end{figure}

\end{subappendices}

\clearpage

\begin{subappendices}
\section{Effect of Acquisition Epochs, $\tau$, on Performance}
\label{appendix:effect_of_aqinterval}

As outlined in the main manuscript, the control vs. shock trade-off must be balanced to ensure good generalization performance of an AL procedure. Acquiring instances too early and frequently can lead to instabilities in the training procedure. Conversely, inadequate sampling of unlabelled instances starves the network of much needed data. To quantify this trade-off, we illustrate in Fig.~\ref{fig:impact_of_aqinterval_sup}, the performance of our family of methods when $\tau=(5,10,15,20)$. Although one value that guarantees best performance for all experiments does not exist, $\tau=10$ or $\tau=15$ seem to outperform the others, on average. 

% \graphicspath{{../impact_of_aqinterval/}}

\begin{figure}[!h]
    \centering
    \begin{subfigure}[h]{0.8\textwidth}
    \centering
    \includegraphics[width=\textwidth]{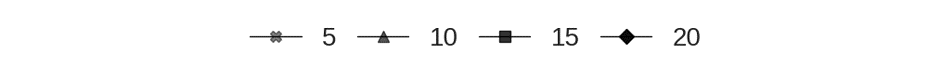}
    \end{subfigure}
    ~
    \begin{subfigure}[h]{0.40\textwidth}
        \centering
        \includegraphics[width=\textwidth]{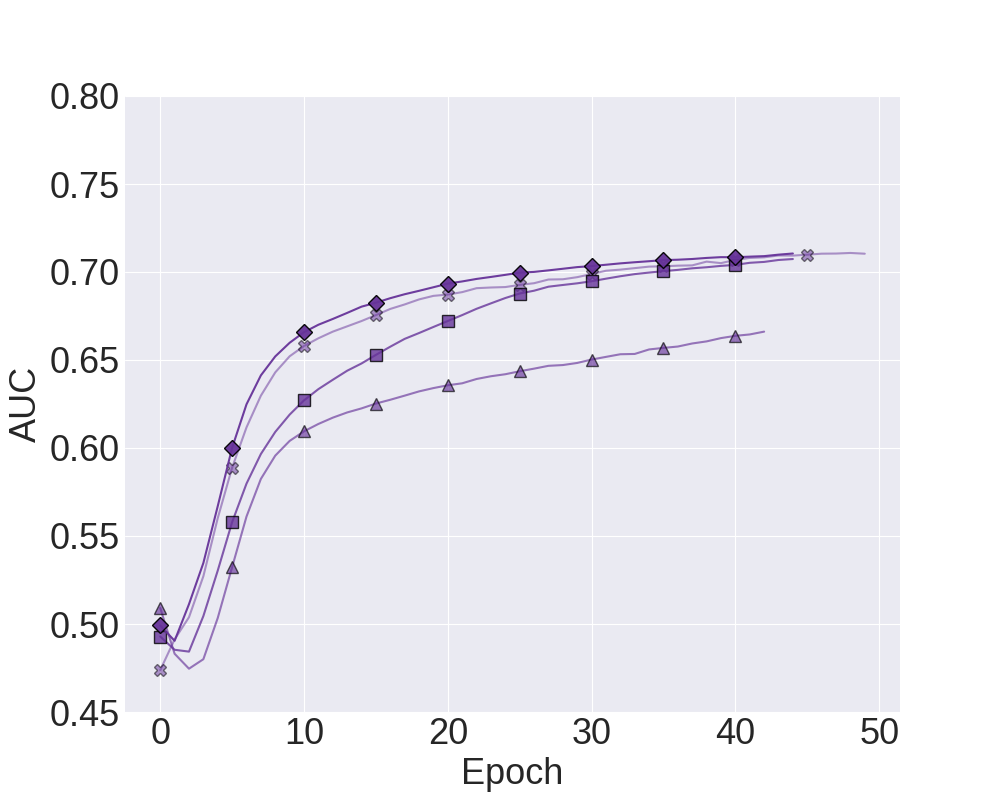}
        \caption{Var Ratio}
    \end{subfigure}
    ~ 
    \begin{subfigure}[h]{0.40\textwidth}
        \centering
        \includegraphics[width=\textwidth]{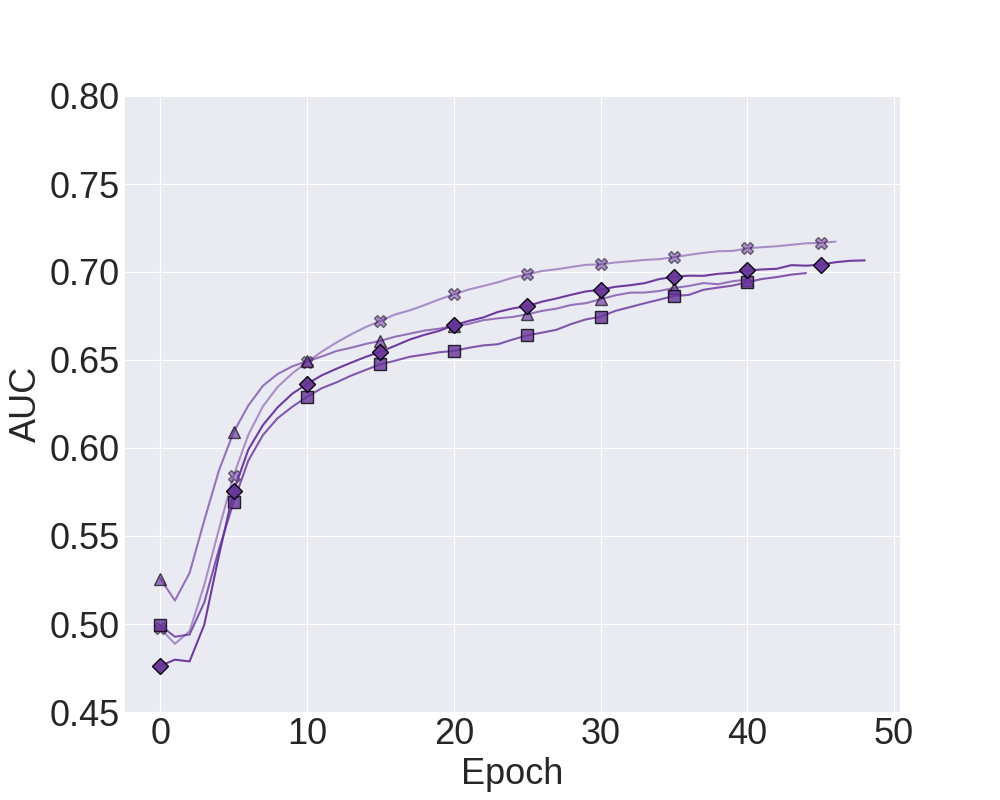}
        \caption{Temporal Var Ratio}
    \end{subfigure}
    ~ 
    \begin{subfigure}[h]{0.40\textwidth}
        \centering
        \includegraphics[width=\textwidth]{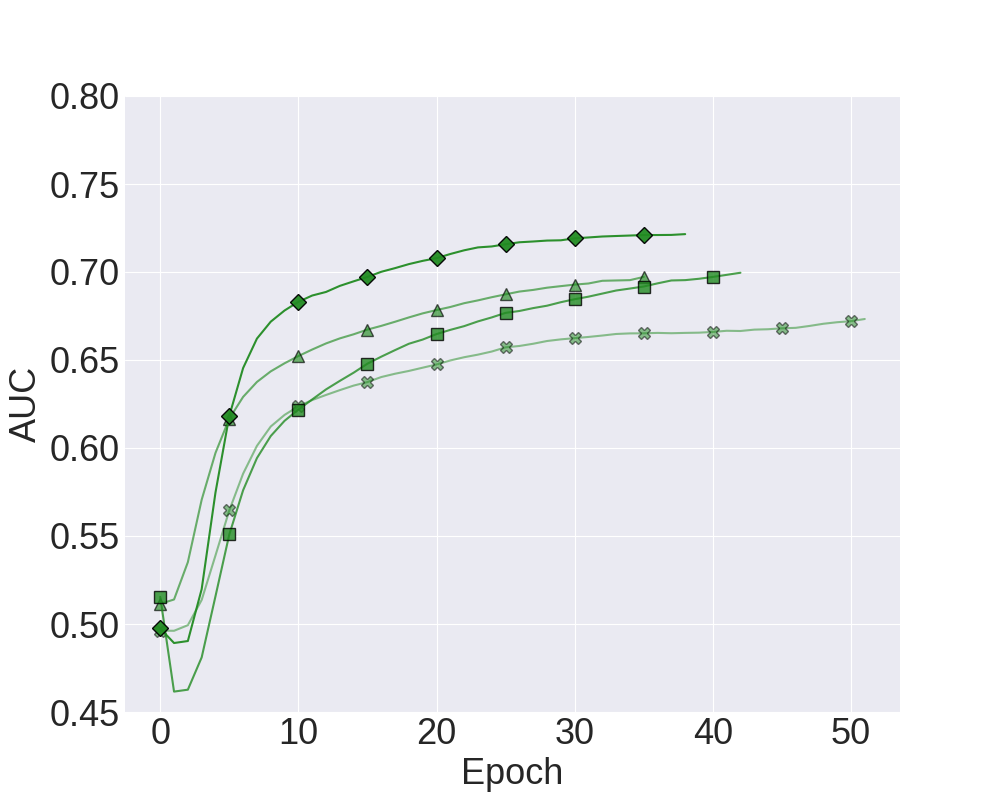}
        \caption{Entropy}
    \end{subfigure}
    ~ 
    \begin{subfigure}[h]{0.40\textwidth}
        \centering
        \includegraphics[width=\textwidth]{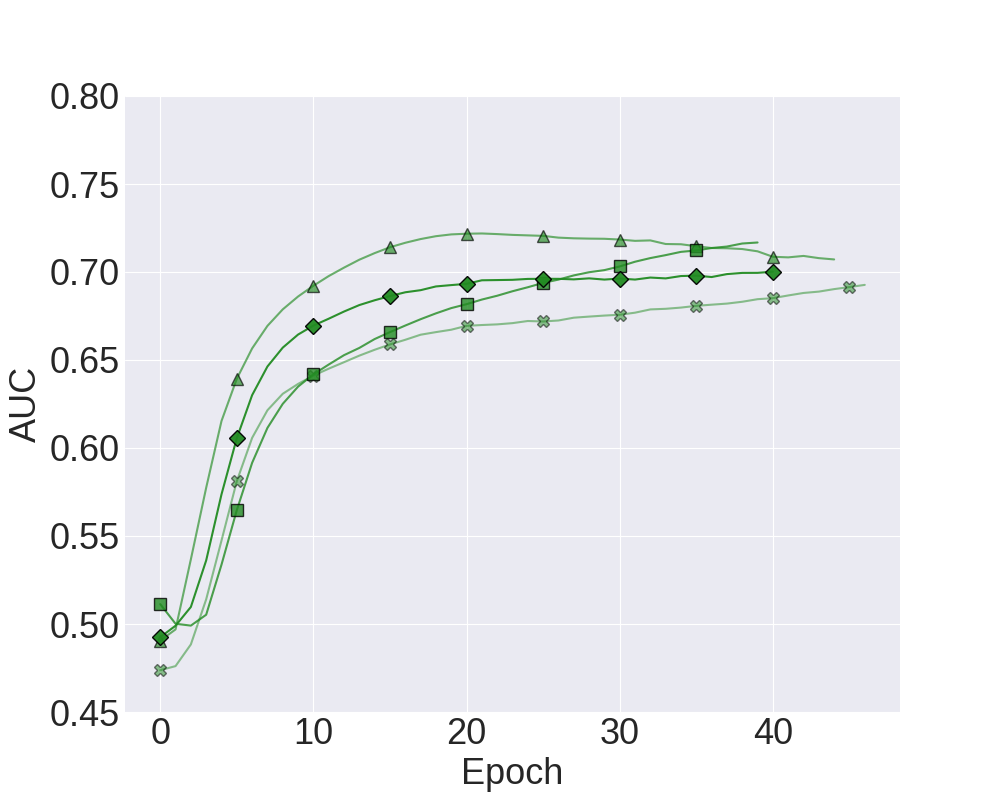}
        \caption{Temporal Entropy}
    \end{subfigure}
    ~ 
    \begin{subfigure}[h]{0.40\textwidth}
        \centering
        \includegraphics[width=\textwidth]{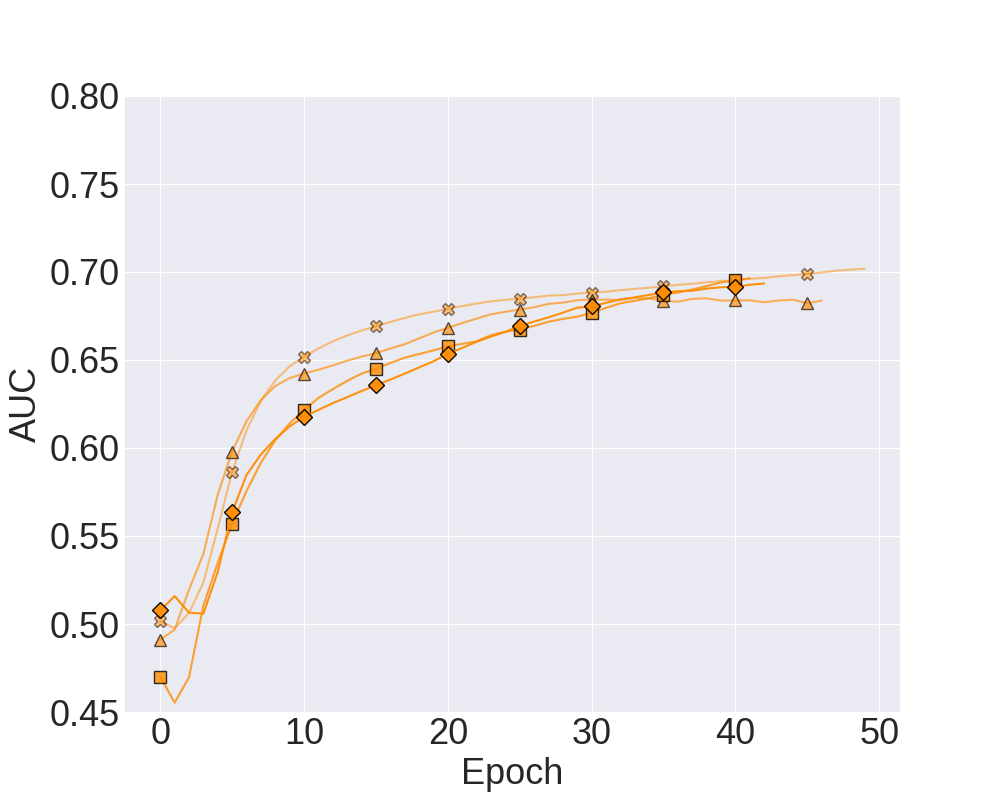}
        \caption{BALD}
    \end{subfigure}
    ~ 
    \begin{subfigure}[h]{0.40\textwidth}
        \centering
        \includegraphics[width=\textwidth]{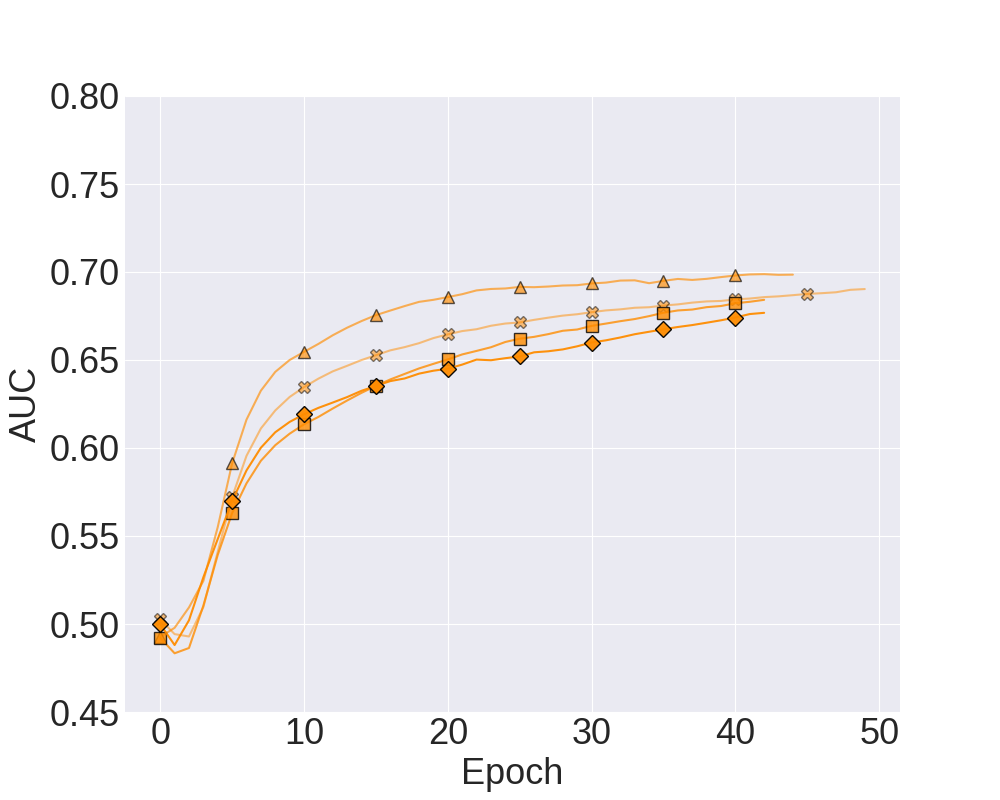}
        \caption{Temporal BALD}
    \end{subfigure}
\end{figure}

\begin{figure}[ht]\ContinuedFloat
    \centering
    \begin{subfigure}[h]{0.40\textwidth}
        \centering
        \includegraphics[width=\textwidth]{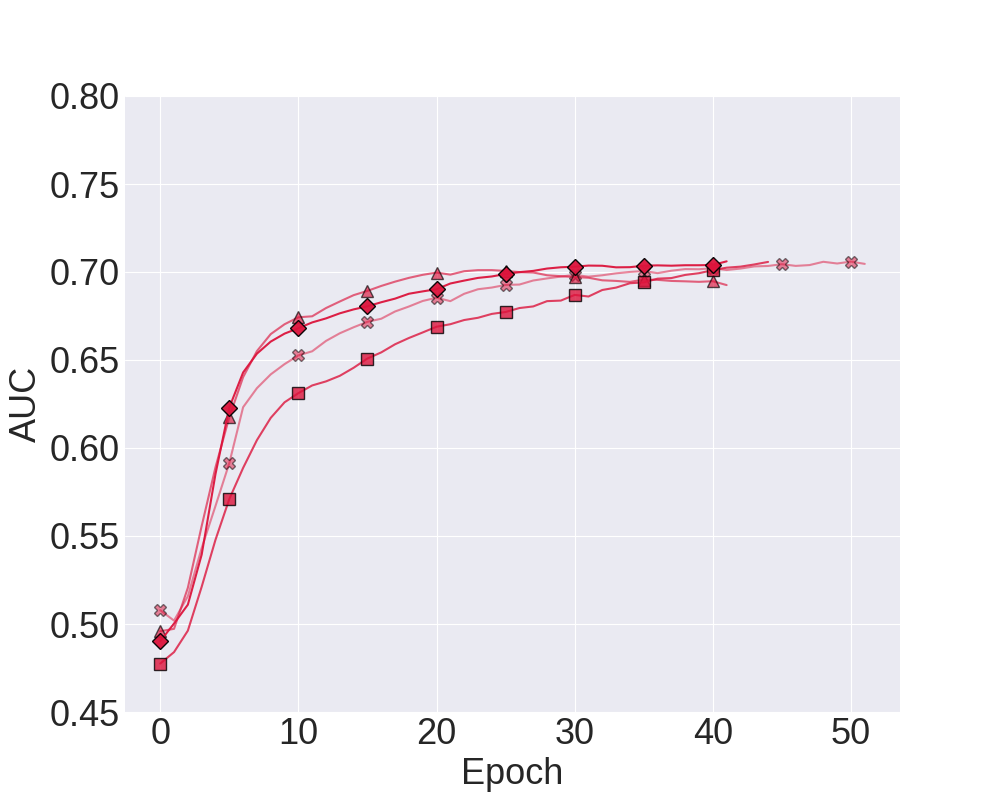}
        \caption{BALC\textsubscript{KLD}}
    \end{subfigure}
    ~ 
    \begin{subfigure}[h]{0.40\textwidth}
        \centering
        \includegraphics[width=\textwidth]{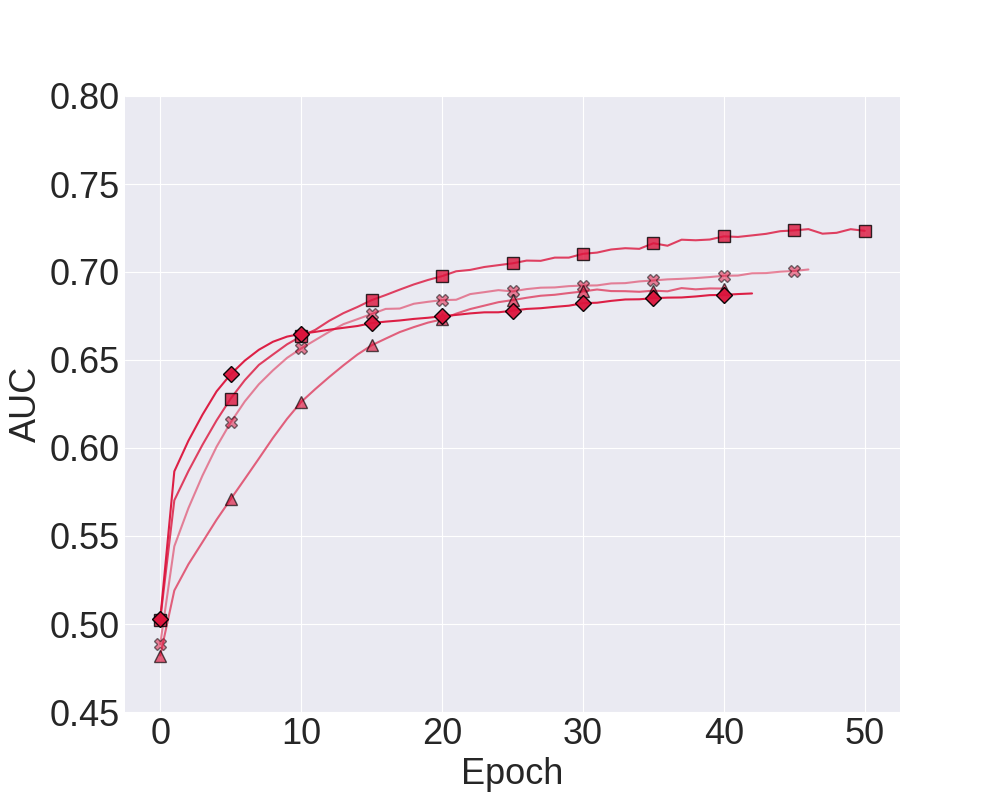}
        \caption{Temporal BALC\textsubscript{KLD}}
    \end{subfigure}
    ~ 
    \begin{subfigure}[h]{0.40\textwidth}
        \centering
        \includegraphics[width=\textwidth]{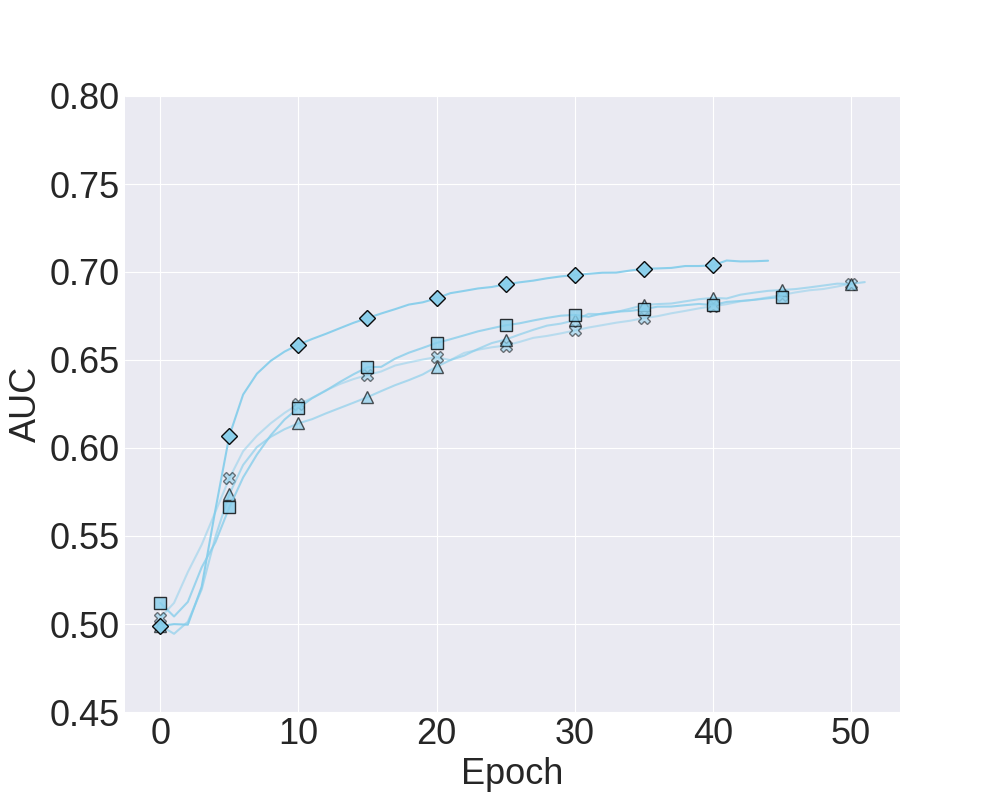}
        \caption{BALC\textsubscript{JSD}}
    \end{subfigure}
    ~ 
    \begin{subfigure}[h]{0.40\textwidth}
        \centering
        \includegraphics[width=\textwidth]{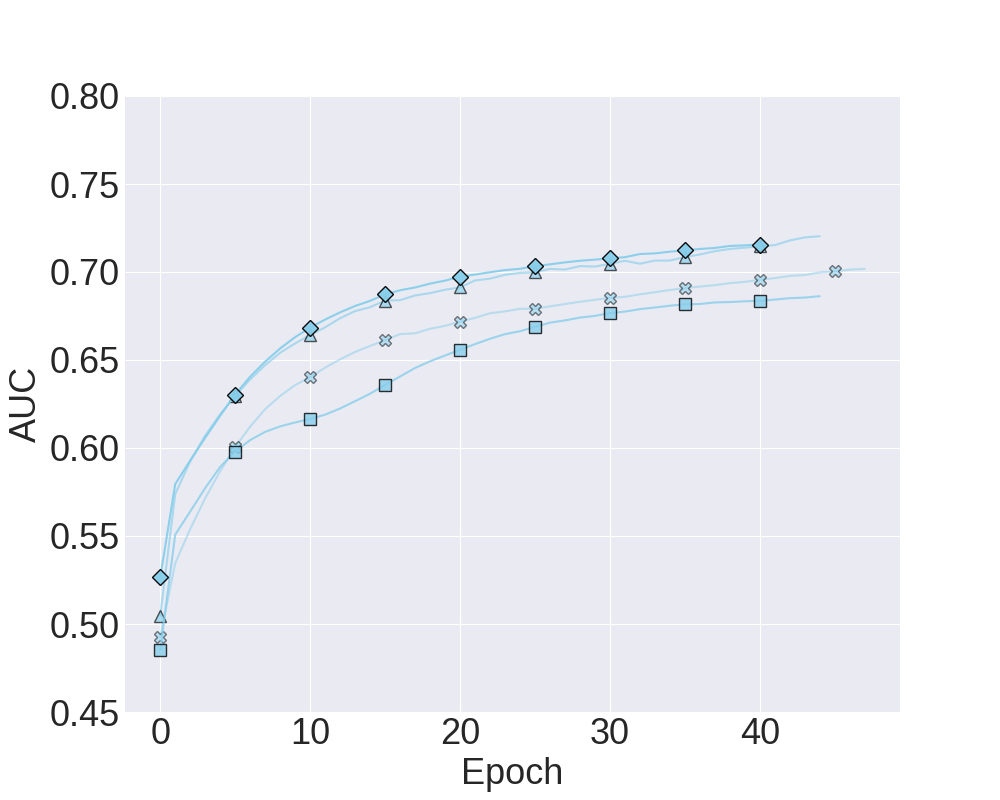}
        \caption{Temporal BALC\textsubscript{JSD}}
    \end{subfigure}
    ~ 
    \caption{Mean AUC of the validation set as a function of acquisition epochs $\tau$ for the different acquisition functions using the MCP method. MC samples and the acquistion percentage were fixed at \textit{T} = 20 and \textit{b} = 2\%, respectively. These experiments are performed on $\mathcal{D}_{2}$ at a fraction $\beta=0.5$. Results are averaged across 5 seeds.}
    \label{fig:impact_of_aqinterval_sup}
\end{figure}

\end{subappendices}

\end{document}